\documentclass[11pt,onecolumn,draftcls,doublespace]{IEEEtran}
\usepackage{cite}

\ifCLASSINFOpdf
  \usepackage[pdftex]{graphicx}
  \graphicspath{{./figs/}}
  \DeclareGraphicsExtensions{.pdf}
\else
  \usepackage[dvips]{graphicx}
  \graphicspath{{./figs/}}
  \DeclareGraphicsExtensions{.eps}
\fi

\usepackage[cmex10]{amsmath}
\usepackage{amssymb,tabularx,multirow}
\usepackage{algorithmic}
\usepackage{array}

\ifCLASSOPTIONcompsoc
  \usepackage[caption=false,font=normalsize,labelfont=sf,textfont=sf]{subfig}
\else
  \usepackage[caption=false,font=footnotesize]{subfig}
\fi

\usepackage{fixltx2e}
\usepackage{dblfloatfix}

\ifCLASSOPTIONcaptionsoff
  \usepackage[nomarkers]{endfloat}
 \let\MYoriglatexcaption\caption
 \renewcommand{\caption}[2][\relax]{\MYoriglatexcaption[#2]{#2}}
\fi
\usepackage{url}

\allowdisplaybreaks

\title{Online Low-Rank Subspace Learning from Incomplete Data: A Bayesian View}

\author{Paris V. Giampouras,
        Athanasios~A.~Rontogiannis,~\IEEEmembership{Member,~IEEE,}
        ~Konstantinos~E.~Themelis and~Konstantinos~D.~Koutroumbas%
\thanks{P.V. Giampouras, A.A. Rontogiannis, K.E. Themelis and K.D. Koutroumbas are with the Institute for Astronomy, Astrophysics, Space Applications and Remote Sensing (IAASARS), National Observatory of Athens, I. Metaxa \& Vas. Pavlou str., GR-15236, Penteli, Greece (e-mail: parisg@noa.gr; tronto@noa.gr; themelis@noa.gr;  koutroum@noa.gr). }
}

\begin{document}
\maketitle
\begin{abstract}
Extracting the underlying low-dimensional space where high-dimensional signals often reside has long been at the center of numerous algorithms in the signal processing and machine learning literature during the past few decades. At the same time, working with incomplete (partly observed) large scale datasets has recently been commonplace for diverse reasons. This so called {\it big data era} we are currently living calls for devising online subspace learning algorithms that can suitably handle incomplete data. Their envisaged objective is to {\it recursively} estimate the unknown subspace by processing streaming data sequentially, thus reducing computational complexity, while obviating the need for storing the whole dataset in memory. In this paper, an online variational Bayes subspace learning algorithm from partial observations is presented. To account for the unawareness of the true rank of the subspace, commonly met in practice, low-rankness is explicitly imposed on the sought subspace data matrix by exploiting sparse Bayesian learning principles. Moreover, sparsity,  {\it simultaneously} to low-rankness, is favored on the subspace matrix by the sophisticated hierarchical Bayesian scheme  that is adopted. In doing so, the proposed algorithm becomes adept in dealing with applications whereby the underlying subspace may be also sparse, as, e.g., in sparse dictionary learning problems. As shown, the new subspace tracking scheme outperforms its state-of-the-art counterparts in terms of estimation accuracy, in a variety of experiments conducted on simulated and real data. 
\end{abstract}


\maketitle



\section{Introduction}
\label{sec:intro}
Recent years are by all means characterized by the vast amounts of data, commonly named with the blanket term {\it big data}, generated by a wealth of sources such as social media, environmental monitoring sensors, medical application devices, e-commerce sites etc. to mention just a few. In first place, having at hand a lot of data seems to be fairly advantageous. However, enjoying the merits emerging from this so called data deluge raises a number of issues needed to be properly addressed. Among other things, computational complexity and memory storage requirements are undoubtedly two basic aspects needed to be carefully taken into consideration in the challenging task of devising appropriate processing tools for extracting useful information from big data.

Detecting the underlying low-dimensional space (subspace) where high-dimensional data reside, is at the heart of several signal processing and machine learning tasks, such as network anomalies detection,\cite{mardani_dyn}, image denoising, \cite{elad_image_denoising}, \cite{theodoridis2015machine}, direction of arrival (DOA) estimation, \cite{petrels}, etc. The celebrated PCA indubitably holds a prominent position in the family of this kind of algorithms. However, since PCA is a {\it batch} method  a) its computational complexity scales with the size of the available measurement data and b) it requires the storage of the whole bunch of data in memory. Therefore, its application is becoming practically prohibitive in the big data scenario under study. In light of this, {\it online} subspace estimation (tracking) algorithms, that first came into the scene in the 1970s, \cite{Owsley_78,bunch_78}, have nowadays regain their popularity, \cite{grouse,petrels,Mairal_2010}. These tools build upon the hypothesis that datums are sequentially arriving and thus the unknown subspace is {\it adaptively} estimated each time a new data sample becomes available. Interestingly, this premise, besides reducing the computational complexity, leads to schemes with no need of storing data in memory. Nowadays, in a variety of applications dealing with large scale datasets,  datums to be processed are partly observed i.e., a fraction of them might be missing. Depending on the case, incomplete datasets may result either from applying compressed sensing ideas in an effort to facilitate or account for failures in the data acquisition process, \cite{candes2009exact}, or from the inherent nature of signals met in disparate applications, e.g. collaborative filtering,\cite{su2009survey}, image reconstruction \cite{gull1978image}, etc. Consequently, algorithms that perform  subspace tracking from (possibly highly) incomplete data have flourished notably in the last few years.

Along those lines, the GROUSE algorithm, which brings forth an approach based on stochastic gradient descent on the Grassmanian manifold of subspaces, has been presented in \cite{grouse}. Since stochastic approximation is at the core of GROUSE, its computational complexity classifies it to the low-complexity subspace tracking algorithms, \cite{douk_2008}. Nevertheless, the benefit emanating from the low computational cost is mitigated by the inherent ``barriers'' in the search path, \cite{balzano2014local}, that enforce GROUSE being trapped to local optima. In \cite{petrels}, a second order subspace tracking algorithm, of similar computational complexity to GROUSE, dubbed PETRELS, has been presented. PETRELS is an {\it unconstrained} alternating minimization recursive least squares (RLS)-type algorithm, building upon the seminal PASTd subspace tracking algorithm, \cite{yang_95}, and extending it for handling missing data. Common characteristic of both the aforementioned algorithms is the rather weak assumption that the true rank of the sought subspace is known in advance. This shortcoming, which makes both GROUSE and PETRELS exhibit an unstable behavior in case the assumption does not hold, is addressed in \cite{Mardani2015}, where two different algorithms are described. Therein, an upper bound of the nuclear norm is favorably employed for imposing low-rankness on the unknown subspace matrix, thus robustifying the algorithms in the challenging yet realistic scenario of lacking the knowledge of the subspace rank. In that vein, Algorithm 1 of \cite{Mardani2015} is introduced, deriving from an alternating minimization strategy on an exponentially weighted {\it regularized} cost function. In addition, a more efficient in terms of computational complexity Algorithm 2 is presented, based on a stochastic gradient descend approach. 

In this paper, we deal with the low-rank subspace estimation problem from incomplete data (defined in Section \ref{sec:problem_statement}) in a Bayesian framework and devise a new online variational Bayes subspace learning algorithm, termed OVBSL. A basic feature of the proposed methodology is the adoption of an efficient scheme \cite{tan2009automatic,Babacan2012,bayes_MC_2014} for {\it explicitly} enforcing low-rankness on the subspace matrix, whose columns span the underlying low-dimensional data subspace. To this end a three-level hierarchical Bayesian model is proposed, analytically described in Section \ref{sec:proposed_model}. Interestingly, the adopted strategy, besides making OVBSL robust in the lack of knowledge of the true rank, it  also  favors its disclosure after convergence. An additional feature of OVBSL encapsulated in the espoused Bayesian model is the promotion  of sparsity on the subspace matrix. That said, OVBSL pertains to the sparse subspace learning algorithms such as the streaming sparse PCA method, \cite{yang2015streaming}, and the online sparse dictionary learning algorithm, \cite{Mairal_2010}. Nevertheless, OVBSL deviates from the aforementioned deterministic approaches in the sense that {\it sparsity} is enforced simultaneously  with {\it low-rankness} on the subspace matrix, while additionally incomplete data are appropriately handled. Intractability in deriving the posterior distribution due to our model complexity is herein addressed by the ubiquitous  variational Bayes scheme \cite{vb_tzikas}, \cite{themelis2014} (Section \ref{sec:vb_scheme}). The resulting batch algorithm is suitably adapted to the online processing scenario in Section \ref{sec:online_vb}. According to this online variational Bayes scheme, the subspace matrix is adaptively computed based on time-recursive updates of second order statistics of the latent  (projection of measurement data  on the subspace) and the observed (measurement data) variables, as well as the cross-correlation between observed and latent variables. It should be pointed out that the incorporated online variational Bayes approach departs from the seminal online variational Bayes,\cite{sato2001online}, which has been adopted for attacking the matrix completion problem in \cite{silva2012active}, yet in the context of active learning.

In Section \ref{sec:relation_state_of_the_art}, the relevance of OVBSL to the deterministic PETRELS and Algorithm 1 of \cite{Mardani2015} is brought to light within a {\it maximum a posteriori} (MAP) framework arising from our adopted hierarchical model. It is favorably illuminated that from a deterministic point of view OVBSL departs from the {\it unconstrained} RLS-type PETRELS algorithm and likewise the algorithms of \cite{Mardani2015} can be deemed as an algorithm closely associated with the minimization of an exponentially weighted {\it regularized} least squares cost function. However, the key difference to the algorithms of \cite{Mardani2015} is that instead of utilizing the upper-bound of the nuclear norm introduced in \cite{srebro2005rank}, low-rankness  on the subspace matrix is now aptly provoked by the group-sparsity inducing $\ell_{2}/\ell_1$ norm. As far as the computational complexity of OVBSL is concerned, by virtue of the presumed statistical independence among the elements of the subspace matrix, it is similar to that of GROUSE, PETRELS and the stochastic approximation type Algorithm 2 of \cite{Mardani2015}.

OVBSL shares the compelling characteristic of all Bayesian approaches, that is, it is fully automated. Thus, contrary to its deterministic state-of-the-art rivals, herein no tuning parameters are required. That said, cross-validation which is impractical in online applications, is avoided. Moreover, since all parameters are treated as variables, OVBSL instead of point estimates, provides the sufficient statistics of the probability distributions of all the involved parameters, thus offering more valuable supplementary information, compared to its deterministic counterparts. As demonstrated on simulated and real data experiments in Section \ref{sec:experimental}, it presents superior estimation performance than three of the state-of-the-art algorithms described earlier. To validate this, online matrix completion and (either sparse or non sparse) subspace recovery from missing data are simulated as case studies. Finally, the hyperspectral image reconstruction and the eigenface learning problems are examined, corroborating the effectiveness of OVBSL on real data\footnote{A preliminary version of a part of this work was presented in \cite{giampouras2015online}.}.
\\
\emph{Notation}: Column vectors are represented as boldface lowercase letters, e.g. ${\mathbf x}$, and matrices as boldface uppercase letters, e.g. ${\mathbf X}$, while, unless otherwise explicitly stated, $x_i$ is the $i$th element of ${\mathbf x}$ and $x_{ij}$ the $ij$th element of ${\mathbf X}$ . In particular, small boldface calligraphic letters are used to denote columns of a matrix $\mathbf{X}$ (i.e. $\boldsymbol{\mathit{x}}_i$) and regular boldface letters to denote rows, that is ${\mathbf x}^T_j$ and  $(\cdot)^T$ denotes transposition. Moreover $\mathbf{I}_k$ is the $k \times k$ identity matrix, $\|\!\cdot\!\|_2$ is the standard $\ell_2$-norm, $\|\!\cdot\!\|_F$ stands for the Frobenius norm, $\|\cdot\|_{\ast}$ is the nuclear norm, $\odot$ denotes Hadamard entrywise product,  $\langle \cdot \rangle$ is the expectation operator, $\rm{diag}({\mathbf x})$ denotes a diagonal matrix whose diagonal entries are the elements of ${\mathbf x}$, $\rm{diag}(\mathbf{X})$ is a column vector whose entries are the diagonal elements of the square matrix $\mathbf{X}$, Trace$({\mathbf X})$ is the trace of the square matrix ${\mathbf X}$, $|\mathbf{X}|$ its determinant and $\rm{span}(\mathbf{X})$ is the range (column space) of matrix $\mathbf{X}$. Finally, $\mathcal{N}(\mathbf{x}; \boldsymbol \mu, \boldsymbol \Sigma)$ denotes the Gaussian distribution with mean $\boldsymbol \mu$ and covariance matrix $\boldsymbol \Sigma$. $\mathcal{GIG}(x; p, a, b)$ is the one-dimensional generalized inverse Gaussian distribution defined as
\[ \mathcal{GIG}(x; p, a, b) = \frac{ (a/b)^{p/2} \mathrm{exp}\left[ (p - 1 ) \mathrm{log}x - (a x + \frac{b}{x})/2  \right]  }{2 {\cal K}_{p}( \sqrt{a b}) }, \]
where $x > 0$, $a > 0$, $b > 0$, $p$ is real, and ${\cal K}_{p}(\cdot) $ denotes the modified Bessel function of second kind with $p$ degrees of freedom. The pdf of the Gamma distribution is 
\[ \mathcal{G}(x; \zeta, \tau) = \mathrm{exp}\left[ (\zeta - 1 ) \mathrm{log}x - x\tau - \mathrm{log}\Gamma(\zeta) + \zeta \mathrm{log}\tau  \right]  ,
\] 
where $\Gamma(\cdot)$ is the gamma function, while 
\[
\mathcal{IG}(x; \zeta, \tau) = \mathrm{exp}\left[ -(\zeta + 1 ) \mathrm{log}x - \frac{\tau}{x}- \mathrm{log}\Gamma(\zeta) + \zeta \mathrm{log}\tau     \right] 
\]
is the inverse Gamma distribution.
\section{Problem Statement} \label{sec:problem_statement}
Let $n$ be the time-index and $\mathbf{y}(n)$ a sequence of high-dimensional $K\times 1$ vectors of observations that lie in a linear low-dimensional subspace of rank $r(n)$ with $r(n)\ll K$. Both the linear subspace and its rank may be time-varying. Accordingly, the observations at time $n$ can be expressed as,
\begin{equation}
\mathbf{y}(n) = \mathbf{U}(n)\mathbf{c}(n),
\label{eq:obsrv1}
\end{equation}
where $\mathbf{U}(n)$ is a $K \times r(n)$ matrix whose columns span the underlying data subspace and vector $\mathbf{c}(n)$ contains the coefficients of $\mathbf{y}(n)$ in this subspace. Since, in general, the true rank $r(n)$ of $\mathbf{U}(n)$ is unknown and in order to account for noisy observations, we may assume that our data are produced based on the following linear regression model 
\begin{equation}
\mathbf{y}(n) = \mathbf{W}(n)\mathbf{x}(n)+ \mathbf{e}(n),
\label{eq:obsrv2}
\end{equation}
where $\mathbf{W}(n)$ is a $K \times L$ subspace matrix with $K \gg L \geq r(n)$ and $\mathrm{span}(\mathbf{U}(n)) \subseteq \mathrm{span}(\mathbf{W}(n))$. Moreover, in $(\ref{eq:obsrv2})$, the $L \times 1$  vector $\mathbf{x}(n)$ is the low-dimensional representation of $\mathbf{y}(n)$ in the subspace spanned by the columns of $\mathbf{W}(n)$ and $\mathbf{e}(n)$ is additive Gaussian noise. In other words, besides the noise, a reasonable {\it overestimate} of the true rank of the unknown data subspace is considered in our data generation model. 

To generalize our model, we further assume that a) the unknown subspace matrix $\mathbf{W}(n)$ may be sparse, a condition appearing in several applications and b) part of the entries of $\mathbf{y}(n)$ are missing. The latter means that what we actually have is not $\mathbf{y}(n)$, but $\mathbf{z}(n)$, where 
\begin{align}
\mathbf{z}(n) = \boldsymbol{\phi}(n)\odot \mathbf{y}(n) = \boldsymbol{\Phi}_n\mathbf{y}(n). 
\label{eq:zn}
\end{align}
In (\ref{eq:zn}), $\boldsymbol{\phi}(n)$ is a $\{0,1\}$-binary $K \times 1$ vector having $0$'s at the positions where $\mathbf{y}(n)$ has missing entries and $1$'s elsewhere and $\boldsymbol{\Phi}_n = \mathrm{diag}(\boldsymbol{\phi}(n))$. If we now stack together observation vectors (with possible missing elements) up to time $n$, as {\it rows} in a $n\times K$ matrix $\mathbf{Z}(n)$, yields
\begin{equation}
\mathbf{Z}(n) = \boldsymbol{\Phi}(n) \odot \mathbf{Y}(n) = \boldsymbol{\Phi}(n) \odot \left( \mathbf{X}(n)\mathbf{W}^T(n)+ \mathbf{E}(n) \right),
\label{eq:obsrv3}
\end{equation}
where
\begin{align}
{\mathbf Z}(n) = \left[{\mathbf z}(1),{\mathbf z}(2),\ldots,{\mathbf z}(n)\right]^T 
= [{\boldsymbol {\mathit z}}_1(n),{\boldsymbol {\mathit z}}_2(n),\ldots,{\boldsymbol {\mathit z}}_K(n)],\label{eq:Zn} \\
{\mathbf Y}(n) = \left[{\mathbf y}(1),{\mathbf y}(2),\ldots,{\mathbf y}(n)\right]^T 
= [{\boldsymbol {\mathit y}}_1(n),{\boldsymbol {\mathit y}}_2(n),\ldots,{\boldsymbol {\mathit y}}_K(n)],\label{eq:Yn} \\
\boldsymbol{\Phi}(n) = [{\boldsymbol \phi}(1),{\boldsymbol \phi}(2),\ldots,{\boldsymbol \phi}(n)]^T = [\boldsymbol{\varphi}_1(n),\boldsymbol{\varphi}_2(n),\ldots,\boldsymbol{\varphi}_K(n)], \label{eq:phi} \\
{\mathbf X}(n) = \left[{\mathbf x}(1),{\mathbf x}(2),\ldots,{\mathbf x}(n)\right]^T   
= [{\boldsymbol {\mathit x}}_1(n),{\boldsymbol {\mathit x}}_2(n),\dots,{\boldsymbol {\mathit x}}_L(n)]
\label{eq:Xn}
\end{align}
and $\mathbf{E}(n) = [\mathbf{e}(1),\mathbf{e}(2),\ldots,\mathbf{e}(n)]^T$. In addition, we define the subspace matrix $\mathbf{W}(n)$ row- and columnwise as\footnote{Recall that in (\ref{eq:Zn})-(\ref{eq:Wn}), small boldface calligraphic letters have been used to denote columns of matrices and regular boldface letters to denote rows.} 
\begin{align}
{\mathbf W}(n) = \left[{{\mathbf w}}_1(n),{{\mathbf w}}_2(n),\ldots,{{\mathbf w}}_K(n)\right]^T   
 = \left[ {\boldsymbol {\mathit w}_1}(n), {\boldsymbol {\mathit w}_2}(n),\dots,{\boldsymbol {\mathit w}_L}(n)   \right].
\label{eq:Wn}
\end{align}
It can be noticed from Eqs. (\ref{eq:Zn})-(\ref{eq:Wn}) that the row size of matrices ${\mathbf Z}(n), {\mathbf Y}(n), \boldsymbol{\Phi}(n)$ and ${\mathbf X}(n)$ increases with time, while ${\mathbf W}(n)$ is a time-varying fixed size $K \times L$ matrix. 

The goals of this work are a) the estimation and tracking of the underlying low-dimensional subspace where measurement data reside,  b) the estimation of the low-rank representation of data in this subspace in time and, as a by-product, c) the recovery of the complete measurement data marix $\mathbf{Y}(n)$ via online matrix completion.  In this context, given the batch of incomplete data $\mathbf{Z}(n)$, we aim at estimating the unknown low-rank subspace matrix $\mathbf{W}(n)$ and the latent matrix of projections $\mathbf{X}(n)$ in this subspace. However, in case of streamingly received data, the use of a batch iterative solver entails the processing of the whole bunch of data that are available up to every time instant, rendering the whole procedure computationally prohibitive and thus practically infeasible. A way to alleviate this impediment is by employing online data handling, whereby incomplete observation vectors $\mathbf{z}(n)$ are acquired and processed sequentially to learn and track $\mathbf{W}(n)$ and provide estimates of the vectors of coefficients $\mathbf{x}(n)$. 

In the following, we tackle the aforementioned problem using a Bayesian approach. First, an appropriate Bayesian model is defined that effectively promotes the low-rankness of the sought subspace through column sparsity inducing Laplace priors. As it will become clear below, the adopted modeling aims at revealing the true data subspace (spanned by the columns of $\mathbf{U}(n)$) and its true rank $r(n)$, starting from an overestimate $L$ of it. Based on the proposed Bayesian model, a variational Bayes batch iterative subspace estimation algorithm is developed, which after suitable adjustments leads to an efficient online subspace learning scheme. 
\section{The proposed Bayesian model}\label{sec:proposed_model}
To develop a Bayesian inference method, first a Bayesian model must be defined consisting of a) the likelihood function of the data and b) suitable priors assigned to the parameters of the model. The likelihood function of the observed data depends on the statistical properties of the additive noise, which is commonly taken to be Gaussian with zero mean and constant variance. In this work, in order to place more importance on recent data and downgrade older measurements which is meaningful under time-varying conditions, we employ a so-called forgetting factor $\lambda$ with $0 \ll \lambda<1$ and define the noise distribution as\footnote{From (\ref{eq:Edist}), more recent error vectors have smaller variance compared to older ones, which is equivalent to recent measurements being considered more reliable than older ones.},
\begin{equation}
\mathbf{E}(n) = \prod_{i=1}^{n}\mathcal{N}(\mathbf{e}(i)|\mathbf{0},\beta^{-1}\lambda^{i-n}\mathbf{I}_K),
\label{eq:Edist}
\end{equation}
where $\beta$ is a noise precision parameter, while we define
\begin{align}
 \boldsymbol{\Lambda}(n) = \mathrm{diag}([\lambda^{n-1},\lambda^{n-2},\ldots,\lambda,1]^T).
\end{align}
In the following, whenever not necessary, the time index $n$ is omitted to simplify derivations. The time index is reestablished in Section \ref{sec:online_vb}, where the new online subspace estimation algorithm is presented. In this context,  based on  (\ref{eq:obsrv3}) and the noise distribution given in (\ref{eq:Edist}), the likelihood function of the measurement data is expressed as 
\begin{align}
\label{eq:lklhood}
p( \mathbf{Z}\mid \mathbf{X},{\mathbf W},\beta) = \prod^n_{i=1}p(\mathbf{z}(i) \mid \mathbf{x}(i),\mathbf{W},\beta) = \prod_{i=1}^{n}\prod_{k\in \mathcal{I}_{\boldsymbol{\phi}(i)}}\mathcal{N}( z_k(i)\mid \mathbf{w}_k^T \mathbf{x}(i), \beta^{-1}\lambda^{i-n}),
\end{align}
where $\mathcal{I}_{\boldsymbol{\phi}(i)}$ is the set of indices for which the corresponding entries of $\boldsymbol{\phi}(i)$ are $1$.

Now that the likelihood function has been defined, we proceed by presenting the prior distributions imposed on the subspace matrix $\mathbf{W}$ and the coefficients matrix $\mathbf{X}$. These priors aim at simultaneously decreasing the rank and imposing sparsity on the unknown subspace matrix $\mathbf{W}$. Recall that the matrix product $\mathbf{X}\mathbf{W}^T$ in (\ref{eq:obsrv3}) is equivalently written as the sum of the outer products between the columns of $\mathbf{X}$ and $\mathbf{W}$ i.e.,
\begin{align}
 \mathbf{X}\mathbf{W}^T = \sum^L_{l=1}{\boldsymbol {\mathit x}}_l{\boldsymbol {\mathit w}^T_l}.
 \label{eq:outer_product}
\end{align}
 From (\ref{eq:outer_product}) it is readily seen that the rank of the matrix $\mathbf{X}\mathbf{W}^T$ equals to the number $L$ of the rank-one terms existing into this summation. Hence, a natural approach to reduce the rank $L$ of the sought subspace is to somehow eliminate some of the rank-one contributing terms in (\ref{eq:outer_product}). A relevant scheme, \cite{Babacan2012}, reduces the rank by imposing column sparsity {\it jointly} on $\mathbf{X}$ and $\mathbf{W}$. Herein, as in \cite{Babacan2012}, this sparsity constraint is integrated in the modeling of the prior distributions of ${\boldsymbol {\mathit x}}_l$ and ${\boldsymbol {\mathit w}_l}$, as explained below. At the same time, as stated earlier, in several applications (e.g. \cite{cai2007spectral,zhang2011large}) the subspace matrix $\mathbf{W}$ is required to be sparse. That said, joint sparsity on ${\boldsymbol {\mathit x}}_l$ and ${\boldsymbol {\mathit w}_l}$ and the sparse structure on subspace matrix $\mathbf{W}$ are simultaneously incorporated in the modeling process of the corresponding prior distributions. In light of this, three-level hierarchical priors\footnote{Hierarchical priors are required in order to ensure {\it conjugacy} with respect to the likelihood as well as among them, which is a prerequisite for deriving a tractable posterior inference procedure, \cite{theodoridis2015machine,themelis2014}.} are assigned to the columns of $\mathbf{X}$ and $\mathbf{W}$. At the first level of hierarchy the following Gaussian priors are defined, 
\begin{align}
p(\mathbf{X}\mid \mathbf{s},\beta) = \prod_{l=1}^{L}\mathcal{N}(\boldsymbol {\mathit x}_l \mid \mathbf{0}, \beta^{-1} s^{-1}_l\boldsymbol{\Lambda}^{-1}) \label{prior:X},  \\
p(\mathbf{W}\mid \mathbf{s},\boldsymbol \Gamma,\beta) = \prod_{l=1}^{L}\mathcal{N}( {\boldsymbol {\mathit w}_l} \mid \mathbf{0},\beta^{-1}s^{-1}_l \boldsymbol\Gamma^{-1}_l	) \label{prior:W},
\end{align}
where $\mathbf{s}=[s_1,s_2,\dots,s_L]^T$, $\mathbf{\Gamma} = [\boldsymbol{\gamma}_1,\boldsymbol{\gamma}_2,\dots,\boldsymbol{\gamma}_L]$, $\boldsymbol{\gamma}_l = [\gamma_{1l},\gamma_{2l},\ldots,\gamma_{Kl}]^T$ and $\boldsymbol \Gamma_l = \mathrm{diag}(\boldsymbol{\gamma}_l)$ for $l=1,2,\ldots,L$. Using Bayesian terminology, in our problem the entries of $\mathbf{W}$ correspond to the {\it hidden} variables, while the rows of $\mathbf{X}$ are the so-called {\it latent} variables, since each observation vector $\mathbf{z}(i)$ is associated with a latent vector $\mathbf{x}(i)$, \cite{theodoridis2015machine}. Therefore, similar to the observations matrix $\mathbf{Z}$, the latent coefficient vectors in $\mathbf{X}$ should be also exponentially weighted, as is done in (\ref{prior:X}) by incorporating $\boldsymbol{\Lambda}^{-1}$ in the covariance matrix of $\boldsymbol{\mathit{x}}_l$'s. The prior distribution of $\mathbf{X}$ in (\ref{prior:X}) can be written in an equivalent form with respect to the rows of $\mathbf{X}$ as follows
\begin{align}
p(\mathbf{X}\mid \mathbf{s},\beta) = \prod_{i=1}^{n}\mathcal{N}(\mathbf{x}(i) \mid \mathbf{0}, \beta^{-1} \lambda^{i-n}\mathbf{S}^{-1}), \label{alt_prior:X}
\end{align}
where $\mathbf{S} = \mathrm{diag}(\mathbf{s})$. Note that it is the form of the prior in (\ref{alt_prior:X}) that is mainly used in the analysis of the next sections, although ({\ref{prior:X}) serves in this section to show how the rank is reduced by the proposed model. More specifically, it can be observed from ({\ref{prior:X}) and ({\ref{prior:W}) that the $l$th columns of $\mathbf{X}$ and $\mathbf{W}$ share the same {\em joint sparsity promoting parameters} $s_l$'s. At the same time, the diagonal matrix $\boldsymbol\Gamma_l$ which arises in the prior distribution of $\mathbf{W}$ is responsible for {\em independently} imposing sparsity on the entries of the $l$th column of the subspace matrix\footnote{In case $\mathbf{W}$ is not sparse, we set $\boldsymbol{\Gamma}_l = \mathbf{I}_K$ in (\ref{prior:W}) and no prior applies to $\boldsymbol{\Gamma}$, i.e. Eqs (\ref{prior:gamma}) and (\ref{prior:rho}) are needless.}. In particular, some of the $s_l$'s take large values when Bayesian inference is performed and as a result, both the $l$th columns of $\mathbf{X}$ and $\mathbf{W}$ are driven to zero. Notably, in cases where a parameter $s_l$ does not enforce joint sparsity, the $k$th element of the $l$th column of $\mathbf{W}$ may be independently led to zero by the corresponding {\it subspace sparsity promoting} parameter $\gamma_{kl}$ of $\boldsymbol \Gamma_l$. Based on these premises, at the second level of the hierarchy we define the following conjugate inverse Gamma distributions for $\mathbf{s}$ and $\boldsymbol{\Gamma}$,
\begin{align}
p(\mathbf{s} \mid \boldsymbol{\delta}) = \prod^L_{l=1}\mathcal{IG}(s_l \mid \frac{K+n+1}{2},\frac{\delta_l}{2}), \label{prior:s} \\
p(\mathbf{\Gamma} \mid \boldsymbol{\cal P}) = \prod^K_{k=1}\prod^L_{l=1}\mathcal{IG}(\gamma_{kl}\mid1,\frac{\rho_{kl}}{2}). \label{prior:gamma}
\end{align} 
where $\boldsymbol{\delta} = [\delta_1,\delta_2,\ldots,\delta_L]^T$ and $\boldsymbol{\cal P}$ is the $K \times L$ matrix whose entries are the  $\rho_{kl}$'s.  Finally, at the third level of the hierarchy, conjugate Gamma distributions are defined for the scale parameters $\delta_l$'s and $\rho_{kl}$'s, i.e. 
\begin{align}
p(\delta_l) = \mathcal{G}(\delta_l;\mu,\nu) \\
p(\rho_{kl}) = \mathcal{G}(\rho_{kl};\psi,\xi). \label{prior:rho}
\end{align}
By integrating out $\mathbf{s}$ from (\ref{prior:X}) and (\ref{prior:W}) using (\ref{prior:s}) with $\boldsymbol{\Gamma}$ kept fixed, we are led to a heavy-tailed multiparameter Laplace-type distribution for the joint prior of $\mathbf{X}$ and $\mathbf{W}$ that promotes joint column sparsity, as is shown in Appendix \ref{sec:apendix}. Similarly, by fixing $\mathbf{s}$, from (\ref{prior:W}) and (\ref{prior:gamma}) we get a multiparameter Laplace prior that imposes sparsity on $\mathbf{W}$. 

 The proposed Bayesian model is concluded by assigning a conjugate to the likelihood Gamma prior to  the precision of the noise $\beta$ as follows,
\begin{align}
p(\beta)=\mathcal{G}(\beta;\kappa,\theta)  \label{prior:beta}.
\end{align} 
In the next section, based on the multi-hierarchical model described in this section and presented graphically in Fig. \ref{fig:bmodel}, an approximate Bayesian inference scheme is derived for low-rank sparse subspace learning from partial observations.
\begin{figure}
\centering
 \includegraphics[width=0.86\textwidth]{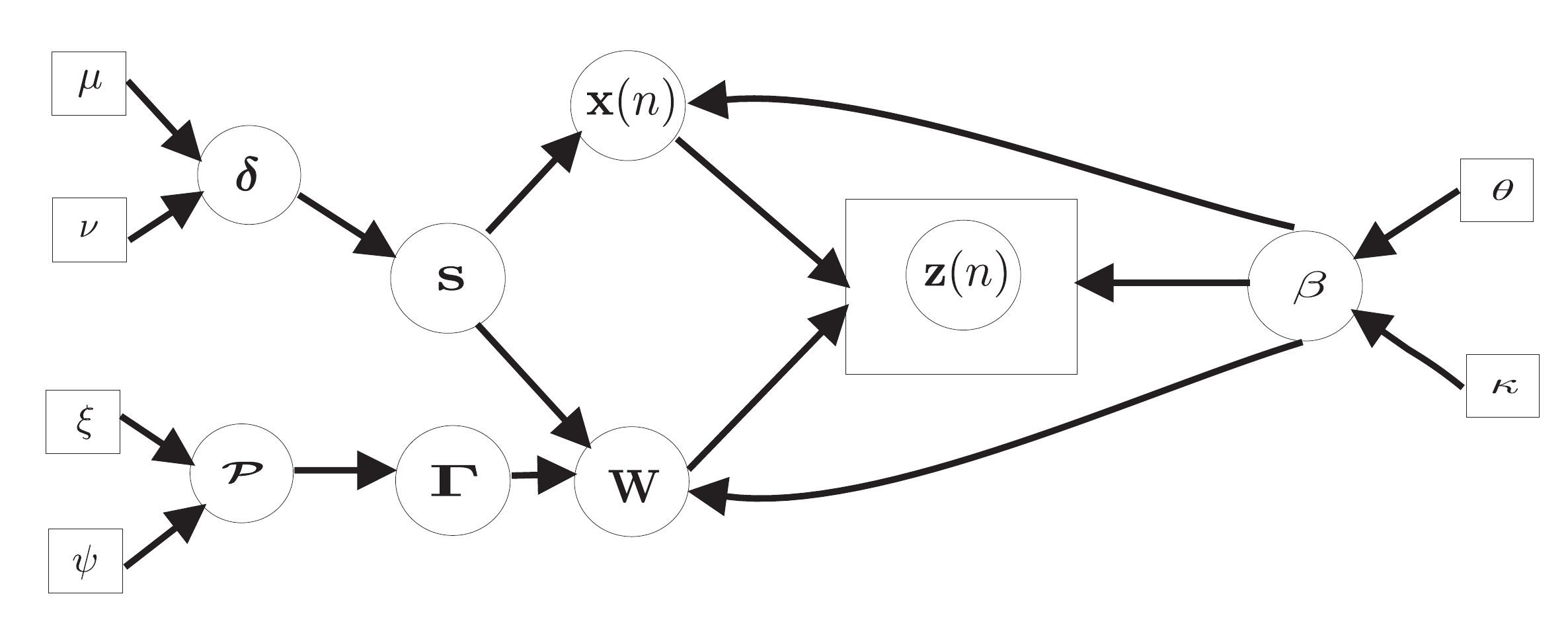}
 \caption{Directed acyclic graph of the proposed Bayesian model.}
\label{fig:bmodel}
\end{figure}

\section{Batch Variational Bayes Inference}\label{sec:vb_scheme}
Inferring the joint posterior distribution of multiple variables given the data boils down to an intractable process when it comes to composite Bayesian models, such as those springing from hierarchical dependences of the involved variables, which are modeled by suitable priors. This is also the case for the Bayesian model described in the previous section, and graphically depicted in Fig.\ref{fig:bmodel}. Following the Bayes' theorem, the exact joint posterior of our variables given the observations is obtained by 
\begin{align}
 p(\mathbf{X},\mathbf{W},\mathbf{s},\mathbf{\Gamma},\boldsymbol{\delta},\boldsymbol{\cal P},\beta\mid \mathbf{Z}) =  \frac{p(\mathbf{Z},\mathbf{X},\mathbf{W},\mathbf{s},\boldsymbol{\Gamma},\boldsymbol{\delta},\boldsymbol{\cal P},\beta)}{\int p(\mathbf{Z},\mathbf{X},\mathbf{W},\mathbf{s},\boldsymbol{\Gamma},\boldsymbol{\delta},\boldsymbol{\cal P},\beta)d\mathbf{X}d\mathbf{W}d\mathbf{s}d\boldsymbol{\Gamma}d\boldsymbol{\delta}d\boldsymbol{\cal P}d\beta}.
 \label{exact_posterior}
\end{align}
Apparently, getting a closed form expression for the posterior given in (\ref{exact_posterior}) involves the daunting task of estimating the integral at the denominator.
To obviate obstacles of this type, plentiful approximate inference schemes have come to light in literature, \cite{Minka_exp_prop,Cemgil2012}. Herein, the ubiquitous variational Bayes inference approach is adopted, \cite{vb_tzikas}. The basic premise of this approach inspired from the field of statistical physics is the assumption that the posterior distribution can be approximately expressed in a factorized form. Based on this particular hypothesis, the exact joint posterior $p(\mathbf{X},\mathbf{W},\mathbf{s},\mathbf{\Gamma},\boldsymbol{\delta},\boldsymbol{\cal P},\beta\mid \mathbf{Z})$ is approximated by $q(\mathbf{X},\mathbf{W},\mathbf{s},\mathbf{\Gamma},\boldsymbol{\delta},\boldsymbol{\cal P},\beta)$, defined as
\begin{align}
 q(\mathbf{X},\mathbf{W},\mathbf{s},\mathbf{\Gamma},\boldsymbol{\delta},\boldsymbol{\cal P},\beta) = q(\beta)\prod^n_{i=1}{q(\mathbf{x}(i))}\prod^K_{k=1}\prod^L_{l=1}q(w_{kl})\prod^L_{l=1}q(s_l)\prod^L_{l=1}q(\delta_l)\prod^K_{k=1}\prod^L_{l=1}q(\gamma_{kl})q(\rho_{kl}).
 \label{approx_posterior}
\end{align}
From (\ref{approx_posterior}) it is easily noticed that there has been considered full statistical {\it a posteriori} independence among the rows of $\mathbf{X}$, as well as among all the elements of the subspace matrix $\mathbf{W}$. As far as $\mathbf{x}(i)$'s are concerned, being statistical independent is something that is naturally brought up due to the presumed independence among the corresponding observation vectors $\mathbf{z}(i)$'s. On the other hand, the posterior independence imposed on the entries of $\mathbf{W}$ gives rise to  coordinate-descend recursions for retrieving $w_{nk}$'s. In doing so, as shown later, the computational complexity required to recover $\mathbf{W}$ is significantly reduced. Notably, as implied by (\ref{approx_posterior}), those explicit assumptions on the independence among the rows of $\mathbf{X}$ and the elements of $\mathbf{W}$ dictate relevant statistical independence on the variables of our model belonging to the second and the third level of hierarchy, namely $\mathbf{s}, \boldsymbol{\delta}, \mathbf{\Gamma}$ and $\boldsymbol{\cal P}$.

In an attempt to bring to light the particular way that the posterior distributions $q(\cdot)$'s of all variables in (\ref{approx_posterior}) are recovered according to the variational Bayes scheme, we define the cell array $\boldsymbol{\theta}=\{\mathbf{x}(1),\dots,\mathbf{x}(n),w_{11},\dots,w_{nK},s_1, \ldots,s_L, \gamma_{11},\ldots,\gamma_{KL},\delta_{1},\ldots,\delta_L,p_{11},\ldots,p_{KL}\}$\footnote{Note that for notational convenience, the entries of $\boldsymbol{\theta}$ i.e. the $\boldsymbol{\theta}_i$'s may represent either vectors or scalars.}. The posterior distribution $q(\boldsymbol{\theta}_i)$ of each component $\boldsymbol{\theta}_i$ is then obtained by minimizing the Kullback-Leibler distance between the posterior i.e. $p(\boldsymbol{\theta} \mid \mathbf{Z})$, and the approximate one $q(\boldsymbol{\theta})$ leading to the following closed-form expressions \cite{vb_tzikas}
\begin{align}
q(\boldsymbol{\theta}_i) = \frac{\exp\left(\langle\mathrm{ln}p(\mathbf{Z},\boldsymbol{\theta})\rangle_{i\neq j} \right)}{\int \exp\left(\langle\mathrm{ln}p(\mathbf{Z},\boldsymbol{\theta})\rangle_{i\neq j} \right)d\boldsymbol{\theta}_i}. \label{eq:qtheta}
\end{align}
In the last equation $\langle \cdot \rangle_{i\neq j}$ denotes expectation taken with respect to all $q(\boldsymbol{\theta}_j)$'s but $q(\boldsymbol{\theta}_i)$. Interestingly, through (\ref{eq:qtheta}) the parameters of each posterior $q(\boldsymbol{\theta}_i)$ are expressed in terms of the parameters of the other distributions $q(\boldsymbol{\theta}_j)$'s, for $j \neq i$. Thus, the minimization of the Kullback-Leibler distance gives birth to a cyclic iterative scheme, whereby the parameters of each $q(\boldsymbol{\theta}_i)$ are computed based on the most recent estimates of the parameters of the rest $q(\boldsymbol{\theta}_j)$'s, as it will also become more clear below.

Along those lines, the procedure described earlier is now applied for our three-level hierarchical Bayesian model. In view of this, due to the aforementioned conjugacy of the respective prior distributions, the posterior distribution $q(\mathbf{x}(i))$ of the $i$th coefficient vector does turn out to be Gaussian, i.e.
\begin{align}
q(\mathbf{x}(i))  = \mathcal{N}\left(\mathbf{x}(i)\mid \langle\mathbf{x}(i)\rangle, \boldsymbol{\Sigma}_{\mathbf{x}(i)}\right), 
\end{align}
with mean $\langle\mathbf{x}(i)\rangle$ and covariance matrix $\boldsymbol{\Sigma}_{\mathbf{x}(i)}$ given by,  
\begin{align}
\langle\mathbf{x}(i)\rangle &= \langle \beta \rangle \boldsymbol{\Sigma}_{\mathbf{x}(i)}\langle\mathbf{W}\rangle^T \mathbf{z}(i), \label{eq:xmean} \\
\boldsymbol{\Sigma}_{\mathbf{x}(i)} &= \langle \beta \rangle^{-1} \left(\langle \mathbf{W}^T\boldsymbol{\Phi}_i\mathbf{W}\rangle + \langle\mathbf{S}\rangle\right)^{-1}, \label{eq:xcov}
\end{align}
where we recall that $\boldsymbol{\Phi}_i = \mathrm{diag}(\boldsymbol{\phi}(i))$. The expectation term $\langle \mathbf{W}^T\boldsymbol{\Phi}_i\mathbf{W}\rangle$ is expressed as, 
\begin{align}
\langle \mathbf{W}^T\boldsymbol{\Phi}_i\mathbf{W}\rangle = \langle\mathbf{W}\rangle^T\boldsymbol{\Phi}_i\langle\mathbf{W}\rangle+ \sum^K_{k=1}\phi_{ik}\boldsymbol{\Sigma}_{\mathbf{w}_k} \label{eq:wphiw}
\end{align}
where $\boldsymbol{\Sigma}_{\mathbf{w}_k} = \mathrm{diag}([\sigma_{w_{k1}}^2,\sigma_{w_{k2}}^2,\ldots,\sigma_{w_{kL}}^2 ]^T)$ by virtue of the statistical independence assumed for the elements of $\mathbf{W}$. Note that $\sigma_{w_{kl}}^2$ is the variance of $w_{kl}$ whose posterior turns out also to be Gaussian, i.e.
\begin{align}
q(w_{kl}) &= \mathcal{N}(w_{kl}\mid \langle w_{kl}\rangle,\sigma^2_{{w}_{kl}}),  
\end{align}
with 
\begin{align}
\langle w_{kl} \rangle = \langle\beta\rangle \sigma^2_{w_{kl}}\left(\langle\boldsymbol{\mathit{x}}_l\rangle^T\boldsymbol{\Lambda} \boldsymbol{\mathit{z}}_k-\langle \boldsymbol{\mathit{x}}_l^T \boldsymbol{\Lambda}\boldsymbol{\varPhi}_k\mathbf{X}_{\neg l}\rangle\langle\mathbf{w}_{k\neg l}  \rangle   \right) \label{posterior_W_mean},\\
\sigma^2_{{w}_{kl}} = \langle\beta\rangle^{-1}\left(\langle\boldsymbol{\mathit{x}}^T_l\boldsymbol{\Lambda}\boldsymbol{\varPhi}_k\boldsymbol{\mathit{x}}_l\rangle + \langle \gamma_{kl}\rangle \langle s_l\rangle  \right)^{-1} \label{posterior_W_var}.
\end{align}
$\mathbf{X}_{\neg l}$ and $\mathbf{w}_{k\neg l}$ in (\ref{posterior_W_mean}) are the quantities arising after removing the $l$th column and the $l$th element of $\mathbf{X}$ and $\mathbf{w}_k$, respectively and $\boldsymbol{\varPhi}_k  = \mathrm{diag}(\boldsymbol{\varphi}_k)$. As for the expectation terms appearing in (\ref{posterior_W_mean}) and (\ref{posterior_W_var}), it holds,
\begin{align}
 \langle \boldsymbol{\mathit{x}}_l^T \boldsymbol{\Lambda}\boldsymbol{\varPhi}_k\mathbf{X}_{\neg l}\rangle & = \langle \boldsymbol{\mathit x}_l \rangle^T\boldsymbol{\Lambda}\boldsymbol{\varPhi}_k\langle\mathbf{X}_{\neg l}\rangle + \sum^n_{i=1}\lambda^{n-i}\phi_{ik}\boldsymbol{\sigma}_{{x(i)\neg l}}^T, \label{eq:expect1}\\
 \langle\boldsymbol{\mathit{x}}^T_l\boldsymbol{\Lambda}\boldsymbol{\varPhi}_k\boldsymbol{\mathit{x}}_l\rangle & = \langle\boldsymbol{\mathit{x}}_l\rangle^T \boldsymbol{\Lambda}\boldsymbol{\varPhi}_k \langle\boldsymbol{\mathit{x}}_l\rangle+ \sum^n_{i=1}\lambda^{n-i}\phi_{ik}\sigma_{{x}_{il}}, \label{eq:expect2}
\end{align}
with $\boldsymbol{\sigma}_{{x(i)\neg l}}$ standing for the $l$th column of $\boldsymbol{\Sigma}_{\mathbf{x}(i)}$ after removing its $l$th element $\sigma_{x_{il}}$. 

Next, the posterior distributions of the variables $s_l$'s and $\gamma_{kl}$'s  belonging to the second hierarchical level are unfolded. From (\ref{eq:qtheta}) it can be shown that the column sparsity promoting parameters $s_l$'s are {\it a posteriori} distributed according to the following generalized inverse Gaussian distribution,
\begin{align}
q(s_l) = \mathcal{GIG}\left(s_l\mid -\frac{1}{2},\langle \beta \rangle\left(\langle \boldsymbol{\mathit{w}}^T_l\boldsymbol{\Gamma}_l\boldsymbol{\mathit{w}}_l\rangle + \langle \boldsymbol{\mathit{x}}^T_l\boldsymbol{\Lambda}\boldsymbol{\mathit{x}}_l \rangle  \right), \langle \delta_l \rangle \right). \label{eq:qsl}
\end{align}
For the mean $\langle s_l \rangle$ of the $\mathcal{GIG}$ distribution it holds,
\begin{align}
 \langle  s_l \rangle = \sqrt{\frac{\langle\delta_l\rangle}{\langle \beta \rangle\left(\langle \boldsymbol{\mathit{w}}^T_l\boldsymbol{\Gamma}_l\boldsymbol{\mathit{w}}_l\rangle + \langle \boldsymbol{\mathit{x}}^T_l\boldsymbol{\Lambda}\boldsymbol{\mathit{x}}_l \rangle  \right)}}. \label{eq:sl}
\end{align}
Likewise, the posterior distribution of  $\gamma_{kl}$'s that promote independently sparsity on the elements of the subspace matrix $\mathbf{W}$ is the generalized inverse Gaussian 
\begin{align}
q(\gamma_{kl}) = \mathcal{GIG}\left(\gamma_{kl}\mid-\frac{1}{2},\langle\beta\rangle\langle s_l \rangle\langle w^2_{kl}\rangle,\langle\rho_{kl}\rangle\right), \label{eq:qgammakl}
\end{align}
with $\langle w^2_{kl}\rangle = \langle w_{kl} \rangle^2 + \sigma^2_{w_{kl}} $. Hence,
\begin{align}
 \langle \gamma_{kl} \rangle = \sqrt{\frac{\langle\rho_{kl}\rangle}{\langle\beta\rangle\langle s_l \rangle(\langle w_{kl} \rangle^2 + \sigma^2_{w_{kl}})}}.\label{eq:gkl}
\end{align}

As far as the hyperparameters $\delta_l$ and $\rho_{kl}$ of $s_l$ and $\gamma_{kl}$ respectively, are concerned, both are {\it a posteriori} Gamma distributed i.e.,
\begin{align}
q(\delta_l) = \mathcal{G}\left(\delta_l \mid \bar{\mu},\bar{\nu}_l  \right) 
\end{align}
with $\bar{\mu} = \mu + \frac{n+K+1}{2}$ and $\bar{\nu}_l = \nu + \frac{1}{2}\langle \frac{1}{s_l} \rangle $, and 
\begin{align}
q(\rho_{kl}) = \mathcal{G}\left(\rho_{kl}\mid 	\bar{\psi},\bar{\xi}_{kl}  \right) 
\end{align}
with $\bar{\psi} = \psi + 1$ and $\bar{\xi}_{kl}= \xi + \frac{1}{2}\langle \frac{1}{\gamma_{kl}} \rangle $. For the expected values of $\delta_l$ and $\rho_{kl}$, that is $\langle \delta_l \rangle$ and $\langle \rho_{kl} \rangle$ we have,
\begin{align}
 \langle \delta_l \rangle = \frac{\mu + \frac{n+K+1}{2}}{\nu + \frac{1}{2}\langle \frac{1}{s_l} \rangle}   \label{post:exp_delta},\\
 \langle \rho_{kl} \rangle = \frac{\psi + 1}{\xi + \frac{1}{2}\langle \frac{1}{\gamma_{kl}} \rangle } \label{post:exp_rho}.
\end{align}
Using the form of the distributions in (\ref{eq:qsl}) and (\ref{eq:qgammakl}), the expectation terms $\langle \frac{1}{s_l} \rangle$ and $\langle \frac{1}{\gamma_{kl}} \rangle$ arising in (\ref{post:exp_delta}) and (\ref{post:exp_rho}) can be obtained as, 
\begin{align}
 \left< \frac{1}{s_l} \right> & = \frac{1}{\langle s_l \rangle} + \frac{1}{\langle \delta_l \rangle} \label{eq:invsl} \\ 
 \left< \frac{1}{\gamma_{kl}} \right> & = \frac{1}{\langle \gamma_{kl} \rangle} + \frac{1}{\langle \rho_{kl} \rangle} \label{eq:gammakl}
\end{align}

Concluding the posterior distributions of all the involved variables in our hierarchical model, it can be shown that the noise precision $\beta$ is Gamma distributed as follows, 
\begin{align}
q(\beta) & = \mathcal{G}\left(\beta \mid \bar{\kappa},\bar{\theta} \right) \label{eq:postbeta}
\end{align}
where,
\begin{align}
\bar{\kappa} = \kappa + \frac{n\left(K+L\right)+KL}{2}
\end{align}
\begin{align}
\bar{\theta} = \theta + \sum_{k=1}^K\left(\langle\|\boldsymbol{\Lambda}^{\frac{1}{2}}\left(\boldsymbol{\mathit{z}}_k -\boldsymbol{\varPhi}_k \mathbf{X}\mathbf{w}_k  \right) \|^2_2\rangle  + \langle \mathbf{w}^T_k\mathbf{S}\boldsymbol{ \varGamma}_k\mathbf{w}_k \rangle \right) + \sum^L_{l=1}\langle s_l \rangle\langle\boldsymbol{\mathit{x}}^T_l \boldsymbol{\Lambda}\boldsymbol{\mathit{x}}_l  \rangle.	\label{beta_theta_posterior}
\end{align}
and $\boldsymbol{\varGamma}_k = \mathrm{diag}([\gamma_{k1},\gamma_{k2},\ldots,\gamma_{kL}]^T)$. The expectation of $\beta$ is given by $\langle \beta \rangle = \frac{\bar{\kappa}}{\bar{\theta}}$. 
As for the expectation terms arising in (\ref{beta_theta_posterior}), it holds,
\begin{align}
\langle\|\boldsymbol{\Lambda}^{\frac{1}{2}}\left(\boldsymbol{\mathit{z}}_k - \boldsymbol{\varPhi}_k\mathbf{X}\mathbf{w}_k  \right) \|^2_2\rangle  =  \|\boldsymbol{\Lambda}^{\frac{1}{2}}\left(\boldsymbol{\mathit{z}}_k -\boldsymbol{\varPhi}_k \langle\mathbf{X}\rangle\langle\mathbf{w}_k\rangle \right) \|^2_2 + \mathrm{Tr}\left(\langle\mathbf{X}\rangle^T\boldsymbol{\Lambda}\boldsymbol{\varPhi}_k\langle\mathbf{X}\rangle\boldsymbol{\Sigma}_{\mathbf{w}_k}  \right) \nonumber\\ + \langle \mathbf{w}_k\rangle^T \sum^n_{i=1}\phi_{ik}\lambda^{n-i}\boldsymbol{\Sigma}_{\mathbf{x}(i)}\langle\mathbf{w}_k \rangle + \mathrm{Tr}\left(\boldsymbol{\Sigma}_{\mathbf{w}_k }\sum^n_{i=1}\phi_{ik}\lambda^{n-i}\boldsymbol{\Sigma}_{\mathbf{x}(i)}\right) 
\end{align}
\begin{align}
\langle\mathbf{w}^T_k \mathbf{S}\boldsymbol{\Gamma}_l\mathbf{w}_k \rangle = \langle\mathbf{w}_k\rangle^T\langle \mathbf{S}\rangle\langle\boldsymbol{\Gamma}_l\rangle\langle\mathbf{w}_k \rangle + \sum^L_{l=1} \langle s_l \rangle \langle\gamma_{kl}\rangle\sigma^2_{w_{kl}} \label{eq:lastbeta}
\end{align}
The mutual dependence among the moments of all the model parameters, that can be easily observed in the previous expressions, paves the way for an iterative scheme over the involved quantities. It should be emphasized though that since we aim at handling a massive amount of streaming data, the utilization of those expressions ends up to be a prohibitive task. More specifically, as the number $n$ of the observations  increases, calculations that involve  quantities such as $\mathbf{Z},\mathbf{X}$, become increasingly demanding in terms of the memory storage as well as the computational effort needed. In light of this, an online scheme is unraveled in the next section, that favorably adjusts the above defined expressions to the streaming processing scenario. 

\section{Online Variational Bayes subspace estimation}\label{sec:online_vb}
In this section, based on the expressions analytically described earlier for the batch problem, we derive an online variational Bayes sparse subspace estimation algorithm. According to the online scenario, incomplete high dimensional datums $\mathbf{z}(n)$'s are streamingly received at each time instance $n$. Then, the proposed algorithm proceeds by a) computing an estimate $\hat{\mathbf{x}}(n)$ of the coefficients vector of the observations on the subspace acquired in the previous iteration (i.e. $\hat{\mathbf{W}}(n-1)$) and next b) updating {\it elementwise} the subspace matrix $\hat{\mathbf{W}}(n-1)$ to $\hat{\mathbf{W}}(n)$. In the sequel, for notational convenience, we disregard the expectation operator $\langle\cdot \rangle$. Then, with a slight but straightforward abuse of notation and by handling the time index appropriately, we get form (\ref{eq:xmean}), (\ref{eq:xcov}) and (\ref{eq:wphiw}),  
\begin{align}
\hat{\mathbf{x}}(n) = \beta(n-1)\boldsymbol{\Sigma}_{\hat{\mathbf{x}}}(n)\hat{\mathbf{W}}^T(n-1)\mathbf{z}(n), 
\end{align}
\begin{align}
\boldsymbol{\Sigma}_{\hat{\mathbf{x}}}(n) = \beta^{-1}(n-1)\left(\hat{\mathbf{W}}^T(n-1)\boldsymbol{\Phi}_n\hat{\mathbf{W}}(n-1) + \sum^K_{k=1}\phi_{k}(n)\boldsymbol{\Sigma}_{\hat{\mathbf{w}}_k}(n-1)+ \mathbf{S}(n-1) \right)^{-1} \label{post:update_SigmaX}.
\end{align}
Next, we define the following {\it fixed-size with respect to time} quantities,
\begin{eqnarray}
\mathbf{T}(n) = \hat{\mathbf{X}}^T(n)\boldsymbol{\Lambda}(n)\mathbf{Z}(n), \label{eq:Tn}
\end{eqnarray}
\begin{eqnarray}
\mathbf{Q}(n) = \hat{\mathbf{X}}^T(n)\boldsymbol{\Lambda}(n)\hat{\mathbf{X}}(n) + \sum^n_{i=1}\lambda^{n-i}\boldsymbol{\Sigma}_{\hat{\mathbf{x}}}(i), \label{eq:Qn}
\end{eqnarray}
and for $k=1,2,\ldots,K$,
\begin{eqnarray} 
\mathbf{P}_{k}(n) = \hat{\mathbf{X}}^T(n)\boldsymbol{\Lambda}(n)\boldsymbol{\varPhi}_k(n)\hat{\mathbf{X}}(n) + \sum^n_{i=1}\lambda^{n-i}\phi_{k}(i)\boldsymbol{\Sigma}_{\hat{\mathbf{x}}}(i), \label{autocorr_matrix}
\end{eqnarray}
\begin{eqnarray}
d_k(n) = \boldsymbol{\mathit{z}}^T_k(n)\boldsymbol{\Lambda}(n)\boldsymbol{\mathit{z}}_k(n). \label{energy_yk}
\end{eqnarray}
The basic idea in any online scheme is the formulation of the various quantities that carry the past knowledge of the relevant process in a time-recursive manner. Interestingly, eqs. (\ref{eq:Tn})-(\ref{energy_yk}) can easily be written in time-recursive forms i.e., 
\begin{align}
\mathbf{T}(n) = \lambda \mathbf{T}(n-1) + \hat{\mathbf{x}}(n)\mathbf{z}^T(n), \label{eq:upTn}
\end{align}
\begin{align}
\mathbf{Q}(n) = \lambda \mathbf{Q}(n-1) + \boldsymbol{\Sigma}_{\hat{\mathbf{x}}}(n) + \hat{\mathbf{x}}(n) \hat{\mathbf{x}}^T(n), \label{eq:upQn}
\end{align}
\begin{align}
 \mathbf{P}_k(n) =\lambda \mathbf{P}_k(n-1) + \phi_k(n)\left(\boldsymbol{\Sigma}_{\hat{\mathbf{x}}}(n) + \hat{\mathbf{x}}(n)\hat{\mathbf{x}}^T(n)\right), \label{eq:upPkn}
\end{align}
\begin{align}
 d_k(n) = \lambda d_k(n-1) + z^2_k(n). \label{eq:updkn}
\end{align}
Moreover, for $k=1,2,\ldots,K$, we define the following matrices that stem from $\mathbf{P}_k(n)$'s with the addition of appropriate diagonal terms,
\begin{eqnarray}
\mathbf{R}_k(n) = \mathbf{P}_k(n) + \boldsymbol{\varGamma}_k (n-1) \mathbf{S}(n-1) \label{eq:upRkn}.
\end{eqnarray}
Having aptly obtained the above computationally efficient formulas, we can now head for online processing. Towards this, the equations derived for the batch case are suitably modified by incorporating the previously  defined recursively computed quantities. More specifically, by substituting (\ref{eq:expect1}), (\ref{eq:expect2}) in (\ref{posterior_W_mean}), (\ref{posterior_W_var}) respectively and using (\ref{eq:Tn}), (\ref{autocorr_matrix}) and (\ref{eq:upRkn}) we get the following time update expressions for the entries of the 
subspace matrix estimate $\hat{\mathbf{W}}$ at time $n$, 
\begin{align}
 \hat{w}_{kl}(n) = \beta(n-1)\sigma^2_{\hat{{w}}_{kl}}(n-1)\left(t_{lk}(n) - \mathbf{r}^T_{k\neg l}(n)\hat{\mathbf{w}}_{k\neg l}(n)\right), \label{update_w} 
\end{align}
\begin{align} 
\sigma^2_{\hat{w}_{kl}}(n) = \beta^{-1}(n-1)r^{-1}_{k,ll}(n),  \label{sigma_w_update}
\end{align}
where $t_{lk}(n)$ is the $lk$th entry of the $L \times K$ matrix $\mathbf{T}(n)$, $\mathbf{r}^T_{k\neg l}(n)$ is the $l$th row of $L \times L$ autocorrelation matrix $\mathbf{R}_k(n)$ after neglecting its $l$th element i.e. $r_{k,ll}$ and finally
\begin{align}
\hat{\mathbf{w}}_{k\neg l}(n) =
[\hat{w}_{k1}(n),\hat{w}_{k2}(n),\dots,\hat{w}_{kl-1}(n),\hat{w}_{kl+1}(n-1),\dots,\hat{w}_{kL}(n-1)]^T. \label{w_vec_neg_l}
\end{align}
From (\ref{update_w}) and (\ref{w_vec_neg_l}) it is readily seen that each element of the $k$th row of  $\mathbf{W}$ is updated at each time instance $n$, taking into account the most recent estimates of the remaining entries of the $k$th row in a cyclic manner. It is worthy to mention that this emerging iterative scheme, resulting from the espoused statistical independence among the elements of $\mathbf{W}$, can be charmingly reckoned as a relevant to the cyclic coordinate-descent strategy \cite{bertsekas1999nonlinear}. Following the same premise, for the column sparsity promoting parameters we get from (\ref{eq:sl}),
\begin{align}
 s_l(n) = \sqrt{\frac{\beta^{-1}(n-1)\delta_l(n)}{\hat{\boldsymbol{\mathit{w}}}^T_l(n)\boldsymbol{\Gamma}_l(n)\hat{\boldsymbol{\mathit{w}}}_l(n) + \sum^K_{k=1}\gamma_{kl}(n)\sigma^2_{\hat{w}_{kl}}(n) + q_{ll}(n)  }}, 
  \label{s_update}
\end{align}
where $q_{ll}(n)$ is the $l$th diagonal element of $\mathbf{Q}(n)$. As for the hyperparameters $\delta_l$'s of the $s_l$'s we have from (\ref{post:exp_delta}), (\ref{eq:invsl}) the following recursive equation 
\begin{align}
 \delta_l(n) = \frac{2\mu + (1-\lambda)^{-1} + K + 1}{2\nu + s_l^{-1}(n-1) + \delta_l^{-1}(n-1) } \label{post:update_delta}.
\end{align}
Note that  in (\ref{post:update_delta}) the size of the effective time window i.e. $(1-\lambda)^{-1}$, is used in place of $n$, as in \cite{themelis2014}. For $\gamma_{kl}$'s that independently favor sparsity on the entries of the subspace matrix $\mathbf{W}$, in an online scheme (\ref{eq:gkl}) takes the form,
\begin{align}
\gamma_{kl}(n) = \sqrt{\frac{\rho_{kl}(n)}{\beta(n-1) s_l(n-1) \left(\hat{w}_{kl}^2(n) + \sigma^2_{\hat{w}_{kl}}(n)   \right)}}
\end{align}
and for the hyperparameters $\rho_{kl}$'s, (\ref{post:exp_rho}) and (\ref{eq:gammakl}) yield
\begin{align}
 \rho_{kl}(n) = \frac{2(\psi +1)}{2\xi + \gamma_{kl}^{-1}(n-1) + \rho_{kl}^{-1}(n-1)}.
\end{align}
Finally, from (\ref{eq:postbeta})-(\ref{eq:lastbeta}) and applying some straightforward algebraic manipulations as in \cite{themelis2014}, we end up with the following efficient formula for computing the noise precision $\beta$, at each time iteration
\begin{align}
 \beta(n) = \frac{2\kappa + \frac{1}{1-\lambda}\left(K+L\right)+KL}{\Big(2 \theta + \sum_{k=1}^K\Big(d_k(n) - \hat{\mathbf{w}}^T_k(n)\boldsymbol{\mathit{t}}_k(n) +   \boldsymbol{\sigma}^T_{\hat{\mathbf{w}}_k}(n)\mathbf{r}_k(n) \Big) + \sum_{l=1}^L s_{l}(n)q_{ll}(n)\Big)}
\end{align}
where $\hat{\mathbf{w}}^T_k(n)$ is the $k$th row of $\hat{\mathbf{W}}(n)$, $\boldsymbol{\mathit{t}}_k(n)$ is the $k$th column of $\mathbf{T}(n)$,  $\boldsymbol{\sigma}_{\hat{\mathbf{w}}_k}(n) = \mathrm{diag}(\boldsymbol{\Sigma}_{\hat{\mathbf{w}}_k}(n))$  and $\mathbf{r}_k(n) = \mathrm{diag}(\mathbf{R}_k(n))$.

As it can be seen, most of the above defined quantities resolve to efficient time-updating formulas. In doing so, the need for taking into consideration the whole bunch of data, which is computationally prohibitive in applications dealing with big data, is eliminated. By collecting and putting in a proper order the previously derived expressions, we are led to the new online variational Bayes sparse subspace learning (OVBSL) algorithm, which is summarized in Table \ref{t:ovbsl}. The algorithm provides at each time iteration not only the sought estimates $\hat{\mathbf{x}}(n)$ and $\hat{\mathbf{W}}(n)$, but also estimates for all parameters of the model described in Section \ref{sec:proposed_model}. Note also that all these parameters are directly linked to specific distributions through the posterior inference analysis of Section \ref{sec:proposed_model}. By carefully inspecting OVBSL in Table \ref{t:ovbsl}, it can be shown that its computational complexity  is $\mathcal{O}(|\boldsymbol{\phi}(n)|L^2 + KL)$, where $|\boldsymbol{\phi}(n)|$ is the number of observed entries at time $n$. It should be emphasized that a significant reduction in the computational complexity has been achieved (which would be otherwise $\mathcal{O}(|\boldsymbol{\phi}(n)|L^3)$) by adopting the element-by-element estimation of $\hat{\mathbf{W}}$ via a coordinate-descent type procedure. As shown in Table \ref{t:ovbsl}, all hyperparameters of OVBSL are set and fixed to very small values at the initialization stage of the algorithm, as is the custom in sparse Bayesian learning schemes \cite{tipping2001sparse}. Hence, parameter fine tuning or cross-validation is entirely avoided and all parameters of the model are inferred from the data, rendering the proposed algorithm ideally accustomed for use in a real-time setting. In the next section the proposed algorithm is set in a unified framework with other related state-of-the-art techniques and its advantages in terms of performance and complexity are highlighted. 

\begin{table}
\centering
\caption{The OVBSL algorithm}
 \begin{tabular}{|l|}
 \hline 
 \bf{Initialize} : $\hat{\mathbf{W}}(0),\mathbf{S}(0),\beta(0),\boldsymbol{\Gamma}_k(0), \boldsymbol{\Sigma}_{\hat{w}_k}(0), k=1,2,\dots,K$ \\
 \bf{Set} $\mathbf{T}(0)=\mathbf{0}, \mathbf{P}_k(0)=\mathbf{0},d_k(0) = 0, k=1,2,\dots,K$\\
 \bf{Set} $\mu = 10^{-6}, \nu = 10^{-6}, \psi = 10^{-6}, \xi = 10^{-6}, \kappa=10^{-6},\theta = 10^{-6}$ \\
 \bf{Set} $\mathbf{Q}(0) = \mathbf{0}, \lambda$ \\
  \bf{for} $n=1,2,\dots$ \\
    \hspace{0.2cm} $\mathrm{Get} \;\; \mathbf{z}(n), \boldsymbol{\phi}(n)$\\
    \hspace{0.2cm}  $\boldsymbol{\Sigma}_{\hat{\mathbf{x}}}(n) = \beta^{-1}(n-1)\left(\hat{\mathbf{W}}^T(n-1)\boldsymbol{\Phi}_n\hat{\mathbf{W}}(n-1) + \sum^K_{k=1}\phi_{k}(n)\boldsymbol{\Sigma}_{\hat{\mathbf{w}}_k}(n-1)+ \mathbf{S}(n-1) \right)^{-1} $ \\
    \hspace{0.2cm}  $\hat{\mathbf{x}}(n) = \beta(n-1)\boldsymbol{\Sigma}_{\hat{\mathbf{x}}}(n)\hat{\mathbf{W}}^T(n-1)\mathbf{z}(n)$\\
    \hspace{0.2cm}  ${\cal \boldsymbol{\Sigma}}(n) = \boldsymbol{\Sigma}_{\hat{\mathbf{x}}}(n) + \hat{\mathbf{x}}^T(n) \hat{\mathbf{x}}(n)$\\
    \hspace{0.2cm}  $\mathbf{Q}(n) = \lambda \mathbf{Q}(n-1) + {\cal \boldsymbol{\Sigma}}(n)$\\
    \hspace{0.2cm} $\mathbf{T}(n) = \lambda \mathbf{T}(n-1) + \hat{\mathbf{x}}(n)\mathbf{z}^T(n)$ \\
    \hspace{0.2cm} \bf{for} $k=1,2,\dots,K,$ \\
    \hspace{0.5cm}$\mathbf{P}_k(n) =\lambda \mathbf{P}_k(n-1) + \phi_k(n){\cal \boldsymbol{\Sigma}}(n)$\\
    \hspace{0.5cm}$\mathbf{R}_k(n) = \mathbf{P}_k(n) + \boldsymbol{\varGamma}_k (n-1) \mathbf{S}(n-1)$ \\
    \hspace{0.5cm}$d_k(n) = \lambda d_k(n-1) + z^2_k(n)$\\
    \hspace{0.5cm}\bf{for} $l=1,2,\dots,L,$\\
    \hspace{0.8cm}$\hat{w}_{kl}(n) = \beta(n-1)\sigma_{\hat{{w}}_{kl}}(n-1)\left(t_{lk}(n) - \mathbf{r}^T_{k\neg l}(n)\hat{\mathbf{w}}_{k\neg l}(n)\right)$\\ 
    \hspace{0.8cm}$\sigma^2_{\hat{w}_{kl}}(n) = \beta^{-1}(n-1)r^{-1}_{k,ll}(n) $\\
    \hspace{0.8cm}$\rho_{kl}(n) = \frac{2(\psi +1)}{2\xi + \gamma_{kl}^{-1}(n-1) + \rho_{kl}^{-1}(n-1)}$\\
    \hspace{0.8cm}$\gamma_{kl}(n) = \sqrt{\frac{\rho_{kl}(n)}{\beta(n-1) s_l(n-1) \left(\hat{w}_{kl}^2(n) + \sigma^2_{\hat{w}_{kl}}(n)   \right)}}$\\
    \hspace{0.5cm}\bf{end}\\
    \hspace{0.5cm}\bf{Set} $\boldsymbol{\Sigma}_{\hat{\mathbf{w}}_k}(n) = \mathrm{diag}\left([\sigma^2_{\hat{w}_{k1}}(n),\sigma^2_{\hat{w}_{k2}}(n), \ldots,\sigma^2_{\hat{w}_{kL}}(n) ]^T\right)$ \\
    \hspace{0.2cm}\bf{end}\\
    \hspace{0.2cm}\bf{for} $l=1,2,\dots,L,$\\
    \hspace{0.5cm}$\delta_l(n) = \frac{2\mu + (1-\lambda)^{-1} + K + 1}{2\nu + s_l^{-1}(n-1) + \delta_l^{-1}(n-1) }$\\
    \hspace{0.5cm}$s_l(n) = \sqrt{\frac{\beta^{-1}(n-1)\delta_l(n)}{\hat{\boldsymbol{\mathit{w}}}^T_l(n)\boldsymbol{\Gamma}_l(n)\hat{\boldsymbol{\mathit{w}}}_l(n) + \sum^K_{k=1}\gamma_{kl}(n)\sigma^2_{\hat{w}_{kl}}(n) + q_{ll}(n)  }} $\\
    \hspace{0.5cm}\bf{end}\\
    \hspace{0.2cm}\bf{Set} $\mathbf{S}(n) = \mathrm{diag}([s_1(n),s_2(n),\dots,s_L(n)]^T)$\\
    \hspace{0.2cm}$\beta(n) = \frac{2\kappa + \frac{1}{1-\lambda}\left(K+L\right)+KL}{\Big(2 \theta + \sum_{k=1}^K\Big(d_k(n) - \hat{\mathbf{w}}^T_k(n)\boldsymbol{\mathit{t}}_k(n) +   \boldsymbol{\sigma}^T_{\hat{\mathbf{w}}_k}(n)\mathbf{r}_k(n) \Big) + \sum_{l=1}^L s_{l}(n)q_{ll}(n)\Big)}$\\
    \bf{end}\\ \hline
 \end{tabular}
 \label{t:ovbsl}
\end{table}
\section{Relation with state-of-the-art}\label{sec:relation_state_of_the_art}
\label{sec:rel}
In this section we investigate and highlight the connection of the new Bayesian algorithm with two other closely related techniques that have recently appeared in the literature, namely the PETRELS algorithm presented in \cite{petrels} and Algorithm 1 of \cite{Mardani2015}. All three algorithms under study are second-order online subspace learning schemes that deal with (possibly highly) incomplete data. Out of the three schemes, only the proposed algorithm has the provision to impose sparsity to the unknown subspace matrix. Hence, to make comparisons more clear we relax this constraint, that is we set $\boldsymbol{\Gamma}_l = {\mathbf I}_K$ for $l=1,2,\dots,L$ in our Bayesian model described in Section \ref{sec:proposed_model}. As we shall see below, this Bayesian model can be considered as a unified framework from which all three schemes may originate. To be more specific, let us first recall the likelihood function of the model given in (\ref{eq:lklhood}), which can be expressed as
\begin{align}
p( \mathbf{Z}\mid \mathbf{X},{\mathbf W},\beta) & \propto \exp \left(-\frac{\beta}{2} \left\| \boldsymbol{\Lambda}^{\frac{1}{2}}(\mathbf{Z} - \boldsymbol{\Phi} \odot (\mathbf{X}\mathbf{W}^T))\right\|_F^2\right).
\label{eq:lifu}
\end{align}
Based on (\ref{eq:lifu}), the maximum likelihood (ML) estimator is obtained by minimizing w.r.t $\mathbf{X}$ and $\mathbf{W}$ the negative log-likelihood, resulting in the following minimization problem  
\begin{align}
\mathrm{(P1)} \;\;\;\;\;\;\;\;\;\;\;\; \min_{\mathbf{X},\mathbf{W}}\frac{\beta}{2} \left\| \boldsymbol{\Lambda}^{\frac{1}{2}}(\mathbf{Z} - \boldsymbol{\Phi} \odot (\mathbf{X}\mathbf{W}^T))\right\|_F^2 \nonumber
\end{align}
The so-termed PETRELS algorithm presented in \cite{petrels} solves $\mathrm{(P1)}$ through an online alternating (between $\mathbf{X}$ and $\mathbf{W}$)  least squares (LS) technique, which provides both the estimates of the subspace matrix $\mathbf{W}(n)$ and the new vector of projection coefficients $\mathbf{x}(n)$ at each time iteration. However, by solving $\mathrm{(P1)}$ PETRELS does not take any special care for revealing the true rank of the sought subspace. The algorithm starts with an overestimate $L$ of the rank (number of columns of $\mathbf{W}$) and the estimates returned by the algorithm are related to a subspace of rank $L$, which may be far from the true rank. 

Let us now consider the likelihood function given in (\ref{eq:lifu}) and the first level (Gaussian) priors of $\mathbf{X}$ and $\mathbf{W}$ in our model given by (\ref{prior:X}) and (\ref{prior:W}) for $s_l = s$, $l=1,2,\ldots,L$, where $s$ is a constant parameter  and not a random variable that can be determined from data. Then (\ref{prior:X}) and (\ref{prior:W}) are rewritten as,
\begin{align}
p(\mathbf{X}\mid s,\beta) \propto \exp \left( -\frac{\beta}{2}s\left\|\boldsymbol{\Lambda}^{\frac{1}{2}}\mathbf{X}\right\|_F^2\right) \label{eq:prX},  \\
p(\mathbf{W}\mid s,\beta) \propto \exp \left( -\frac{\beta}{2}s\left\|\mathbf{W}\right\|_F^2 \right)  \label{eq:prW}.
\end{align}
From the likelihood (\ref{eq:lifu}) and the priors (\ref{eq:prX}) and (\ref{eq:prW}) the maximum a-posteriori probability (MAP) estimator of $\mathbf{X}$ and $\mathbf{W}$ defined as, 
\begin{align}
\min_{\mathbf{X},\mathbf{W}}\left\{ -\log p(\mathbf{X},\mathbf{W}\mid \mathbf{Z}) \right\} \equiv \min_{\mathbf{X},\mathbf{W}}\left\{-\log\left[ p(\mathbf{Z}\mid \mathbf{X},\mathbf{W},\beta)p(\mathbf{X}\mid s,\beta)p(\mathbf{W}\mid s,\beta)\right] \right\},
\end{align}
is expressed as, 
\begin{align}
\mathrm{(P2)} \;\;\;\;\;\;\;\; \min_{\mathbf{X},\mathbf{W}}\frac{\beta}{2} \left[ \left\| \boldsymbol{\Lambda}^{\frac{1}{2}}(\mathbf{Z} - \boldsymbol{\Phi} \odot (\mathbf{X}\mathbf{W}^T))\right\|_F^2 + s\left\|\boldsymbol{\Lambda}^{\frac{1}{2}}\mathbf{X}\right\|_F^2 + s\left\|\mathbf{W}\right\|_F^2 \right]. \nonumber
\end{align}
The minimization problem $\mathrm{(P2)}$ is at the heart of the analysis in \cite{Mardani2015}. Algorithm 1 of \cite{Mardani2015} is a second-order alternating ridge regression type scheme that solves $\mathrm{(P2)}$ sequentially and provides estimates of $\mathbf{W}(n)$ and $\mathbf{x}(n)$ at each time iteration. In \cite{Mardani2015}, to promote the low-rank data representation, the minimization problem is originally formulated as
\begin{align}
\mathrm{(P2')} \;\;\;\;\;\;\;\; \min_{\mathbf{V}} \beta \left[ \frac{1}{2}\left\| \boldsymbol{\Lambda}^{\frac{1}{2}}(\mathbf{Z} - \boldsymbol{\Phi} \odot \mathbf{V})\right\|_F^2 + s\left\|\boldsymbol{\Lambda}^{\frac{1}{2}}\mathbf{V}\right\|_{*}  \right]. \nonumber
\end{align}
Then, in search for a nuclear-norm surrogate that would be amenable to online processing, $||\boldsymbol{\Lambda}^{\frac{1}{2}}\mathbf{V}||_{*}$ in $\mathrm{(P2')}$ is replaced by its upper bound $(||\boldsymbol{\Lambda}^{\frac{1}{2}}\mathbf{X}||_F^2 + ||\mathbf{W}||_F^2)/2$, with $\mathbf {V} = \mathbf{X}\mathbf{W}^T$, thus leading to $\mathrm{(P2)}$. Even though, compared to PETRELS, a more direct promotion of the low-rankness of the underlying subspace is employed in \cite{Mardani2015}, again an overestimate $L$ of the true rank is used and Algorithm 1 of \cite{Mardani2015} lacks a specific mechanism for imposing low-rankness explicitly by reducing the initial rank to the true rank as the algorithm evolves.  In addition, special care should be taken for the parameter $s$ that must be properly selected and updated in the framework of an online scheme. 

Let us, finally, employ the complete Bayesian model of Section \ref{sec:proposed_model} (with the exception of the subspace matrix sparsity promoting parameters $\gamma_{kl}$'s which are set to $1$). In such a case, as shown in Appendix \ref{sec:apendix}, the joint prior of $\mathbf{X}$ and $\mathbf{W}$ can be expressed as 
\begin{align}
p(\mathbf{X},\mathbf{W}\mid \boldsymbol{\delta},\beta) \propto \exp \left( -\beta^{\frac{1}{2}}\sum\limits_{l=1}^L \delta_l^{\frac{1}{2}}\left(\left\| \mathbf{x}_l \right\|_{2,\boldsymbol{\Lambda}}^2 + \left\| \mathbf{w}_l \right\|_2^2 \right)^{\frac{1}{2}} \right). \label{eq:prXW}
\end{align}
From (\ref{eq:lifu}) and (\ref{eq:prXW}) the MAP estimator for $\mathbf{X}$ and $\mathbf{W}$ is now obtained from the solution of the following minimization problem,
\begin{align}
\mathrm{(P3)} \;\;\;\;\;\;\;\; \min_{\mathbf{X},\mathbf{W}} \left[ \frac{\beta}{2} \left\| \boldsymbol{\Lambda}^{\frac{1}{2}}(\mathbf{Z} - \boldsymbol{\Phi} \odot (\mathbf{X}\mathbf{W}^T))\right\|_F^2 + \beta^{\frac{1}{2}}\sum\limits_{l=1}^L \delta_l^{\frac{1}{2}}\left(\left\| \mathbf{x}_l \right\|_{2,\boldsymbol{\Lambda}}^2 + \left\| \mathbf{w}_l \right\|_2^2\right)^{\frac{1}{2}}  \right]. \nonumber
\end{align}
Note that the regularizing summation  term in $\mathrm{(P3)}$ corresponds to the {\it weighted} $\ell_{2}/\ell_{1}$ norm of the matrix $[\boldsymbol{(\Lambda}^{\frac{1}{2}}\mathbf{X})^T \;\;  \mathbf{W}^T]^T$ \cite{Kowalski2009303}, which is known to impose column sparsity \cite{Kowalski2009303} and thus explicitly reducing the rank of $\mathbf{W}$, leading to more consistent estimates. Derived from the Bayesian model of Section \ref{sec:proposed_model}, the minimization problem $\mathrm{(P3)}$ is closely related to the analysis and algorithm presented in the current paper. It should be emphasized though that the proposed algorithm is not a recursive alternating MAP estimation scheme, but a variational Bayes type technique that can be deemed as a generalization of the MAP approach. While a MAP procedure would provide the point estimates of the parameters of interest $\mathbf{X}$ and $\mathbf{W}$, the proposed algorithm returns in addition the approximate distribution of all parameters involved in the model, including the weighting parameters $\delta_l$'s, which are now estimated directly from the data. Summarizing and compared to \cite{petrels} and \cite{Mardani2015} the proposed algorithm a) is equipped with an inherent mechanism for inducing column sparsity and thus reducing the rank of the latent subspace matrix dynamically and b) is fully automatic as all parameters of the model are estimated from the data and thus any need for preselection or fine tuning is entirely avoided.  

In Table \ref{complexity_table}, OVBSL is compared in terms of computational complexity with other related state-of-the-art algorithms. Besides PETRELS and Algorithm 1 of \cite{Mardani2015} mentioned above, two other algorithms are included, namely GROUSE reported in \cite{grouse} and Algorithm 2 of \cite{Mardani2015}, which is a first-order stochastic approximation type scheme. We see from Table \ref{complexity_table} that the proposed algorithm requires less computations per iteration than Algorithm 1, while it has similar complexity with the remaining three algorithms. Note though that, as it will be also shown in the next section, PETRELS and GROUSE perform well under the condition that the true subspace rank $r(n)$ is known, while Algorithm 2, being a first-order algorithm is expected to have a much slower convergence rate compared to the remaining second-order schemes included in Table \ref{complexity_table}. 

\begin{table*}
\centering
\caption{Computational complexity of online subspace learning algorithms}
\begin{tabular}{| c | c | c | c | c |c|}
\hline 
Algorithm & GROUSE \cite{grouse} & PETRELS \cite{petrels} & Alg. 1 of \cite{Mardani2015} & Alg. 2 of \cite{Mardani2015} & OVBSL \\\hline
Comp. complexity & $\mathcal{O}(|\phi(n)|L^2 + KL)$ &  $\mathcal{O}(|\phi(n)|L^2)$  & $\mathcal{O}(|\phi(n)|L^3)$  & $\mathcal{O}(|\phi(n)|L^2 + KL)$ & $\mathcal{O}(|\phi(n)|L^2 + KL)$ \\\hline
\end{tabular}
\label{complexity_table}
\end{table*}
\section{Experimental Results}\label{sec:experimental}
In this section, the effectiveness of the proposed algorithm is corroborated in a variety of experiments carried out on synthetic and real data. 
\subsection{Synthetic data experiments}
 In the following, three different experiments are presented. First we endeavor to unveil the potency of OVBSL in tackling  matrix completion. The ubiquitous subspace estimation problem is subject to scrutiny in the second experiment. It should be noted that, in those two experiments, the sparsity imposition on the subspace matrix from OVBSL is purposely relaxed, that is we set $\boldsymbol{\Gamma}_l=\mathbf{I}_K, \forall l=1,2,\dots,L$. The performance of OVBSL in the challenging {\it sparse} subspace estimation problem is explored in the last experiment of this subsection. Therein, the aforementioned favorable characteristic of OVBSL algorithm, i.e., its potential to impose sparsity on the subspace matrix, is thoroughly investigated. To this end, the parameters $\gamma_{kl}$'s are then considered ``active'', normally taking their values according to the full Bayesian model analytically described above.
\subsubsection{Online matrix completion}
In order to assess the performance of OVBSL algorithm in recovering missing data, we simulate a low dimensional subspace $\mathbf{U} \in\mathcal{R}^{K\times r}$ with $K=500$ and $r=5$ and Gaussian i.i.d entries $u_{kl}\sim \mathcal{N}(0,\frac{1}{K})$. Next, $20000$ $r \times 1$ projection coefficient vectors $\mathbf{c}(n)$ are produced according to a Gaussian distribution $c_{l}(n)\sim \mathcal{N}(0,1)$. The signal $\mathbf{y}(n)$ at time $n$ is then generated by the product $\mathbf{U}\mathbf{c}(n)$ after adding i.i.d Gaussian noise $\mathbf{e}(n)\sim \mathcal{N}(\mathbf{0},\beta\mathbf{I}_K)$. To model the missing entries, we randomly select a fraction  $\pi$ of the entries from each datum $\mathbf{y}(n)$, which are assumed to be known, whereas the rest $(1-\pi)\times 100$\%  of the elements are considered to be missing. To show the merits of the proposed OVBSL algorithm, we compare it to three state-of-the-art techniques, namely GROUSE,\cite{grouse}, PETRELS \cite{petrels} and Algorithm 1 of \cite{Mardani2015}. It is worthy to mention that, as also previously mentioned, both GROUSE and PETRELS hinge on the assumption that the rank of the underlying subspace is known. Contrary, Algorithm 1 of \cite{Mardani2015} utilizes $\ell_2$-norm regularization as described in the previous section, that robustifies the algorithm in the absence of this knowledge. Finally, the step size parameter of GROUSE and the low-rank regularization parameter of Algorithm 1 of \cite{Mardani2015} are set to $0.1$.

In the sequel, to make things more interesting, we adhere to the challenging but realistic scenario whereby the true rank of the underlying subspace is unknown. Along this line, the rank of the subspace matrix is accordingly initialized in all tested algorithms to an overestimate of the true rank, namely  $L=10$. Our initial objective is to demonstrate the effectiveness of the proposed OVBSL algorithm when certain amounts of data are missing. To this end, we carry out two experiments corresponding to different fractions of the observed entries i.e. $\pi = \{0.25,0.75\}$, keeping the noise precision $\beta$ fixed to $10^{3}$. Since the competence of the subspace learning algorithms in tracking possible changes of the sought subspace is of crucial importance in many applications, an abrupt change of the subspace is induced at $n=5000$ for $\pi=0.25$. The performance of the tested algorithms is evaluated in terms of the normalized running average estimation error (NRAEE) defined as: $\mathrm{NRAEE}(n) = \frac{1}{n}\sum^n_{i=1}\frac{\|\hat{\mathbf{y}}(i) - \mathbf{y}(i)\|_2}{\|\mathbf{y}(i)\|_2}$ where $\hat{\mathbf{y}}(i) = \hat{\mathbf{W}}(i)\hat{\mathbf{x}}(i)$ and is shown in Fig. \ref{exp:fig1}a. It is clear that the proposed OVBSL algorithm outperforms its rivals for both values of the fraction of the observed data $\pi$. At the same time, OVBSL is proven to be competent in tracking sudden changes of the latent subspace, since the transient deterioration of its performance caused by the deliberate change induced  at $n=5000$ is swiftly rectified in the subsequent iterations. Notably, in the lack of knowledge of the true rank of the subspace, PETRELS becomes unstable while GROUSE gets stack to local minima. Contrary, algorithm 1 of \cite{Mardani2015} presents a robust behavior, though with clearly  less reconstruction accuracy compared to the proposed OVBSL algorithm. 

Next, we examine the robustness of OVBSL to noise corruption. To do so, we keep the fraction of the observed entries fixed to $\pi=0.4$, focusing on the behavior of OVBSL and the competing schemes for three different values of the noise precision i.e. $\beta = \{ 10^{4},10^{3},10^{2}\}$. Fig. \ref{exp:fig1}b depicts the NRAEE obtained by the tested algorithms in the three different cases examined. It is easily noticed that herein as well, OVBSL achieves higher reconstruction accuracy than the competing schemes for all different $\beta$'s, thus corroborating its strength to various levels of noise corruption.
\begin{figure}
\begin{tabular}{c c}
\includegraphics[width=0.48\textwidth,height=0.35\textwidth]{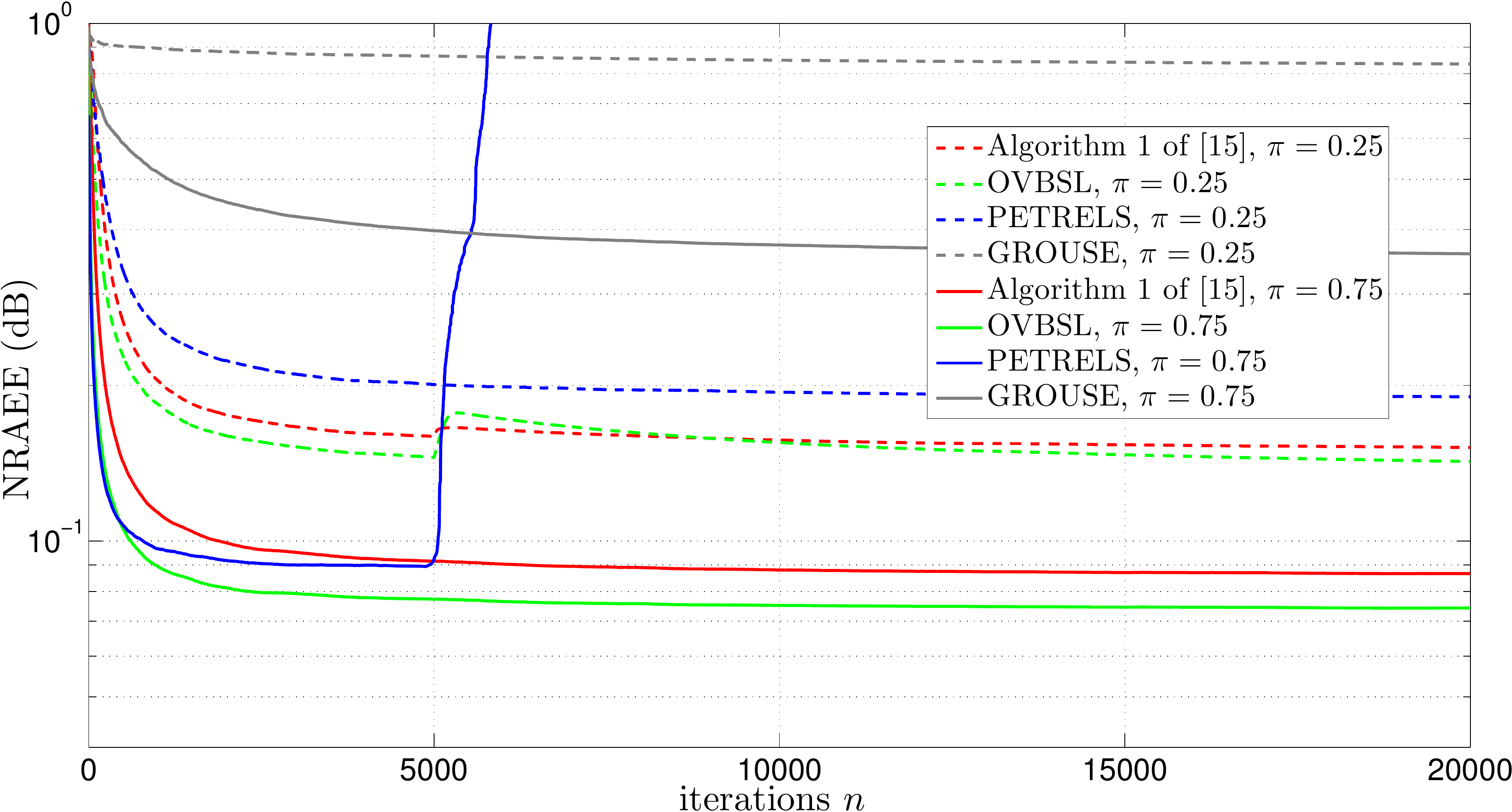} & \includegraphics[width=0.48\textwidth,height=0.35\textwidth]{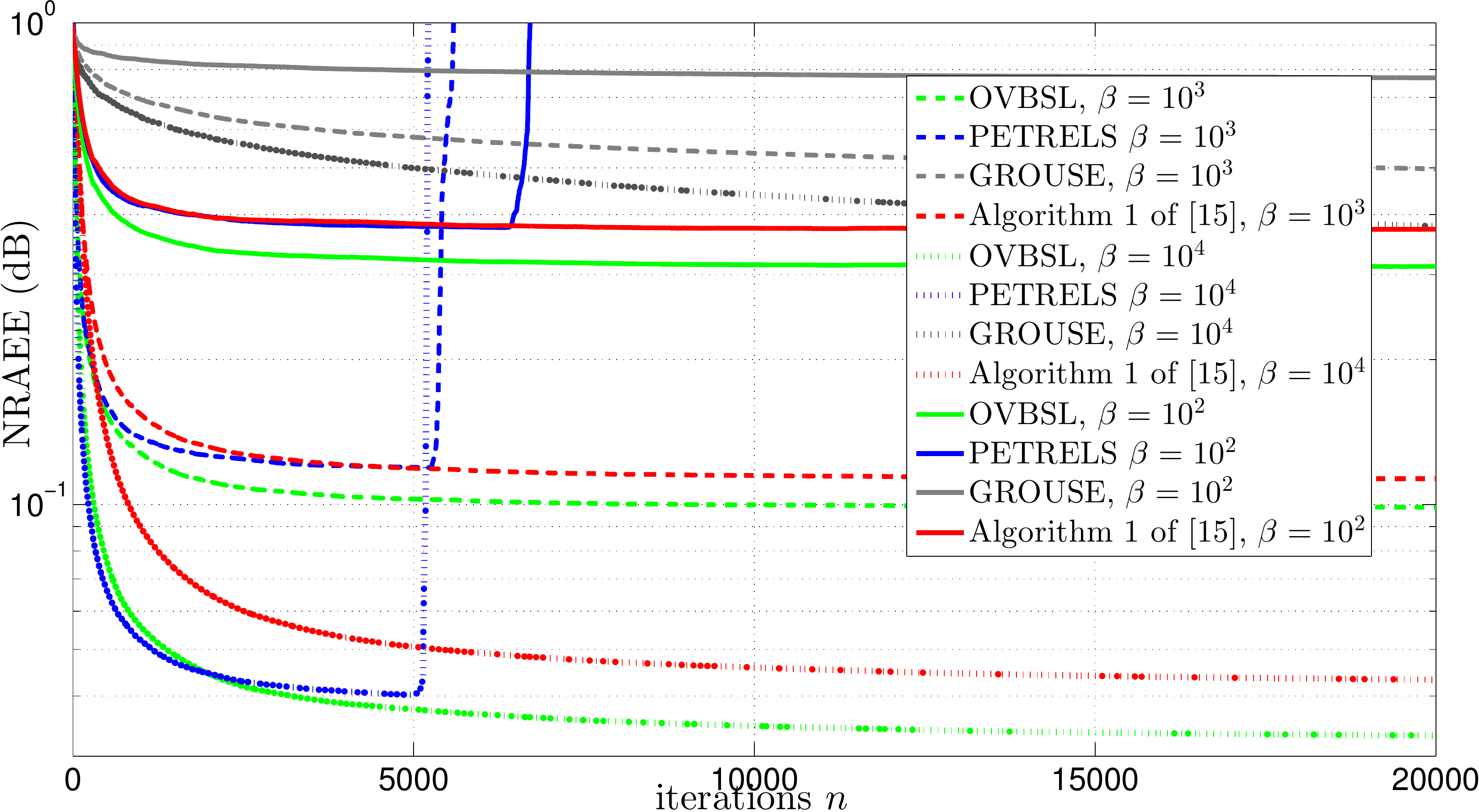} \\
\scriptsize a) Robustness to different fractions of the observed entries ($\pi$) & \scriptsize b) Sensitivity to different levels of noise corruption 
\end{tabular}
\caption{Performance comparison among OVBSL, Algorithm 1 of \cite{Mardani2015}, PETRELS and GROUSE for the matrix completion problem.}
\label{exp:fig1}
\end{figure}
\subsubsection{Online subspace estimation}
In this experiment we aspire to manifest the ability of the proposed OVBSL algorithm in reliably estimating the underlying low-rank subspace. In light of this, OVBSL is compared to Algorithm 1 of \cite{Mardani2015}, which also seems to be robust to subspace rank overestimation. To better explore the efficiency of those two tested algorithms on this finicky task, we keep the noise precision $\beta$ fixed  to $10^{3}$, assuming that half of the data are observed ($\pi=0.5$). The rank $L$ of the subspace matrices of both algorithms is initialized to $25$. Next, following the same process detailed above, incomplete data are generated different times, each corresponding to a specific rank $r$ of the true subspace, namely $r=\{5,10,15,20,25\}$. The estimate of the subspace obtained from both OVBSL and Algorithm 1 of \cite{Mardani2015}, is assessed as time evolves by means of the normalized subspace reconstruction error (NSRE) defined as $\mathrm{NSRE}(n) = \frac{\|\mathcal{P}_{\hat{\mathbf{W}}^{\bot}(n)}\mathbf{U}\|_F}{\|\mathbf{U}\|_F}$\footnote{$\mathcal{P}_{\hat{\mathbf{W}}^{\bot}(n)}\mathbf{U}$ denotes the projection of the true subspace matrix $\mathbf{U}$ to the orthogonal complement of the subspace spanned by the columns of the estimated subspace matrix $\hat{\mathbf{W}}^{\bot}(n)$.}, shown in Fig. \ref{exp:fig2}. Remarkably, OVBSL achieves more accurate estimates of the subspace for all the different ranks tested. Moreover, it should be emphasized that the less the true rank, the more accurate the subspace estimate obtained by OVBSL is. On the other hand, this is not the case for Algorithm 1 of \cite{Mardani2015}, which presents almost the same NSRE regardless of the true rank of the subspace. It should be pointed out that the superior performance of OVBSL as compared to Algorithm 1 of \cite{Mardani2015} in terms of NSRE, does turn out to be in line with its ability in revealing the true rank of the sought subspaces after convergence.
\begin{figure}
\centering
\includegraphics[width=0.7\textwidth,height=0.35\textwidth]{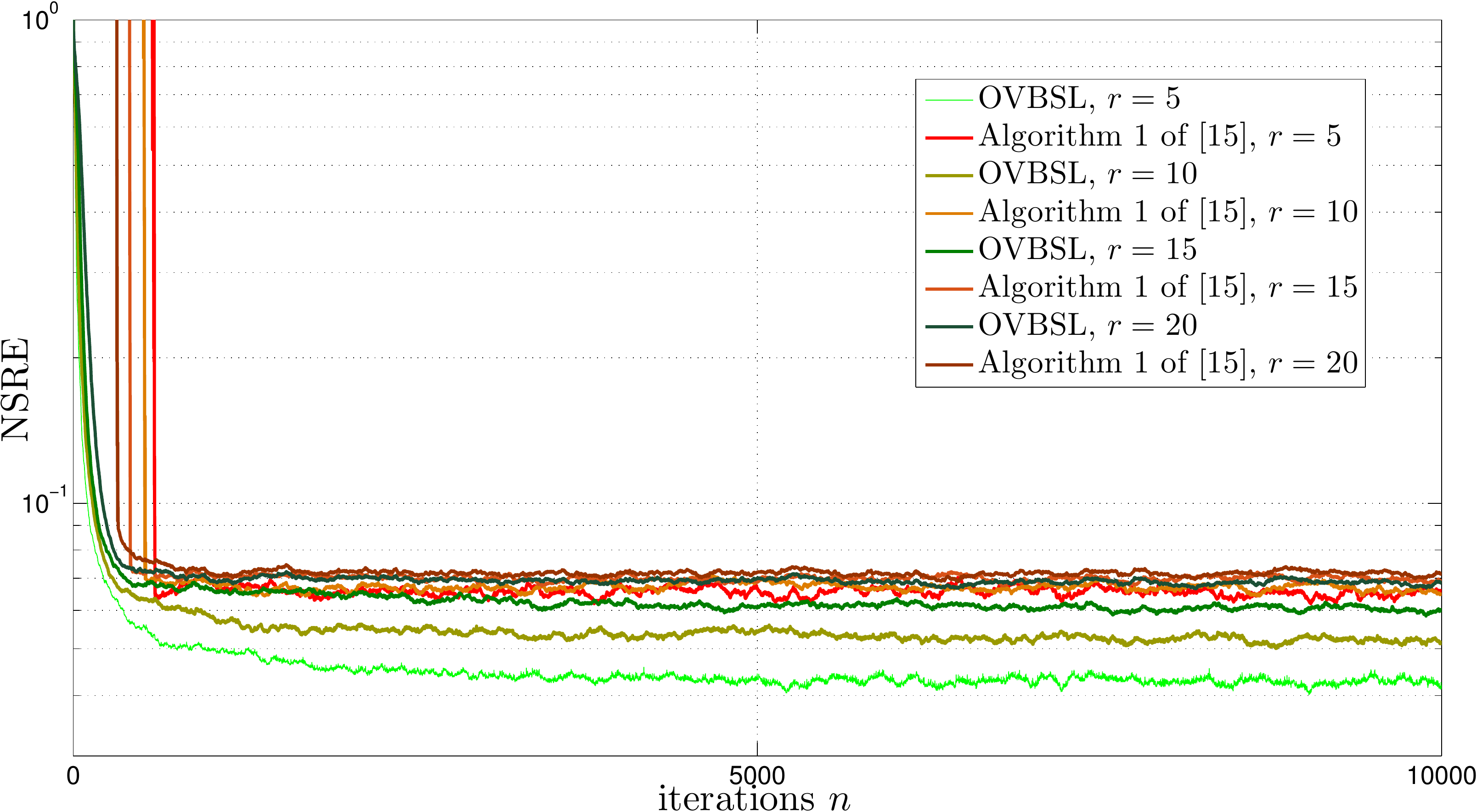} 
\caption{NSRE of OVBSL and Algorithm 1 of \cite{Mardani2015} for the subspace estimation problem.}
\label{exp:fig2}
\end{figure}
\subsubsection{Online sparse subspace estimation}
In the following, the compelling feature of OVBSL to favor {\it sparse} subspace estimates is thoroughly explored. To clearly demonstrate the merits of this key aspect of our newly introduced algorithm, a sparse subspace matrix $\mathbf{U}$  of rank $r=5$ is modeled. Then, the same above-described process is adopted for producing $20000$ projection coefficient vectors $\mathbf{c}(n)$, that give rise to the corresponding signals $\mathbf{U}\mathbf{c}(n)$. Finally, Gaussian i.i.d noise of precision $\beta=10^{3}$ is assumed to contaminate the datums. The benefits emerging from taking into account the sparsity existing in the unknown subspace matrix, come to light by exploring OVBSL's performance for different levels of sparsity imposed on it, namely $0.7$ and $0.9$. For now, focusing on the subspace matrix estimation problem, we depart from the matrix completion problem considering that data are fully observed (hence the fraction of the observed entries $\pi$ equals to $1$) and we test two versions of OVBSL, that is, when sparsity of the subspace a) is taken into account and b) is disregarded in the same way explained earlier. In both cases, the subspace matrices are initialized to an overestimate of the rank, i.e., $L=10$.  Fig. \ref{exp:fig3} depicts the NSRE obtained for the two versions of OVBSL as time evolves. As it can be readily seen, OVBSL achieves subspace estimates of higher accuracy compared to its so to speak non-sparse version. It should be noted that the gains  obtained by the sparse OVBSL are becoming abundantly clear as the sparsity level increases.

Next, OVBSL is probed in the challenging problem of sparse subspace estimation from partially observed data. Towards this, the same experimental setting described above is followed and the sparsity level is set to $0.7$. Then, we consider two different fractions of observed entries, that is $\pi=\{0.5,0.75 \}$. OVBSL is again evaluated for the two cases corresponding to its sparse and non-sparse version, initializing the rank $L$ of subspace matrices to $5$ and using NSRE as the performance metric. From Fig. \ref{exp:fig4b}, it is verified that albeit data are incomplete, taking advantage of the sparsity of the subspace matrix is still meaningful when the assumption of sparse subspace is valid.
\begin{figure}[h]
\centering
\includegraphics[width=0.7\textwidth,height=0.35\textwidth]{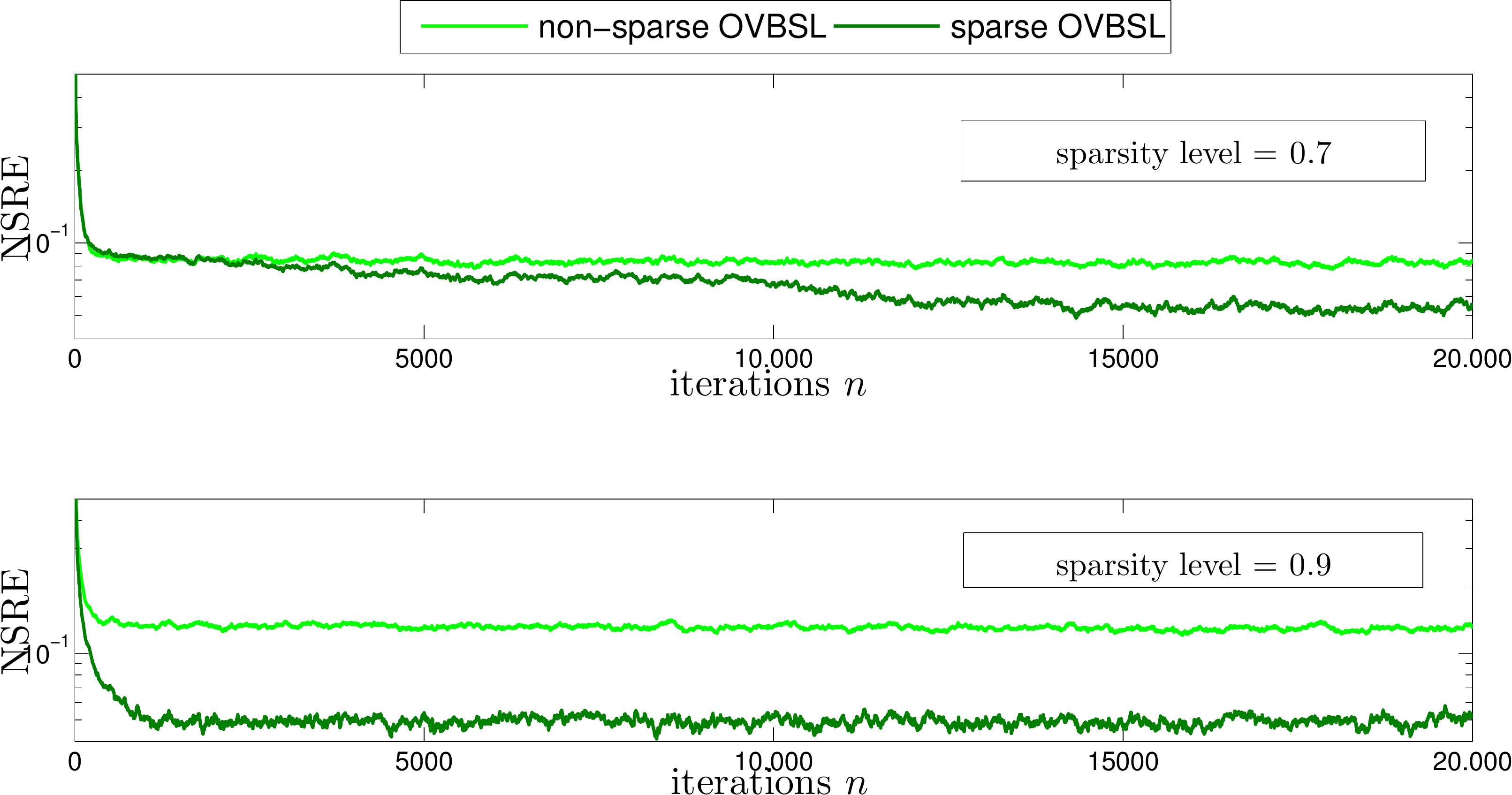}
\caption{Performance comparison between sparse and non-sparse versions of OVBSL for different sparsity levels of the subspace matrix and $\pi = 1$.}
\label{exp:fig3}
\end{figure}
\begin{figure}
\centering
 \includegraphics[width=0.7\textwidth,height=0.35\textwidth]{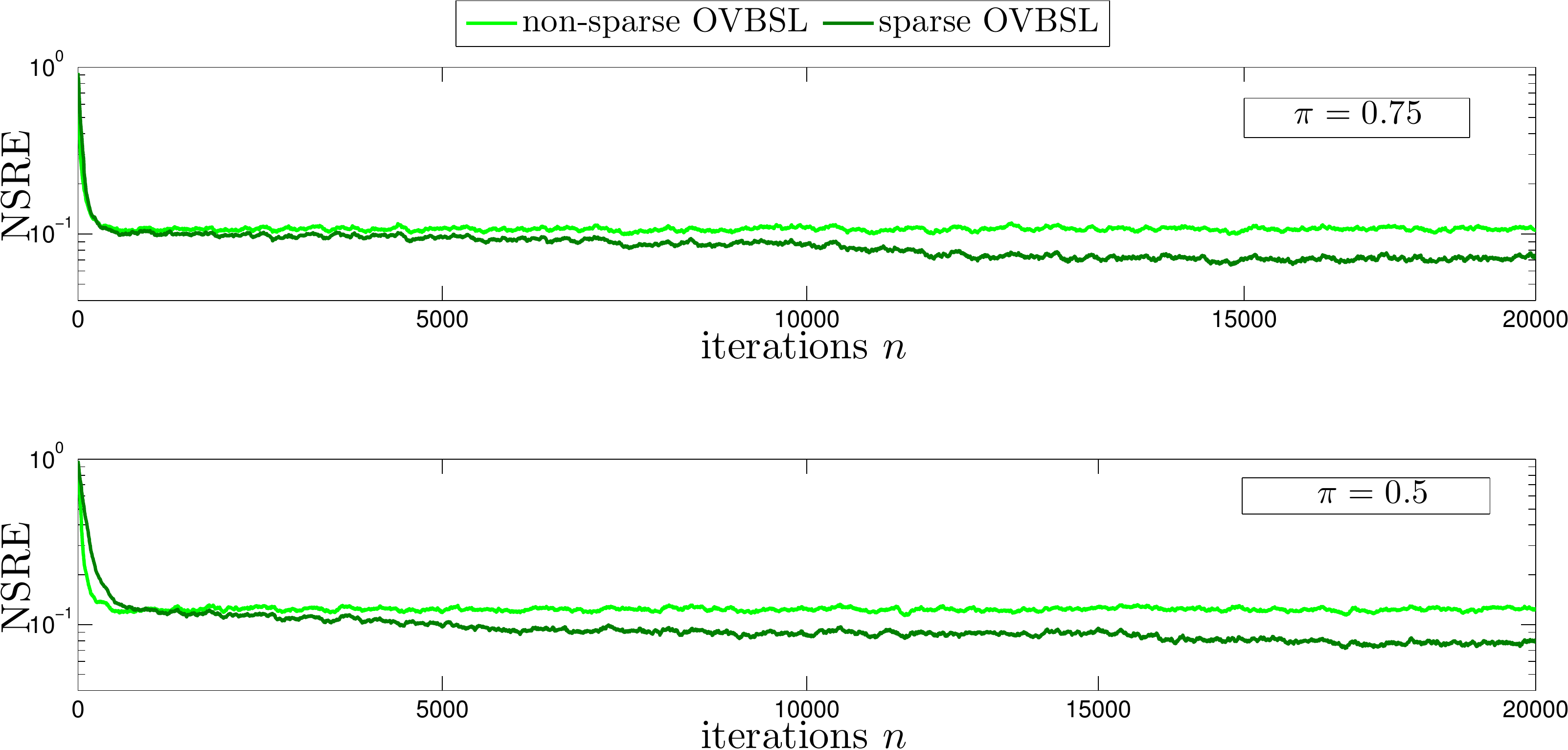}
 \caption{Performance of sparse and non-sparse versions of OVBSL for different percentage of missing entries and  subspace sparsity level $0.7$.}
 \label{exp:fig4b}
\end{figure}
\subsection{Real data experiments}
In this part of the paper, the efficiency of OVBSL algorithm is investigated on real data. More concretely, we conduct two different experiments corresponding a) to hyperspectral image reconstruction out of partially observed measurements and b) to the eigenface learning problem.
\subsubsection{Pixel-by-pixel hyperspectral image recovery}
A hyperspectral image (HSI) is a collection of multiple grayscale images captured at many contiguous spectral bands (channels), thus forming a so-called spectral cube. As a result of this, each pixel in a HSI is represented by a vector of size equal to the number of spectral bands and is called pixel {\it spectral signature}. The entries of this vector are the radiance values of the spatial area corresponding to the pixel in all spectral channels. 
A key characteristic of HSIs is the high degree of correlation they present, both in the spectral and the spatial domains, \cite{giampouras2015simultaneously}. Given a  HSI, let us form a matrix with its rows corresponding to the pixels of the HSI, and its columns to the spectral bands.  In doing so, it can be easily seen that the underlying high coherence appearing both in columns and rows leads to a matrix that may be of very low rank, as compared to its dimensions. Actually, this fact gives us good grounds for exploiting the low-rank structure in favor of recovering HSIs, in cases that data either are partly missing or have suffered by severe noise corruption.

In the following, we test the performance of OVBSL in recovering the Salinas Valley HSI, \cite{giampouras2015simultaneously}, from two different fractions of its entries. The 10th band image of this HSI is shown in Fig. $6$a. Since our proposed algorithm processes the data in an online fashion, we assume that the aforementioned time instances, hereafter, correspond to the sequence of pixels. Put differently, OVBSL processes the pixel spectral signatures (which are the rows of the formed matrix) one-by-one, as if they were becoming available in a streaming fashion. Notably, this type of processing aside from reducing the computational complexity, it alleviates the need for memory storage, thus paving the way for on-board processing. OVBSL is tested for two different fractions of observed entries, namely $\pi=\{0.1,0.2\}$, as illustrated for the 10th band image in Figs. $6$b and $6$e. The rank of the subspace matrix for the two cases is initialized to $L=\{5,10\}$ correspondingly. Figs. $6$c, $6$f and $6$d, $6$g  show the reconstructed $10$th band of the HSI as well as the residual error defined as the mean absolute value of the  difference between the true $\mathbf{Y}$ image and the estimate $\hat{\mathbf{Y}}$ i.e., $\frac{1}{204}\sum^{204}_{i=1} |\mathbf{Y}_i - \hat{\mathbf{Y}}_i|$ (where $\mathbf{Y}_i$ denotes the $i$th band image of $\mathbf{Y}$). Favorably, in both cases OVBSL seems to reconstruct reliably the Salina Valley HSI, thus proving its competence on a real dataset. 

\begin{figure}
\resizebox{.95\textwidth}{0.35\textwidth}{
\begin{tabular}{c | c}
\begin{tabular}{c}
 \centering
\subfloat[$\mathbf{Y}_{10}$ ($10$th band of Salinas HSI)]{\includegraphics[width=0.25\textwidth]{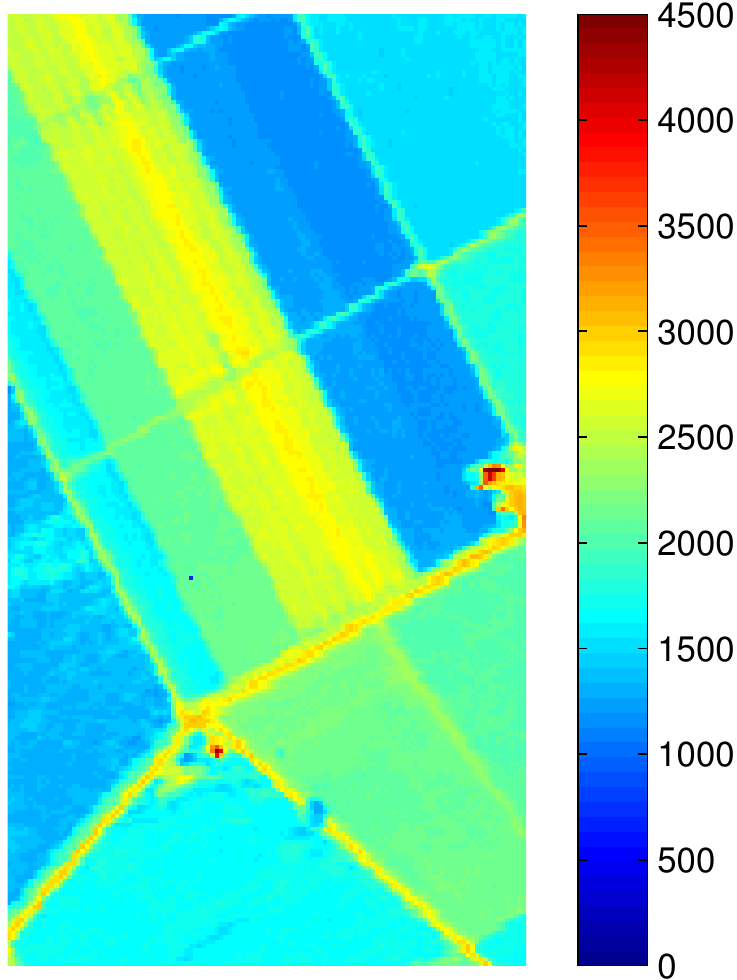}}
\end{tabular} &
\begin{tabular}{c c c}
\subfloat[incomplete image, $\pi=0.2$]{\includegraphics[width=0.25\textwidth]{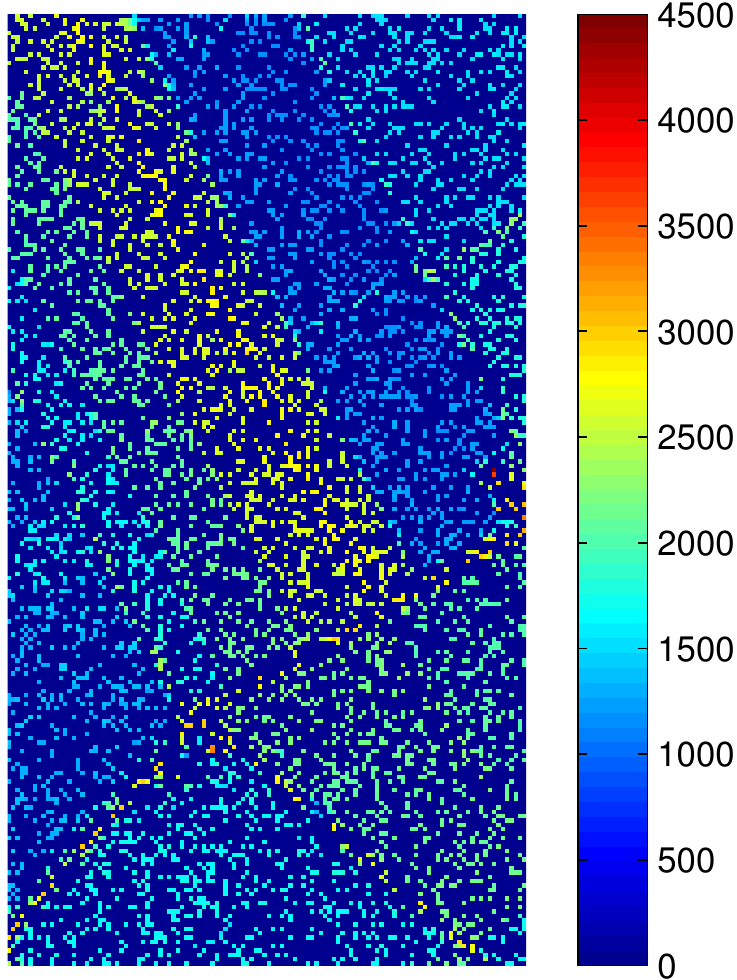}}& \subfloat[reconstructed image $\hat{\mathbf{Y}}_{10}$]{\includegraphics[width=0.25\textwidth]{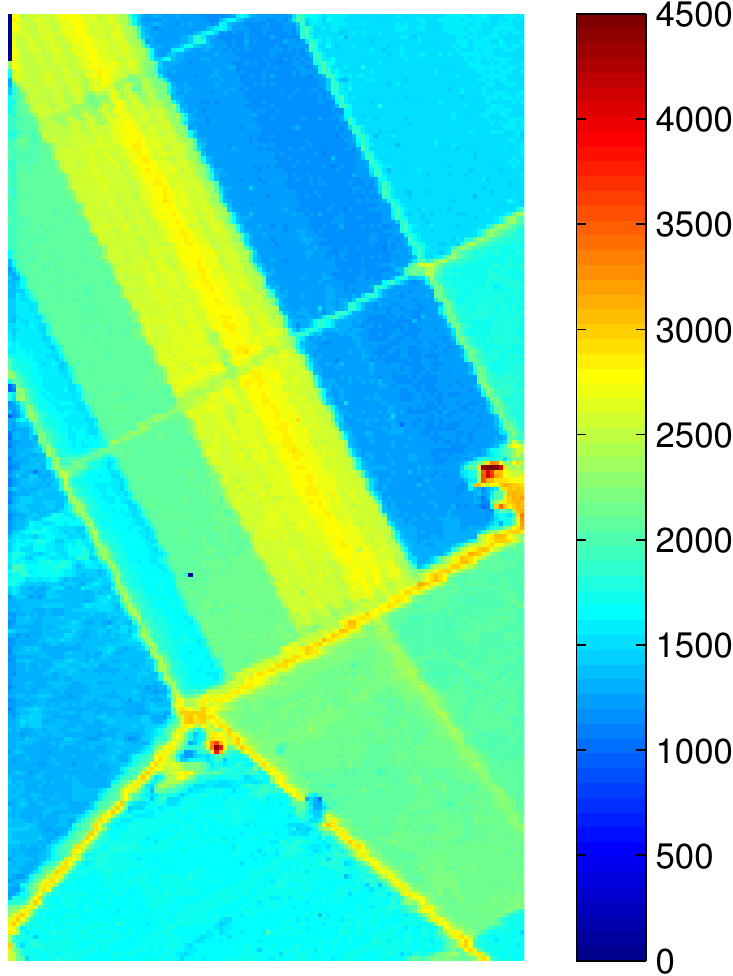}}&\subfloat[residual error]{\includegraphics[width=0.25\textwidth]{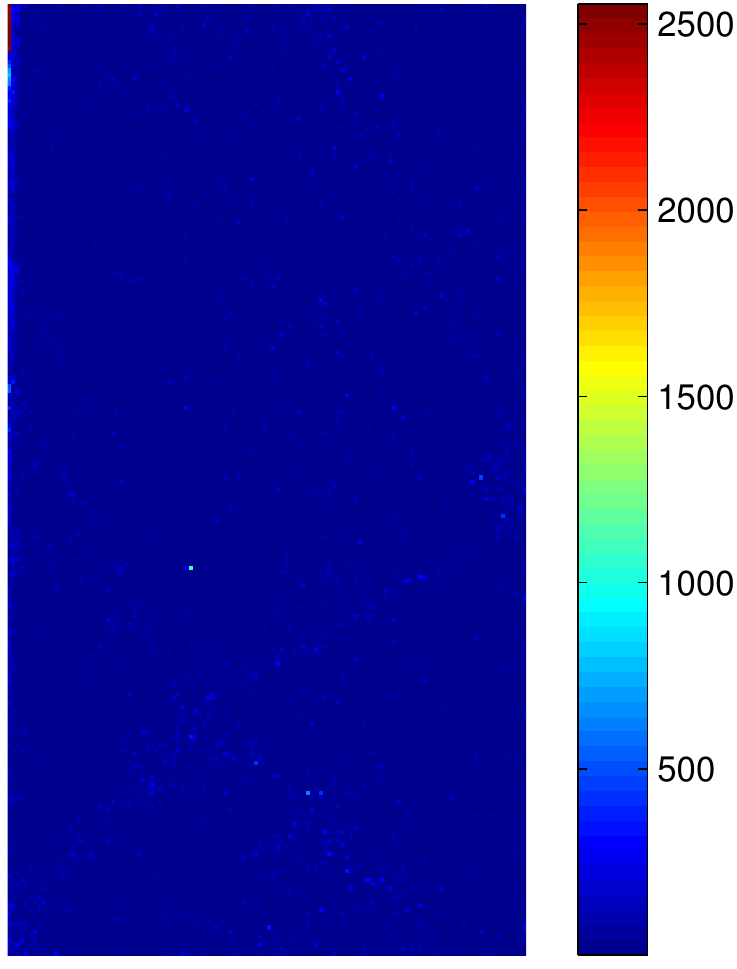}} \\
\subfloat[incomplete image, $\pi=0.1$]{\includegraphics[width=0.25\textwidth]{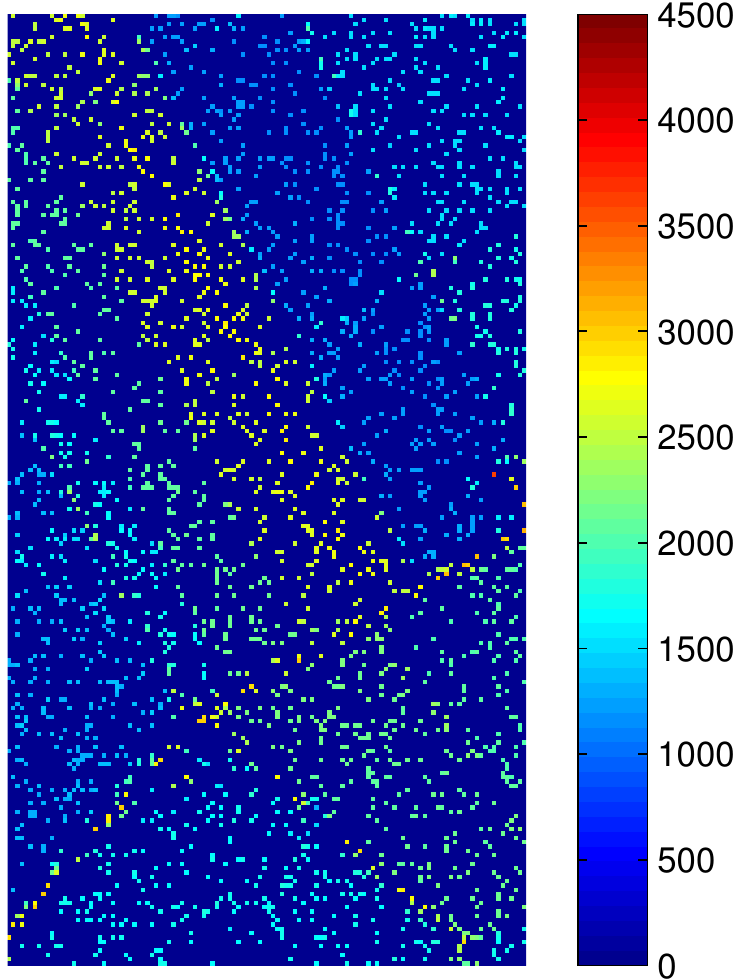}}& \subfloat[reconstructed image $\hat{\mathbf{Y}}_{10}$]{\includegraphics[width=0.25\textwidth]{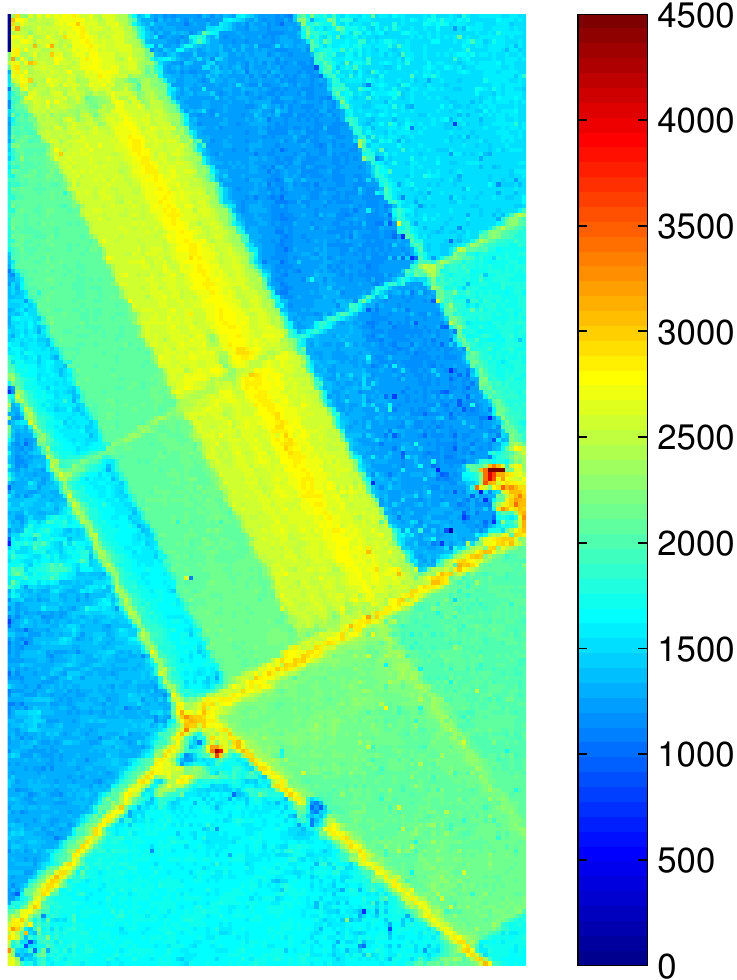}}&\subfloat[residual error]{\includegraphics[width=0.25\textwidth]{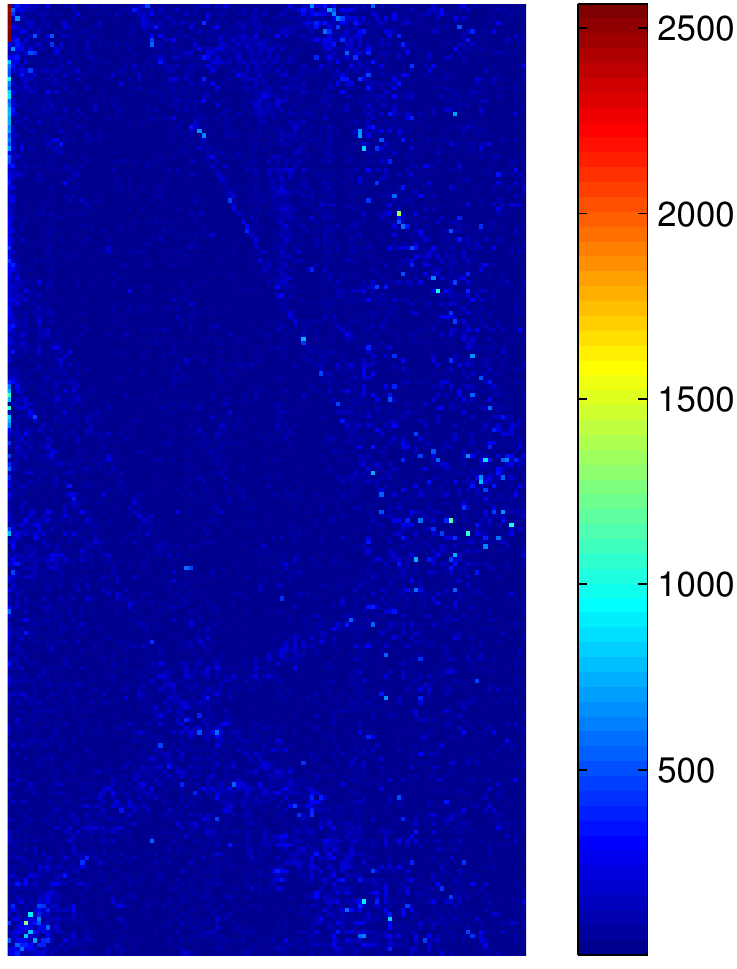}}
\end{tabular}
\end{tabular}}
\label{fig:4}
\caption{Reconstruction of Salinas Valley HSI using OVBSL, for different fractions of observed entries $\pi$.}
\end{figure}
\begin{figure}[t]
\centering
\begin{tabular}{c}
 \begin{tabular}{c c c c c c c}
 \includegraphics[width=0.05\textwidth,height=0.05\textwidth]{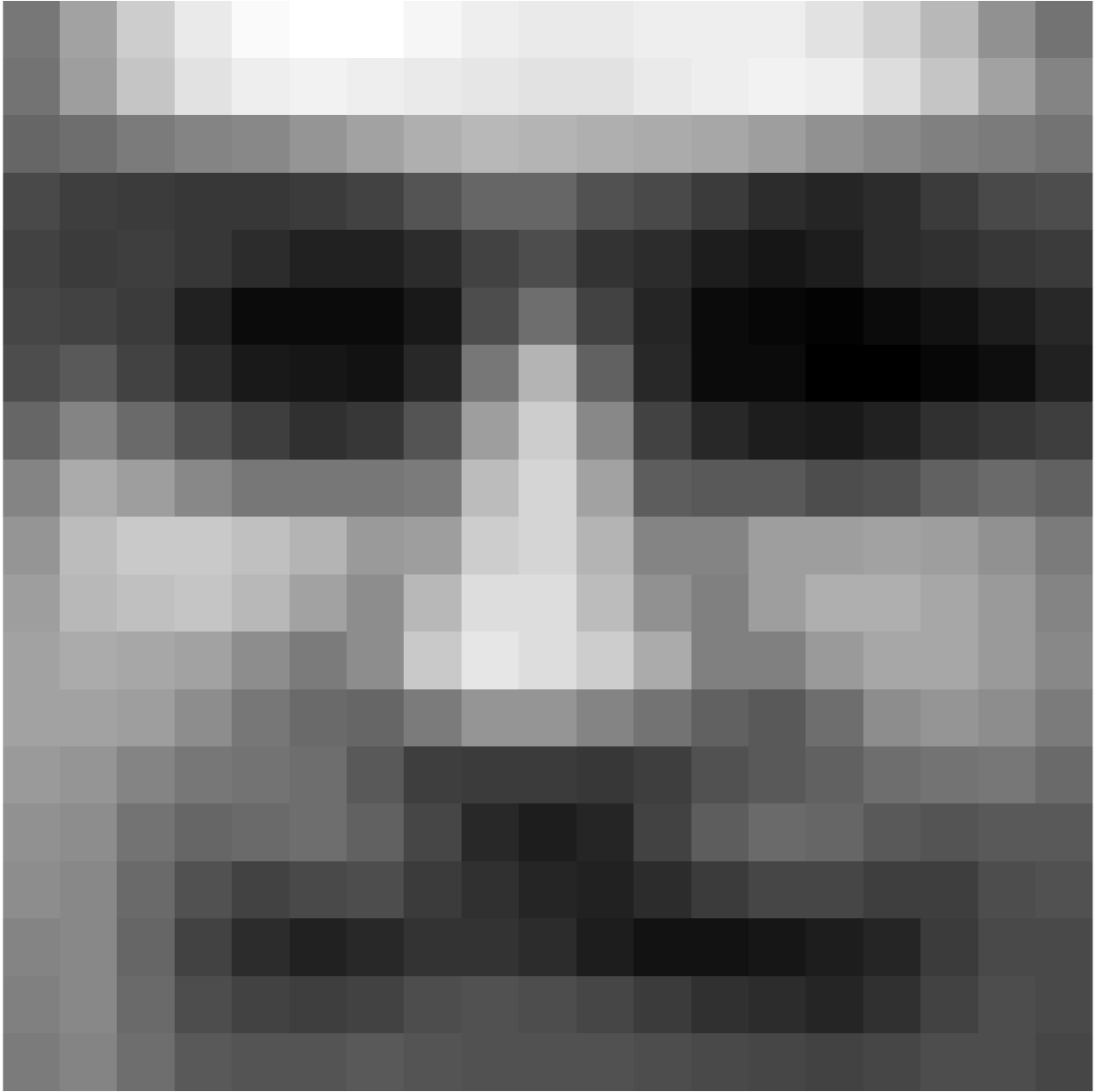}\hspace{-1cm}&\includegraphics[width=0.05\textwidth,height=0.05\textwidth]{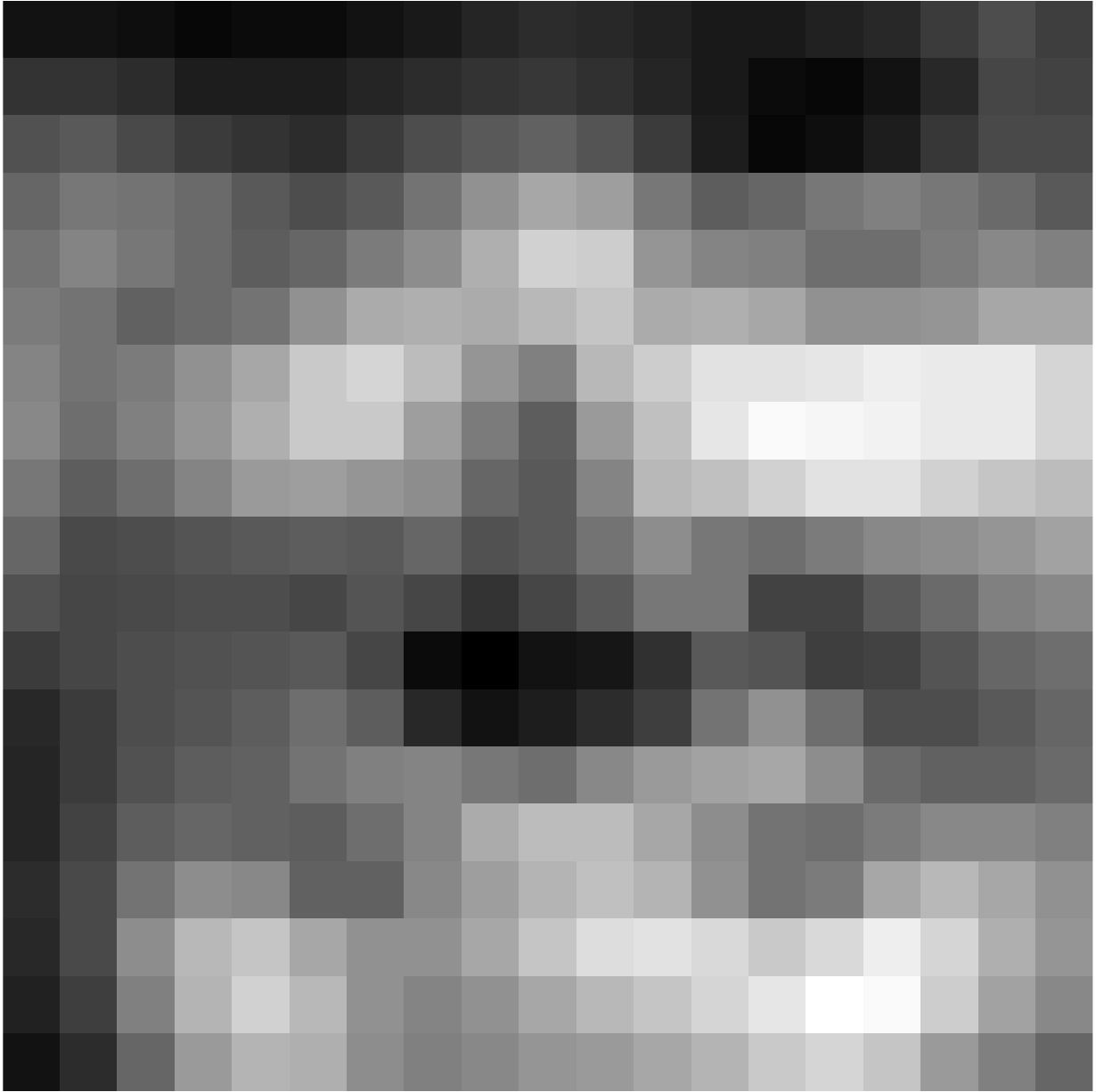}&\includegraphics[width=0.05\textwidth,height=0.05\textwidth]{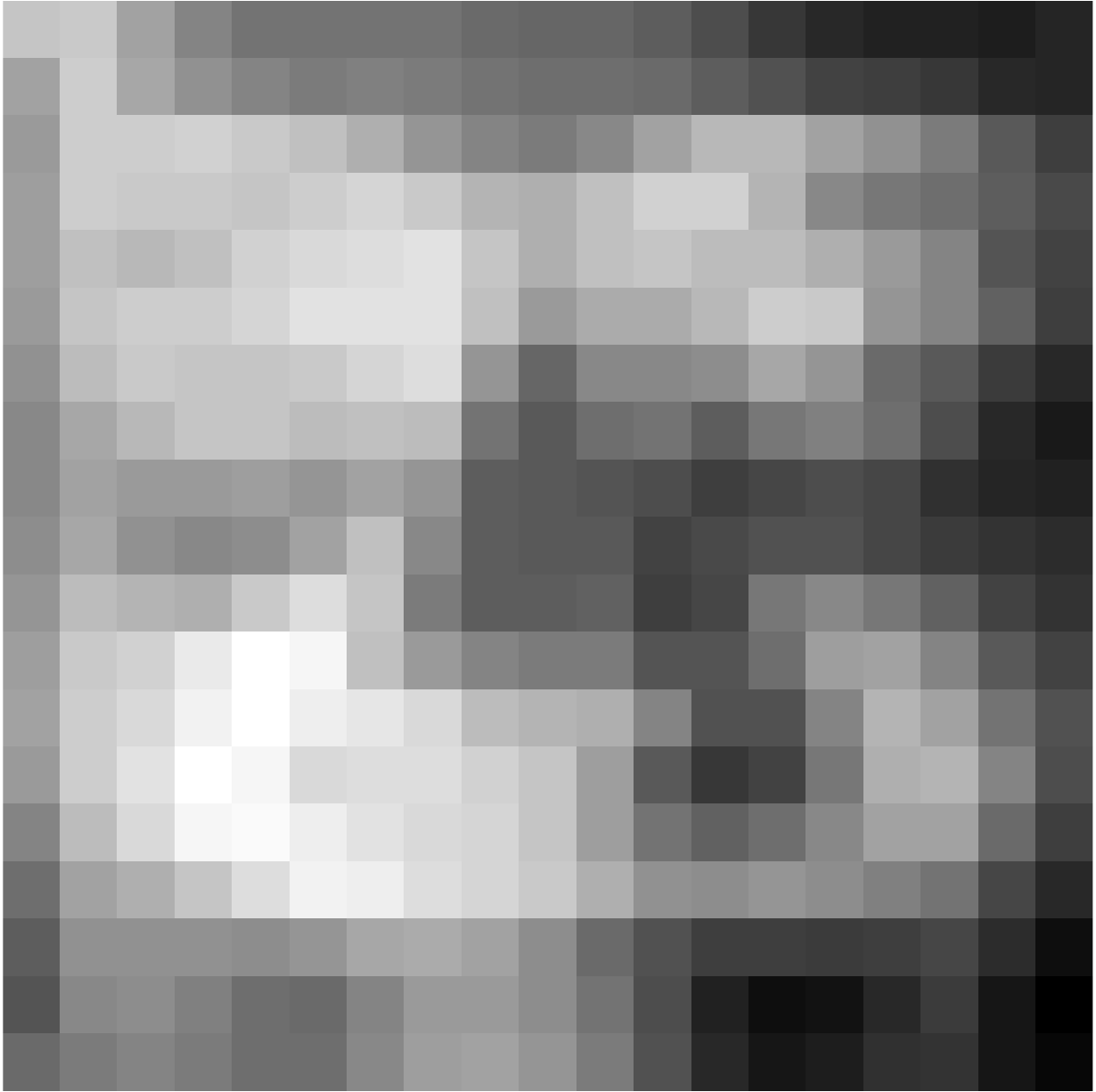}&\includegraphics[width=0.05\textwidth,height=0.05\textwidth]{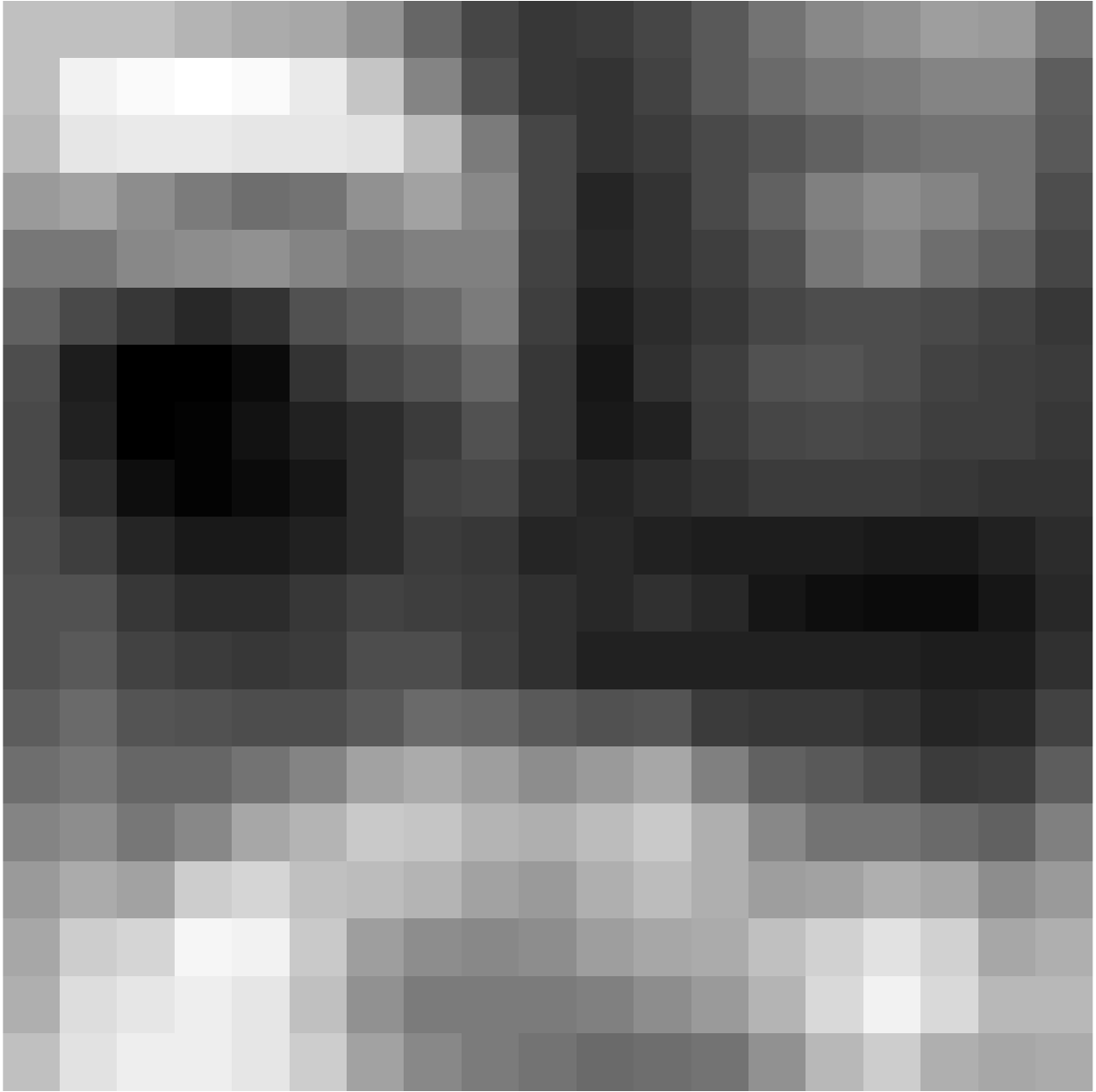}&\includegraphics[width=0.05\textwidth,height=0.05\textwidth]{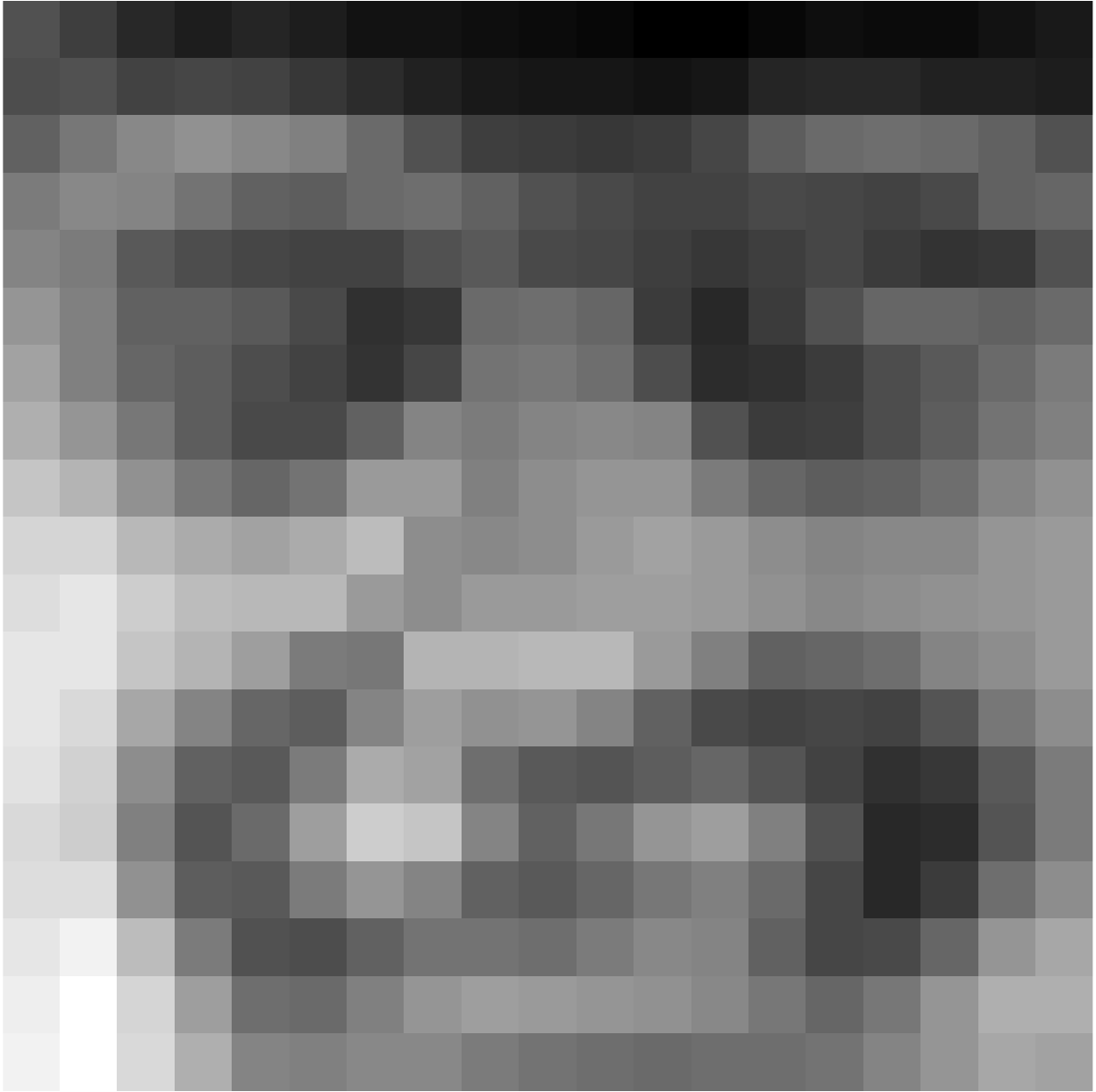} & \includegraphics[width=0.05\textwidth,height=0.05\textwidth]{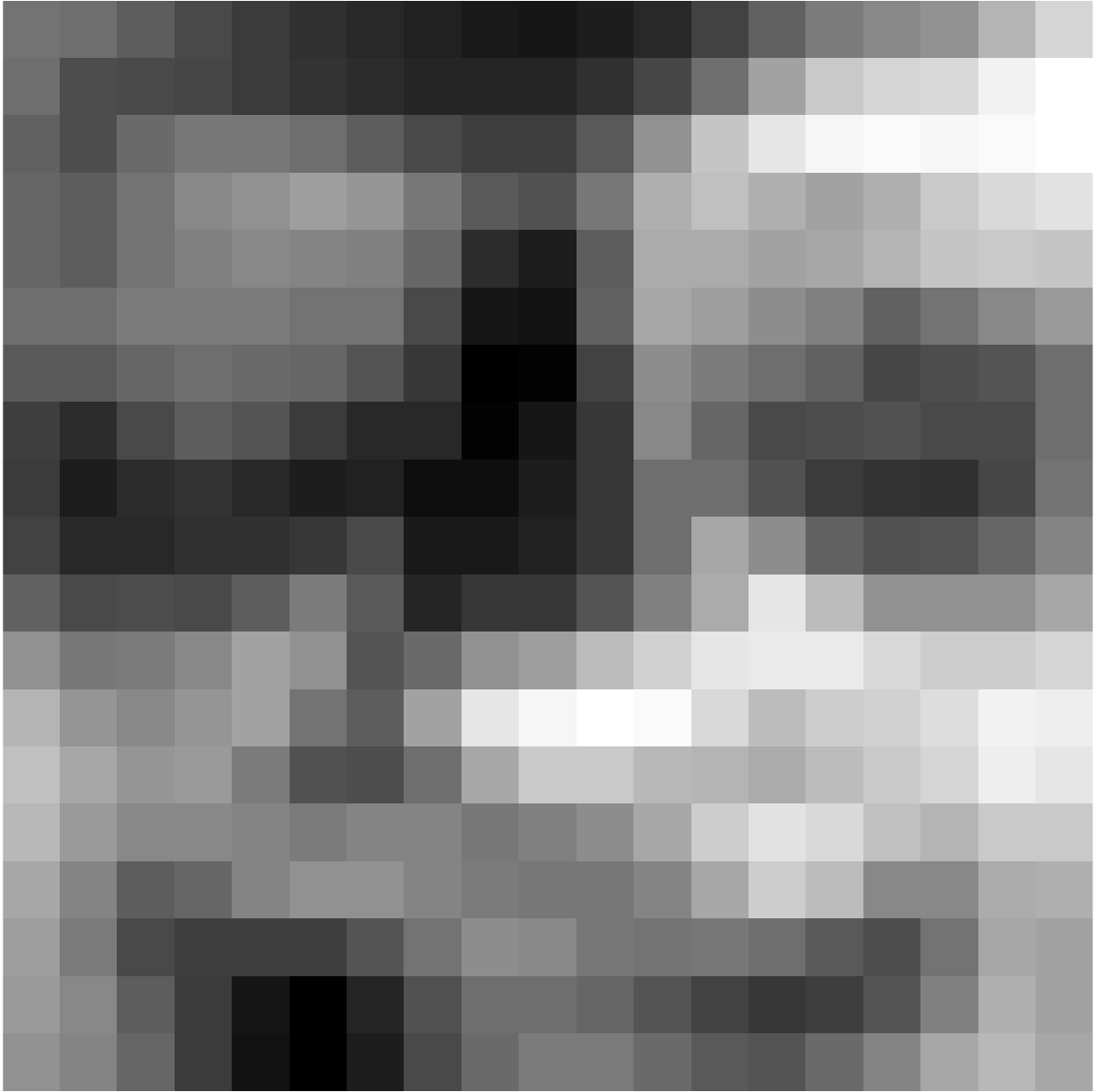}&\includegraphics[width=0.05\textwidth,height=0.05\textwidth]{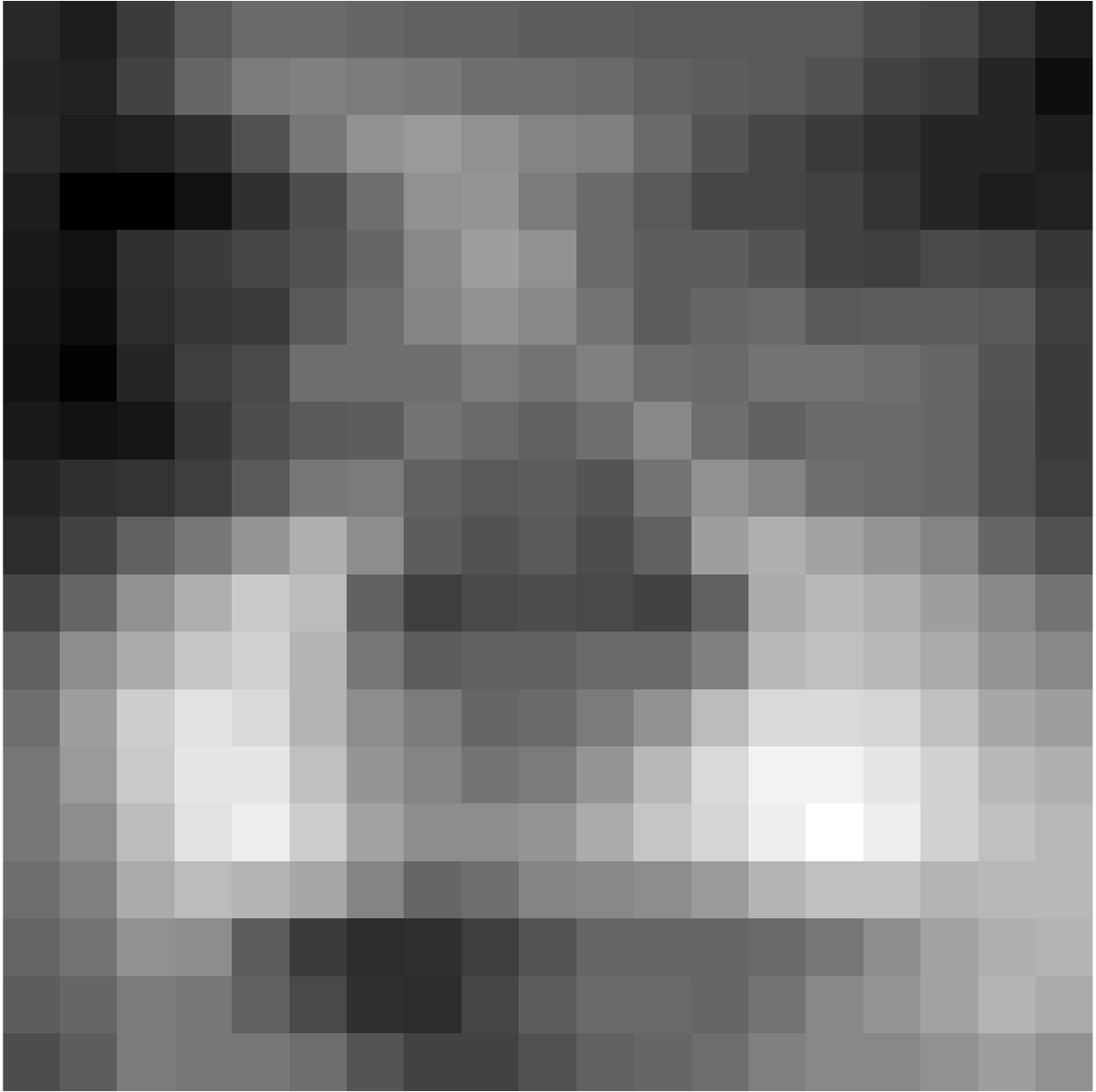} \\
 \includegraphics[width=0.05\textwidth,height=0.05\textwidth]{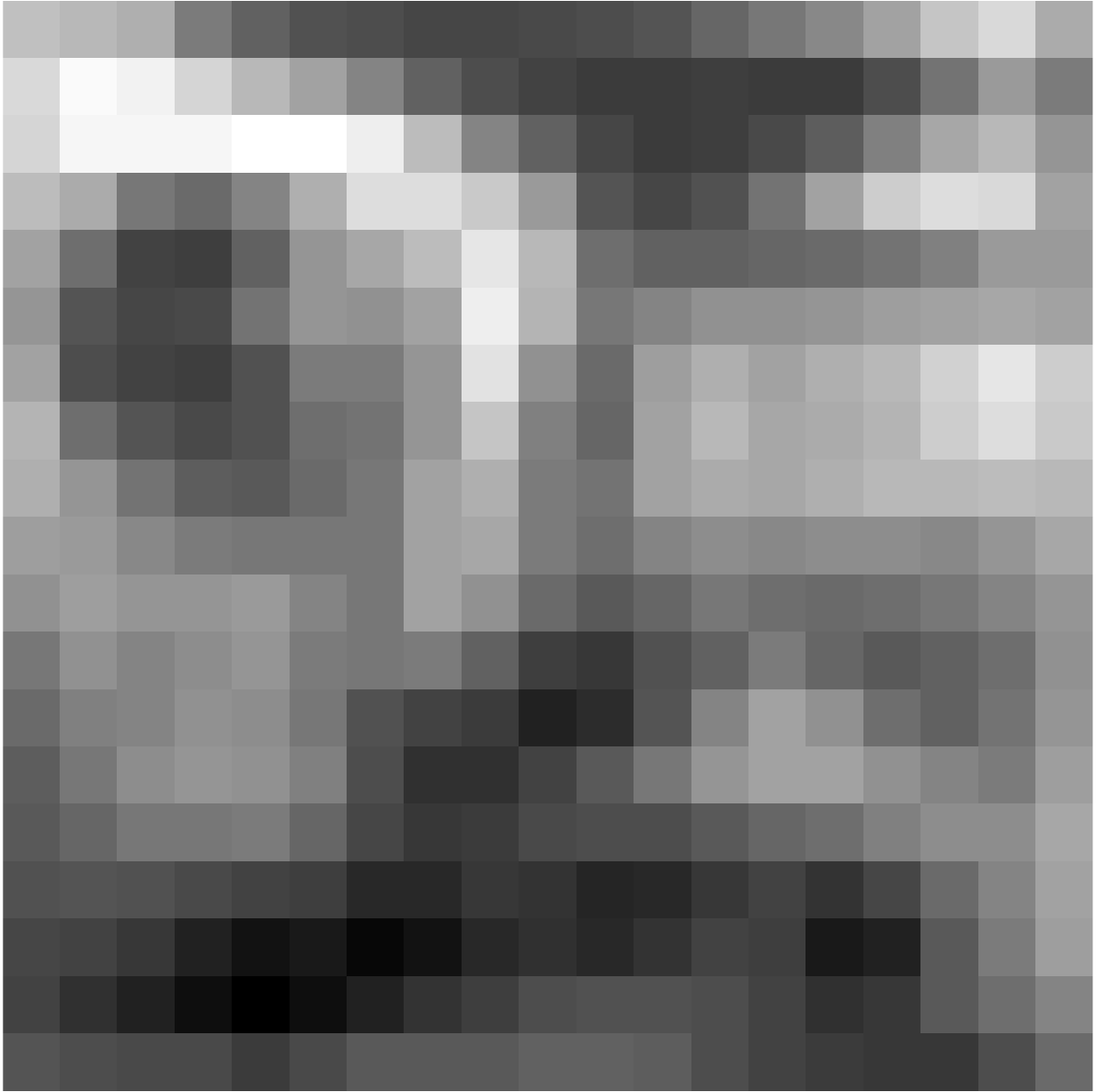}&\includegraphics[width=0.05\textwidth,height=0.05\textwidth]{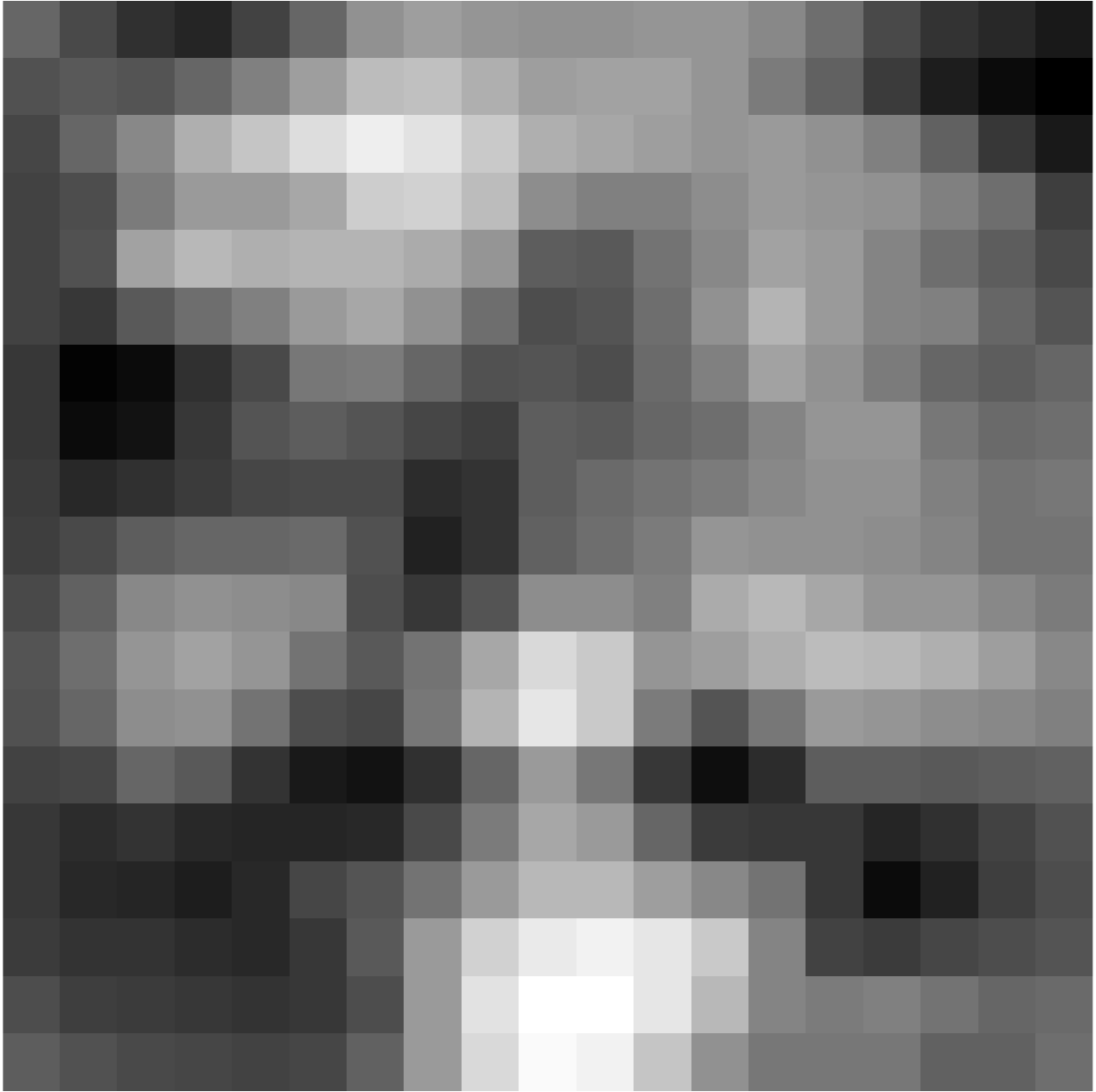}&\includegraphics[width=0.05\textwidth,height=0.05\textwidth]{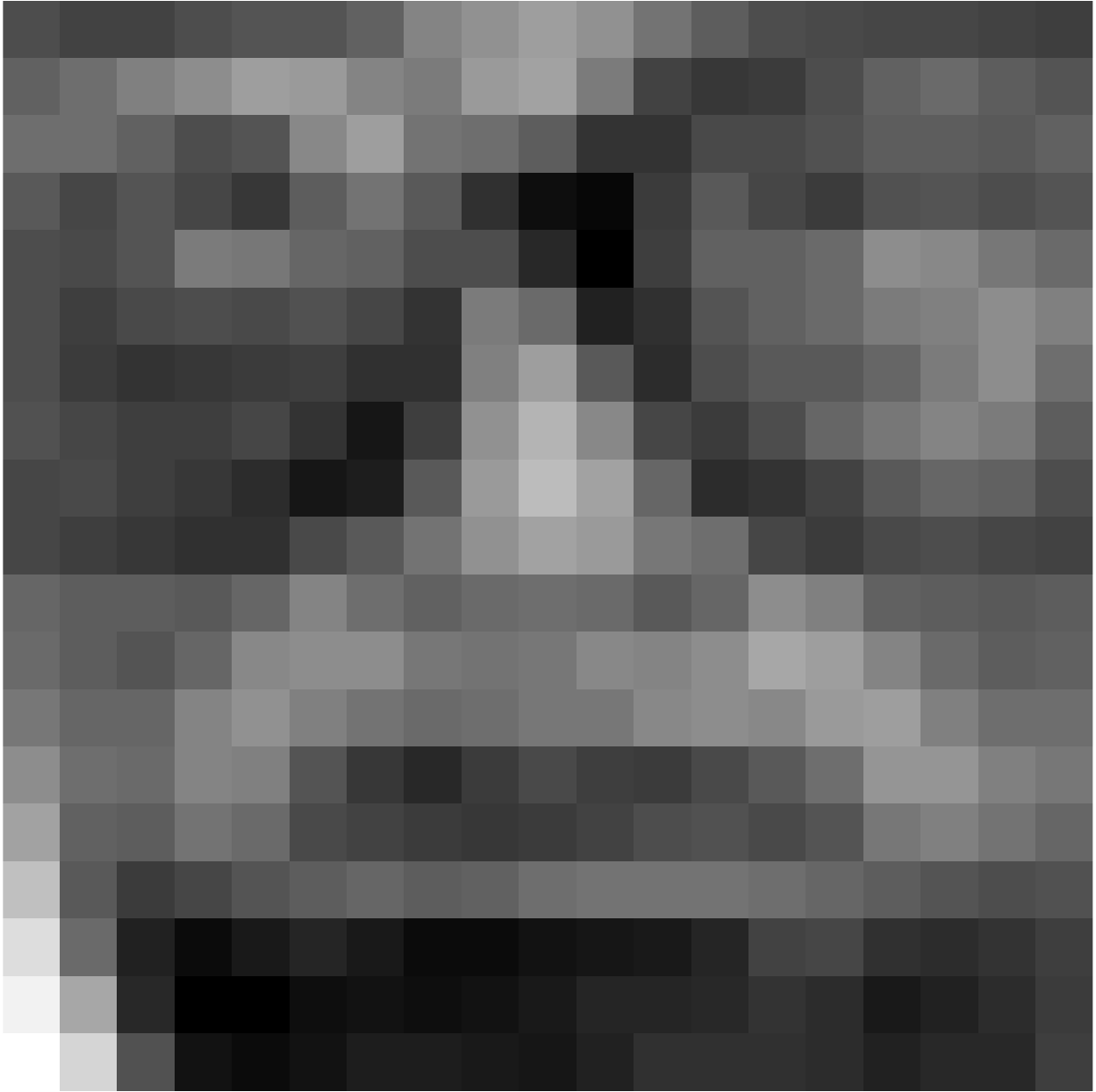} &  \includegraphics[width=0.05\textwidth,height=0.05\textwidth]{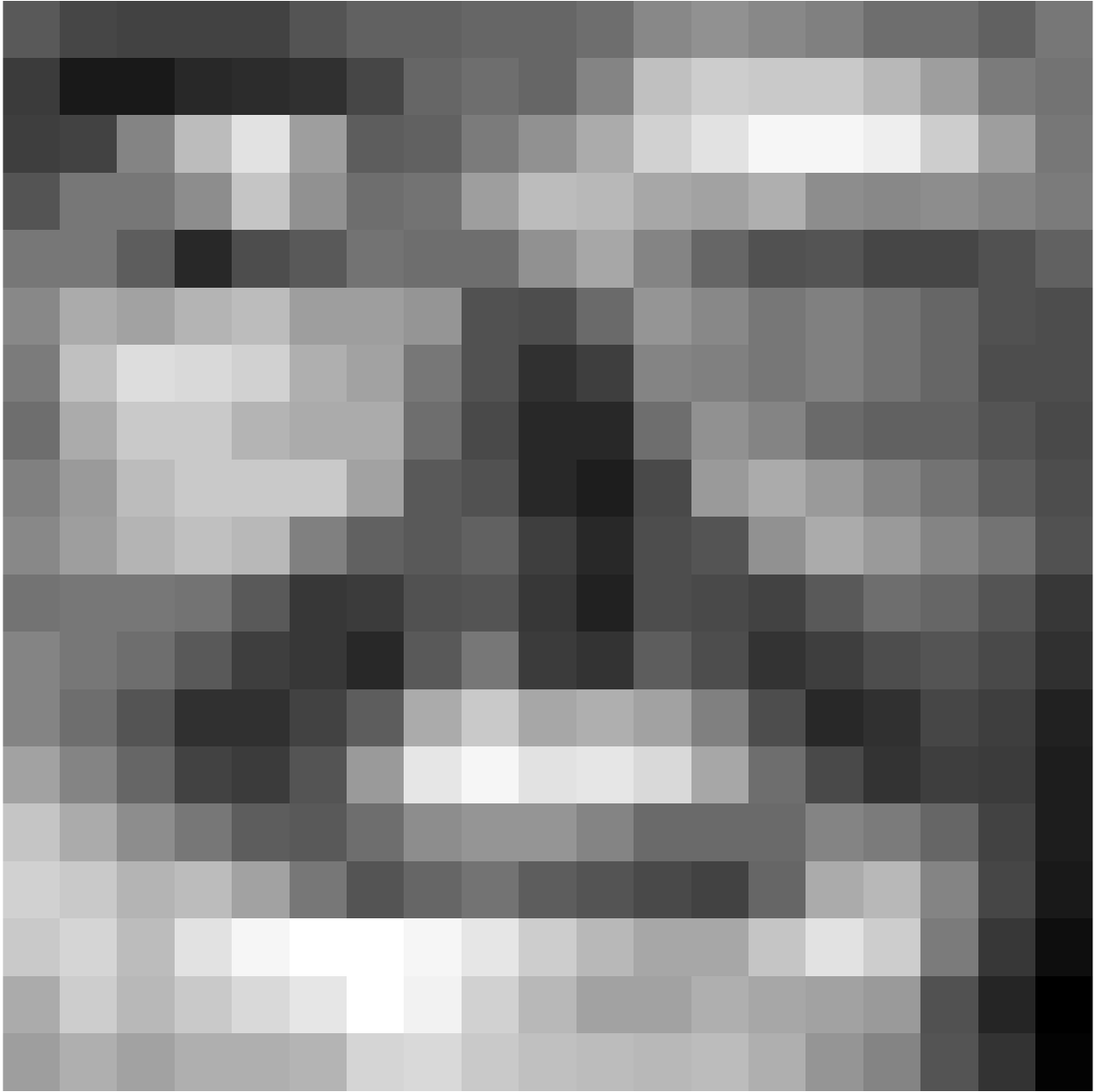}&\includegraphics[width=0.05\textwidth,height=0.05\textwidth]{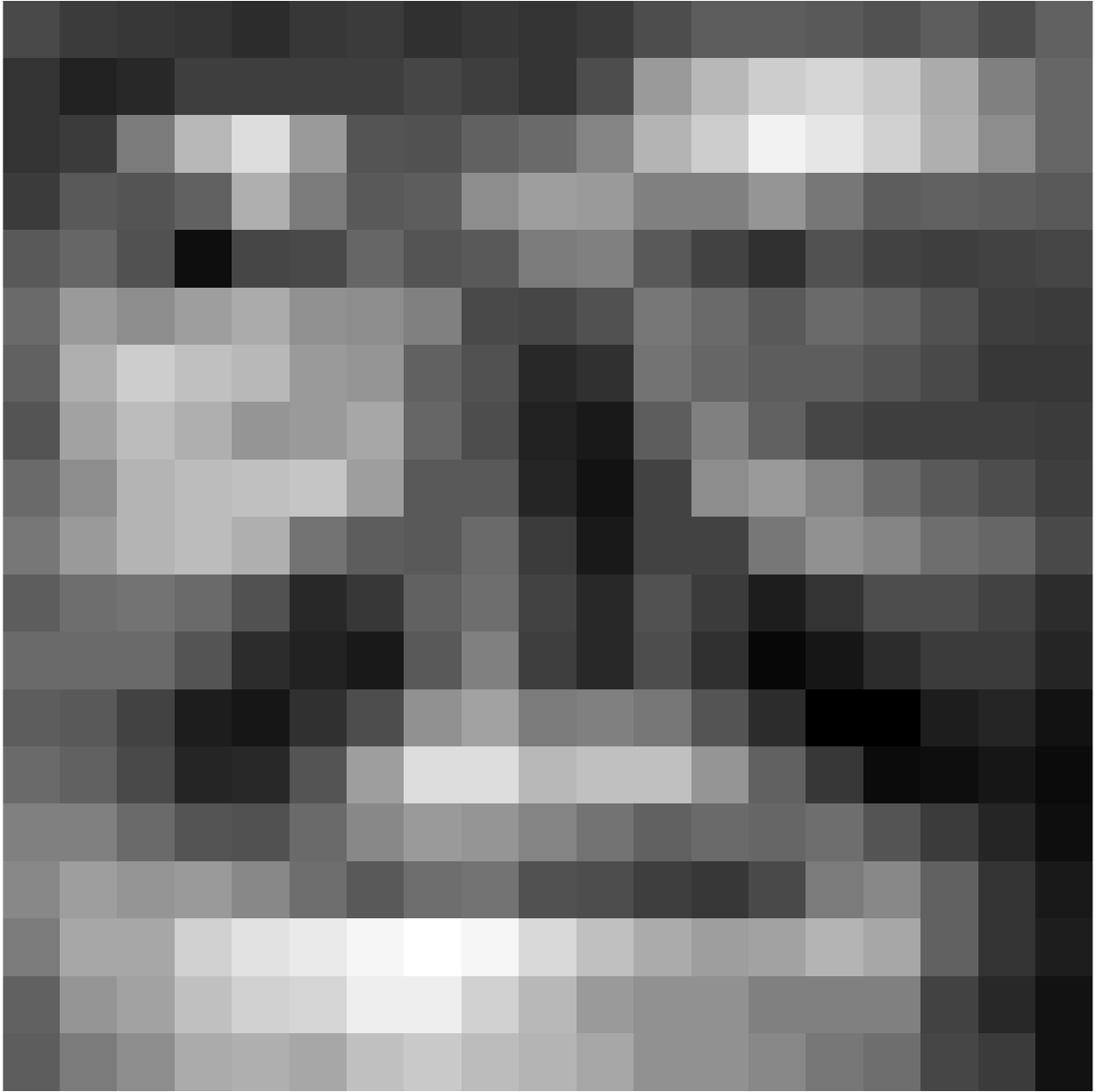}&\includegraphics[width=0.05\textwidth,height=0.05\textwidth]{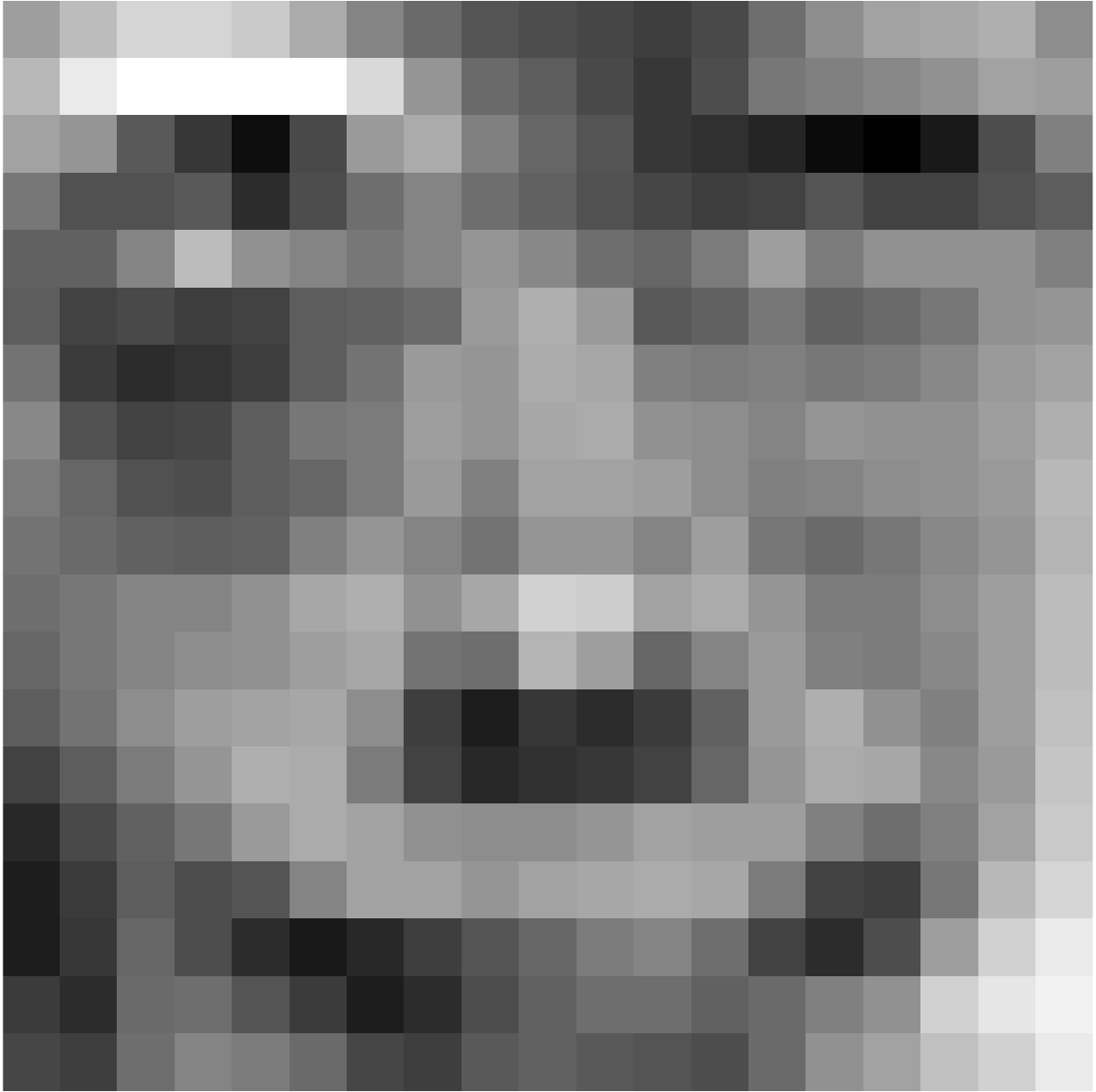}&\includegraphics[width=0.05\textwidth,height=0.05\textwidth]{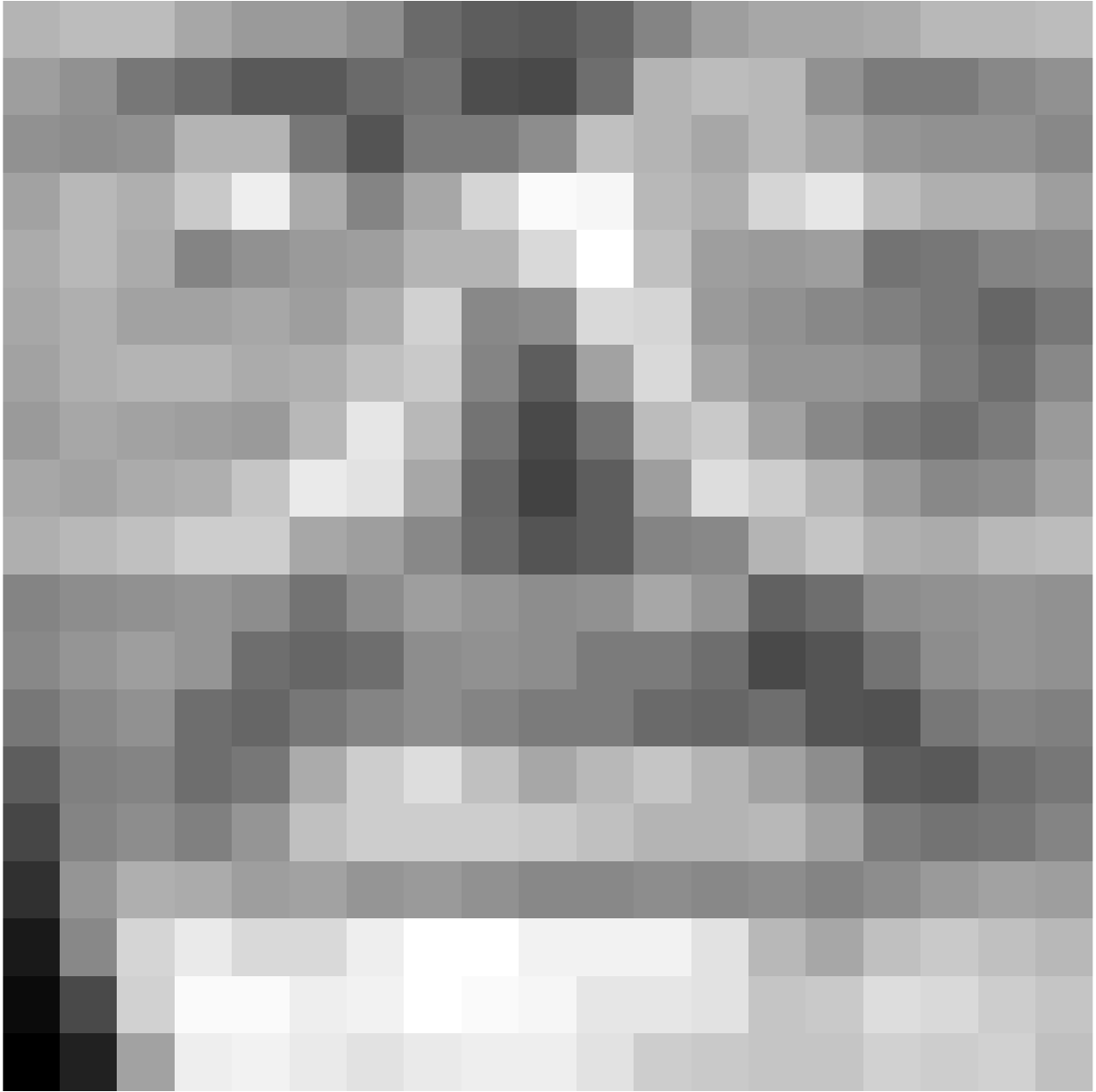}\\
 \includegraphics[width=0.05\textwidth,height=0.05\textwidth]{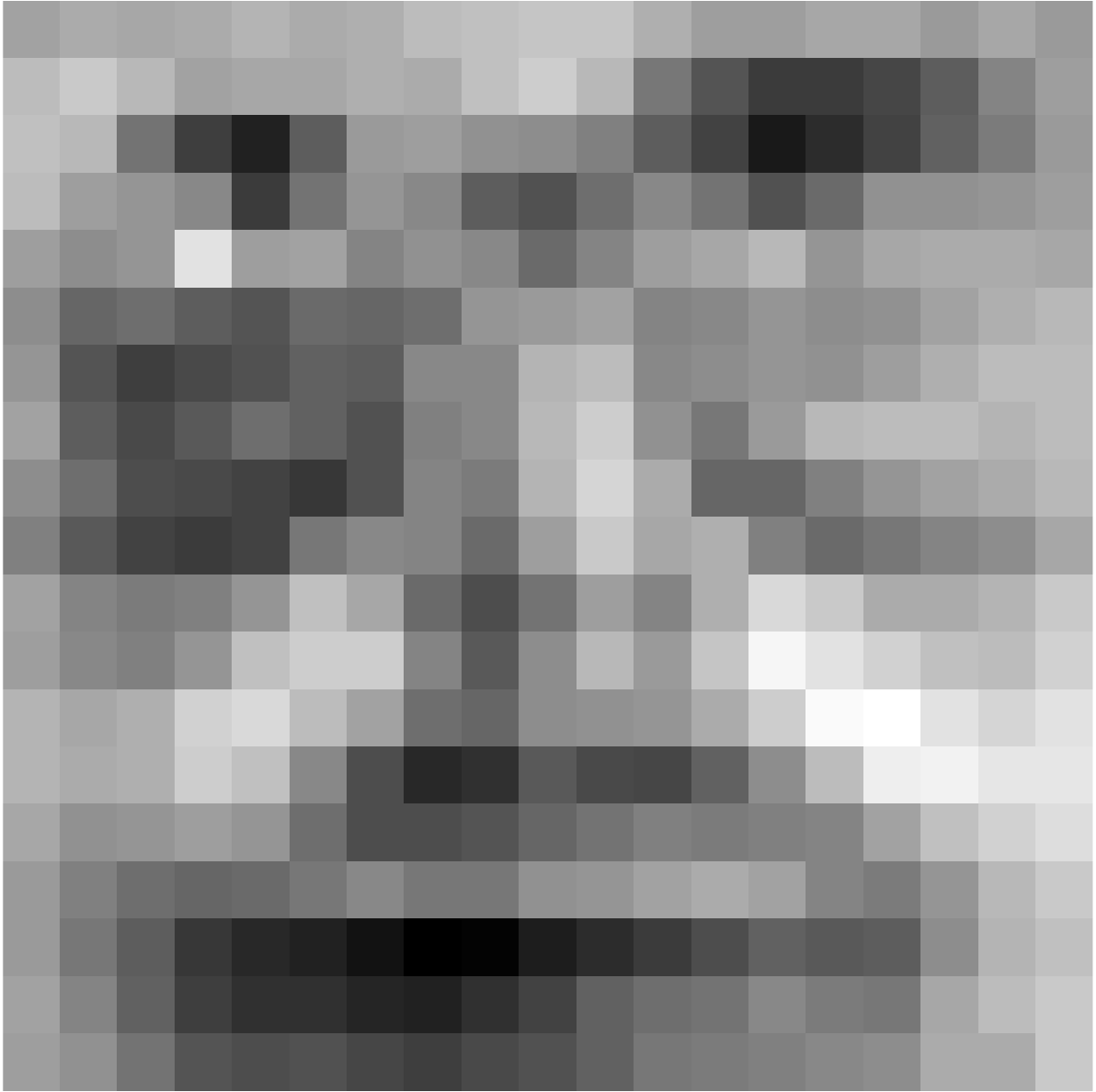}&\includegraphics[width=0.05\textwidth,height=0.05\textwidth]{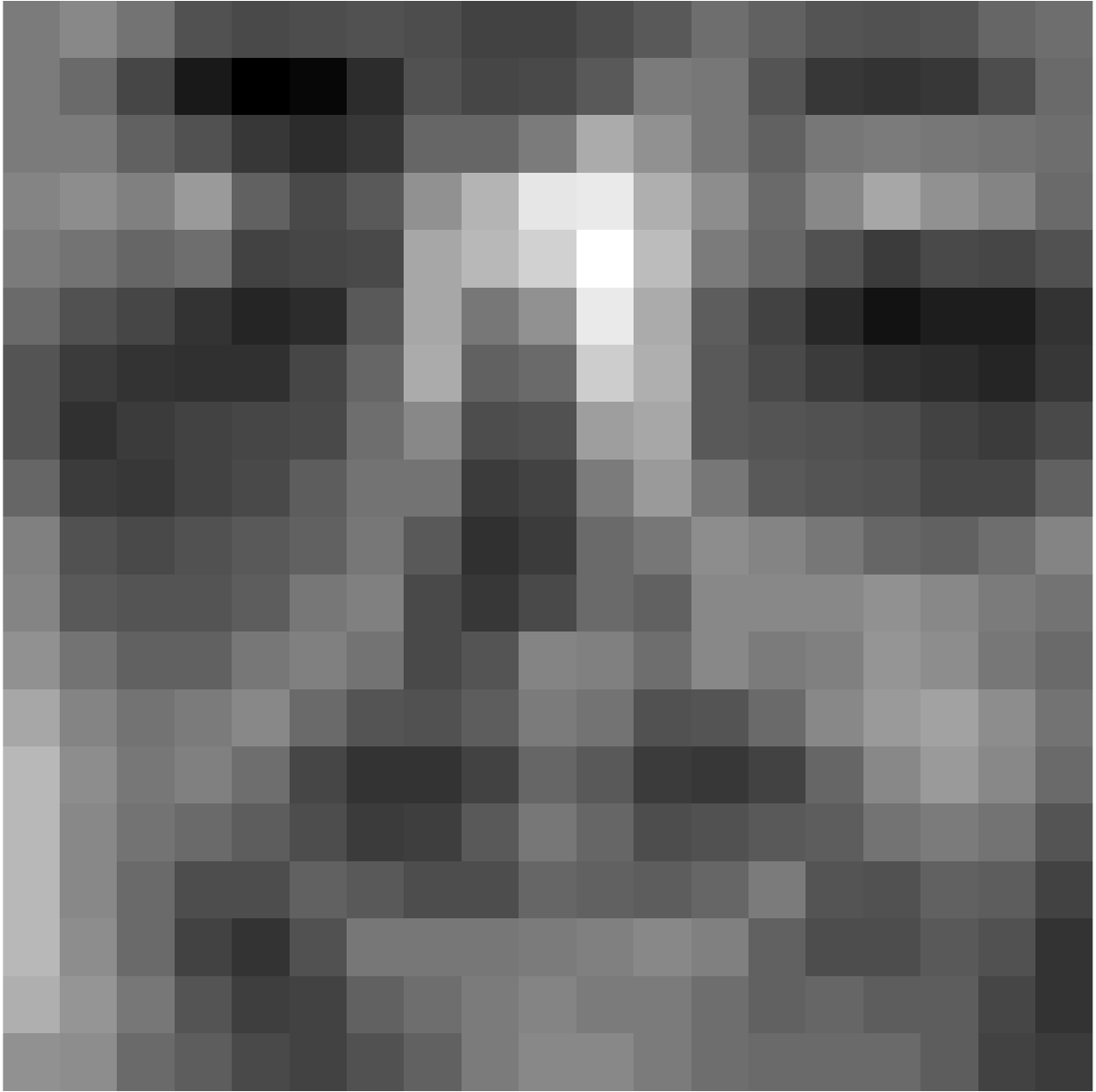}&\includegraphics[width=0.05\textwidth,height=0.05\textwidth]{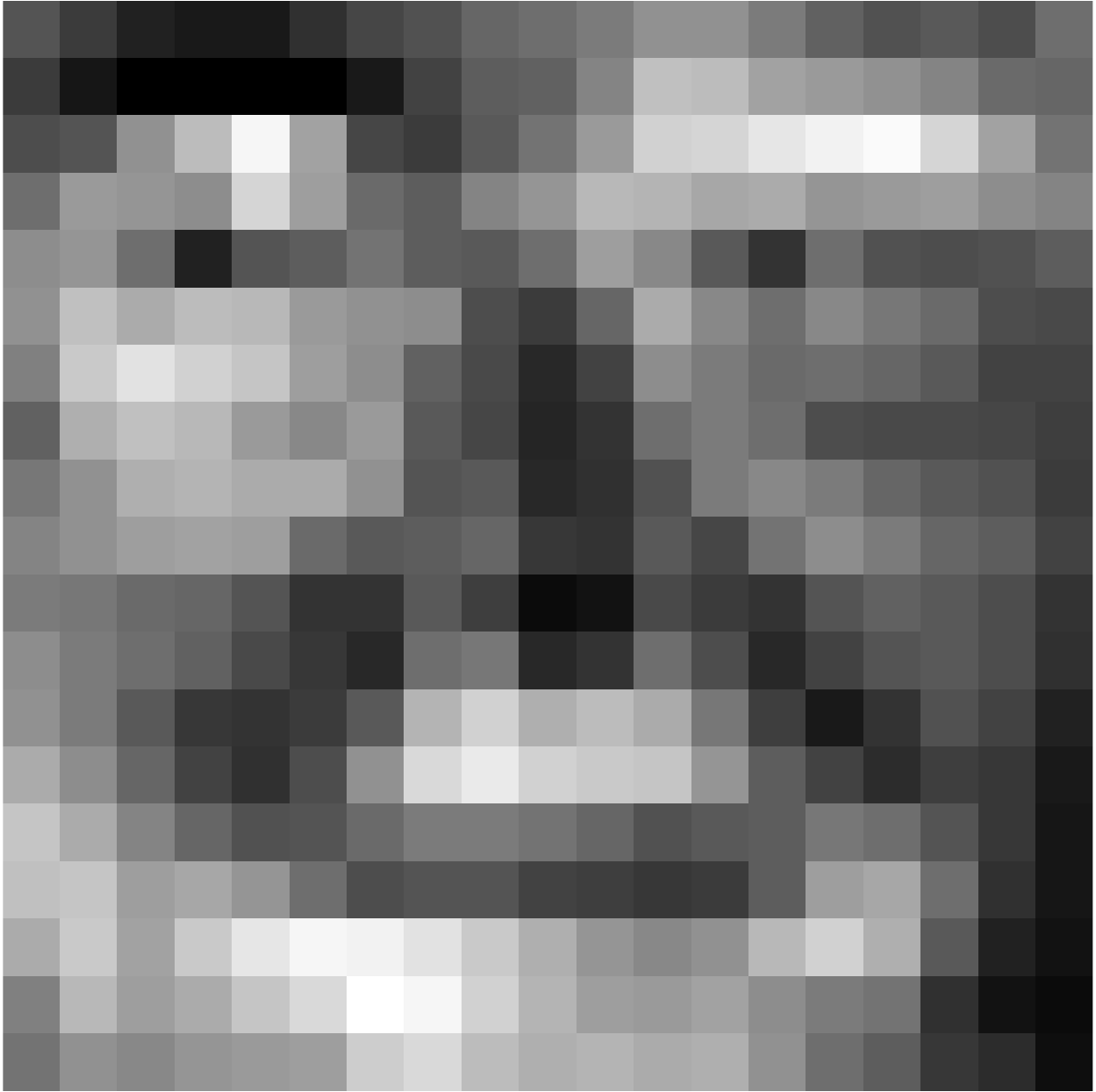}&\includegraphics[width=0.05\textwidth,height=0.05\textwidth]{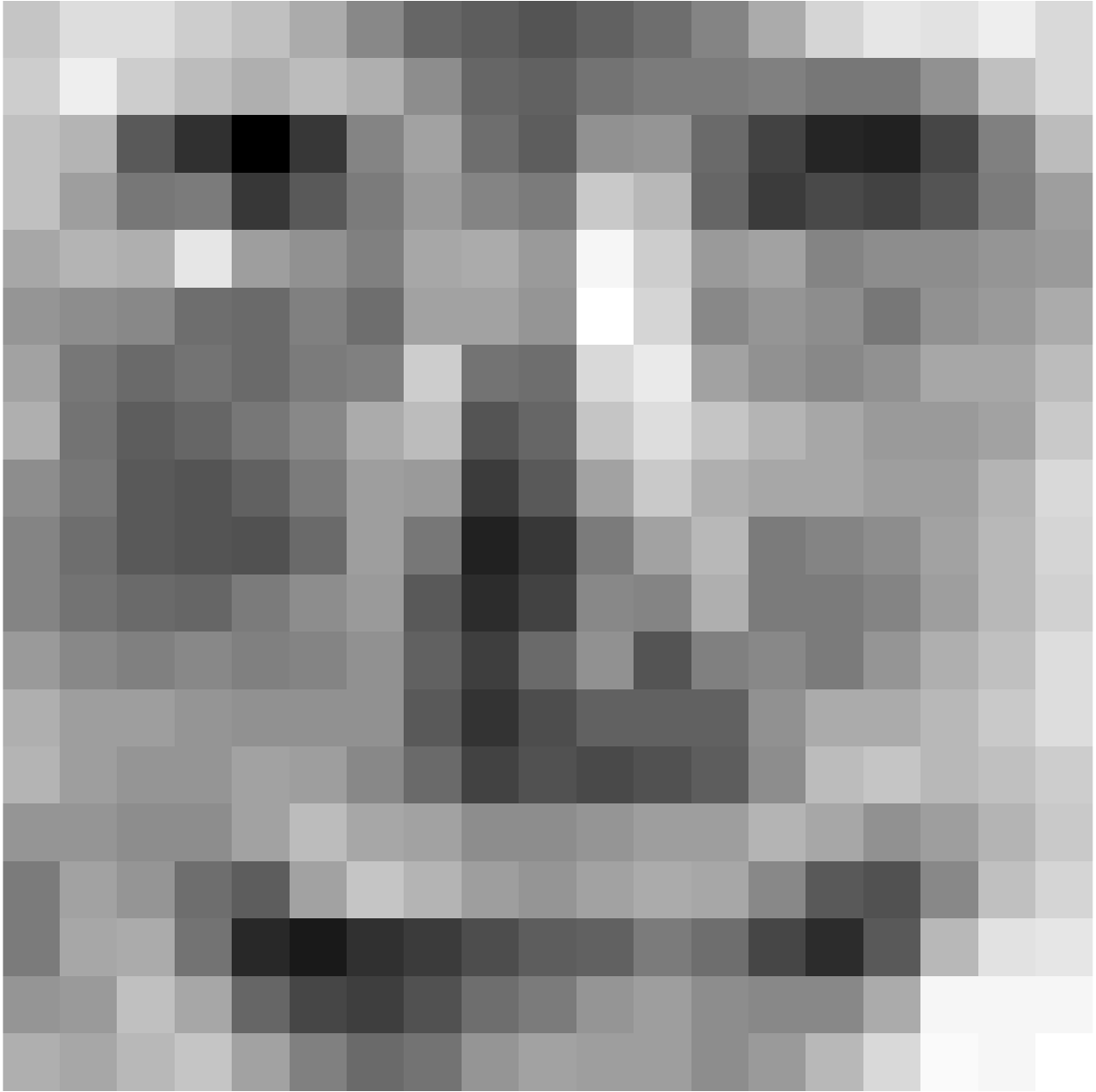}&\includegraphics[width=0.05\textwidth,height=0.05\textwidth]{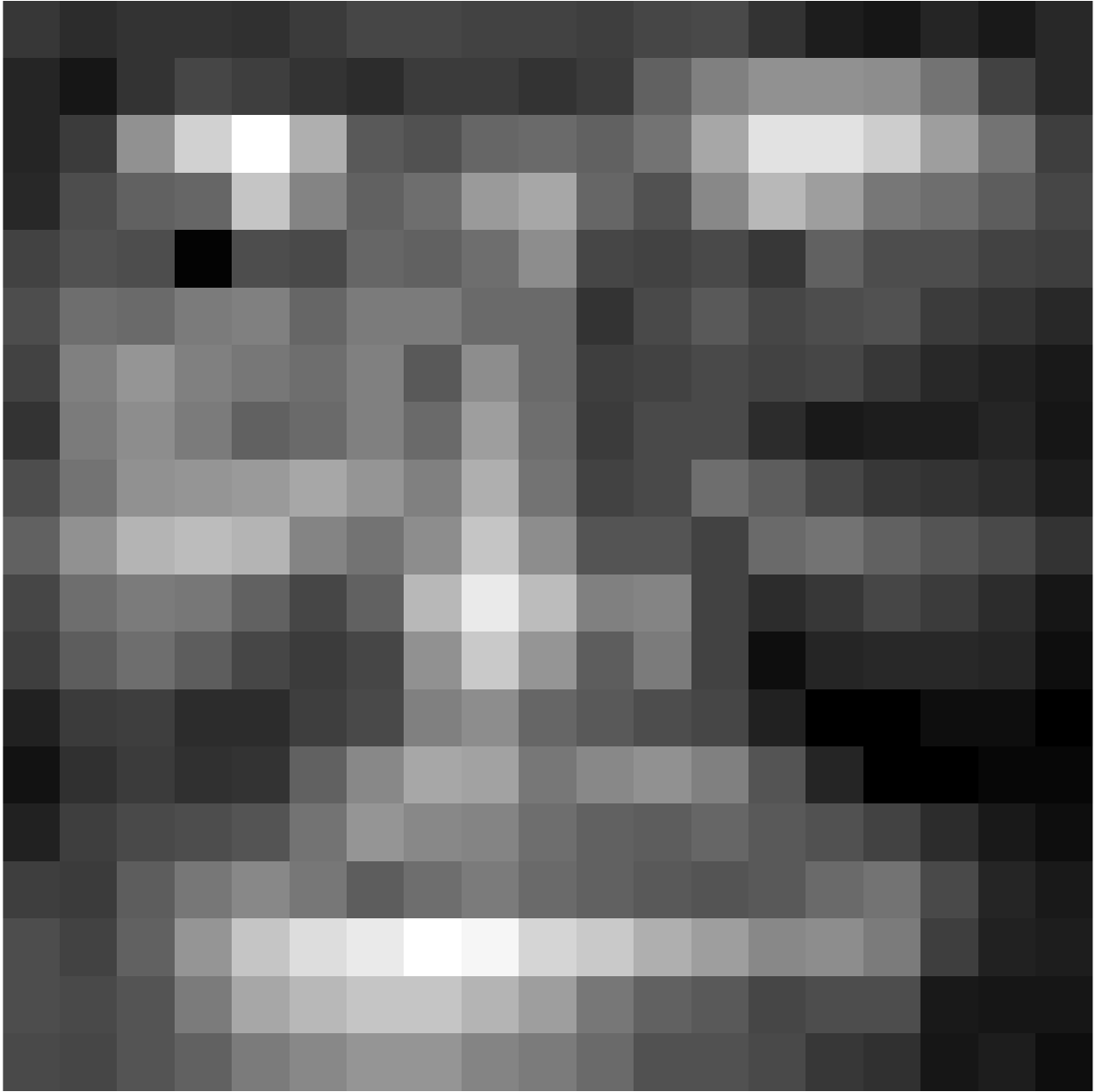}&\includegraphics[width=0.05\textwidth,height=0.05\textwidth]{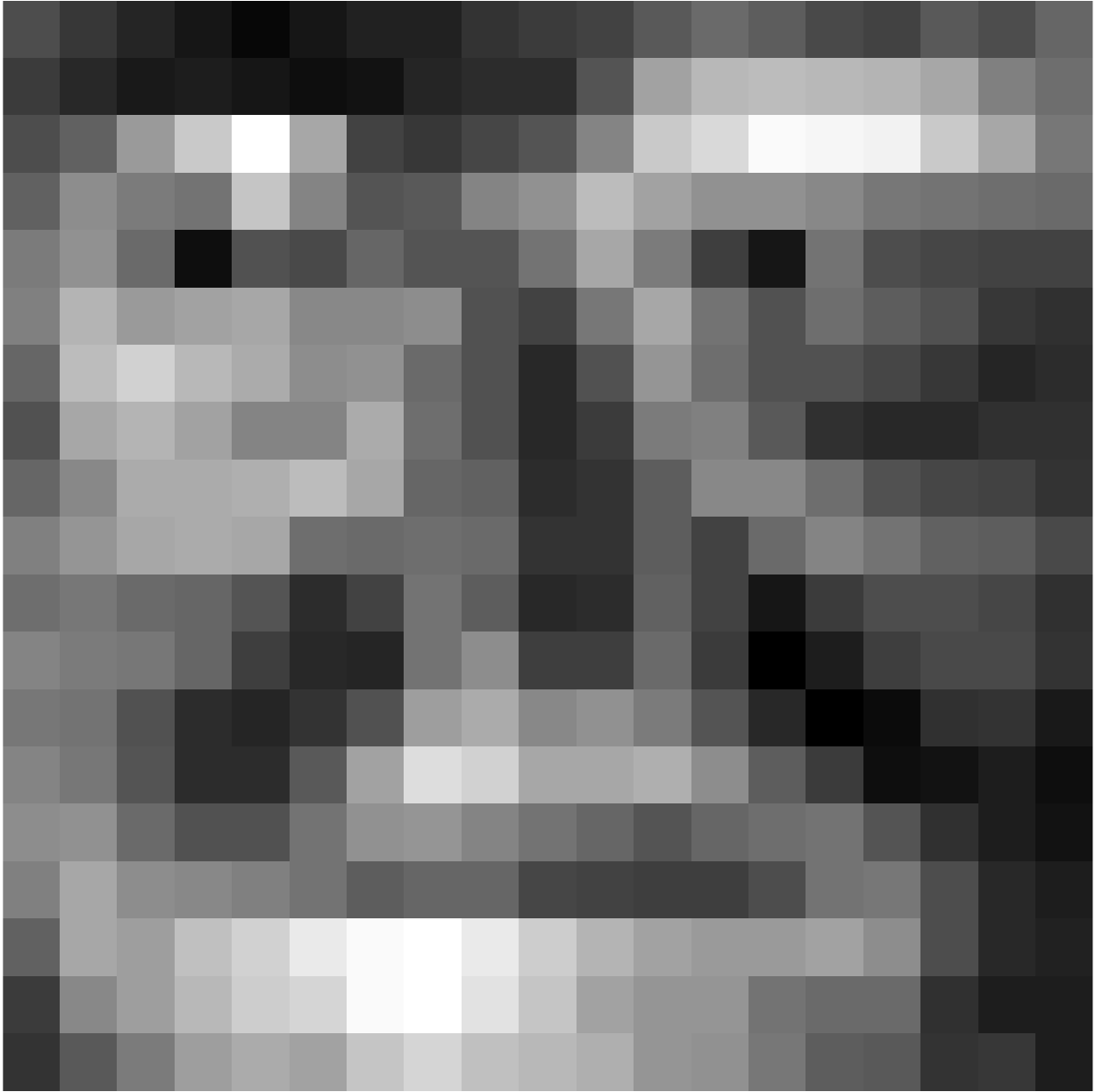}&\includegraphics[width=0.05\textwidth,height=0.05\textwidth]{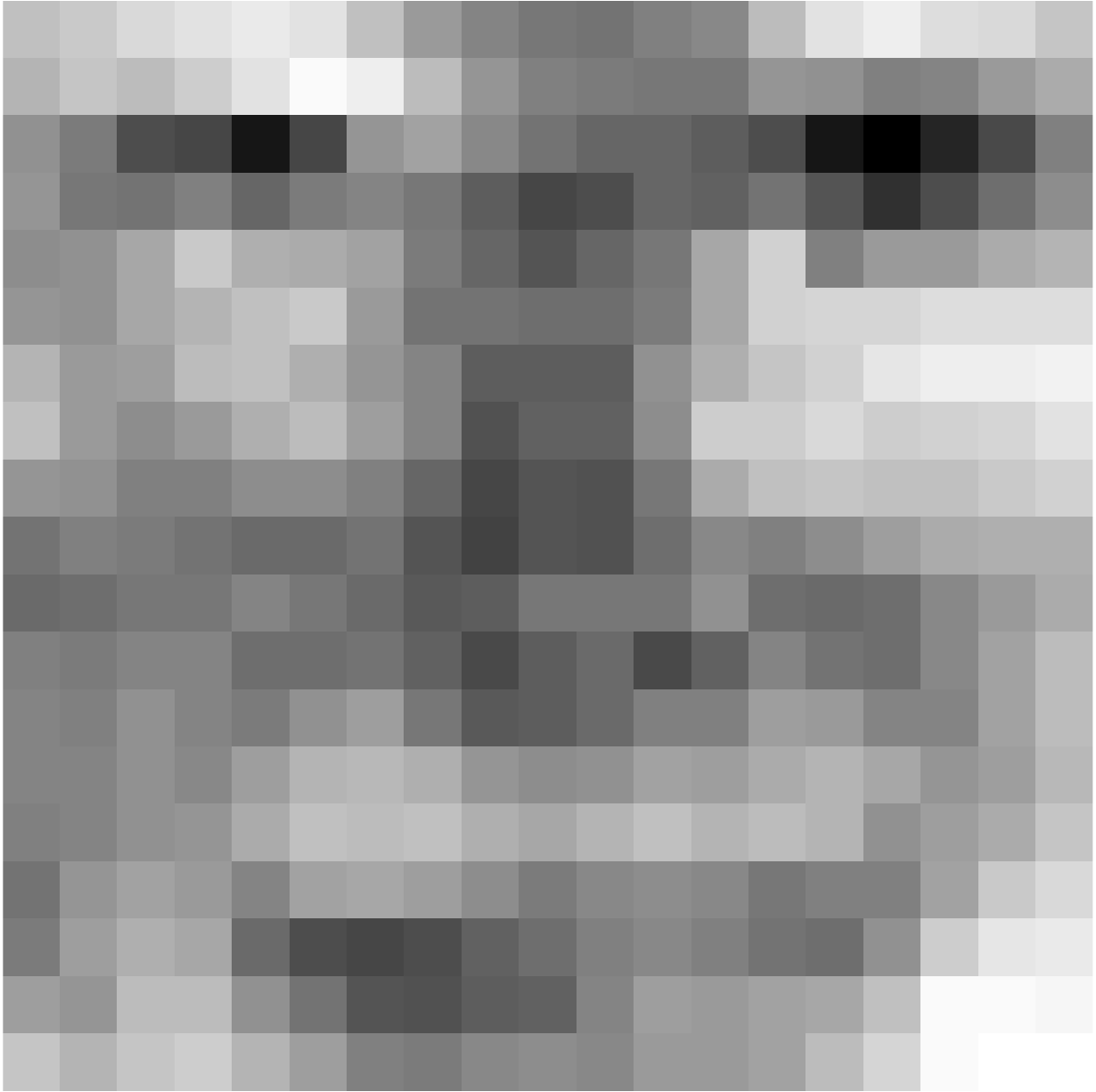} 
 \end{tabular} \\
 \small a) non-sparse OVBSL \\
  \begin{tabular}{c c c c c c c}
  \includegraphics[width=0.05\textwidth,height=0.05\textwidth]{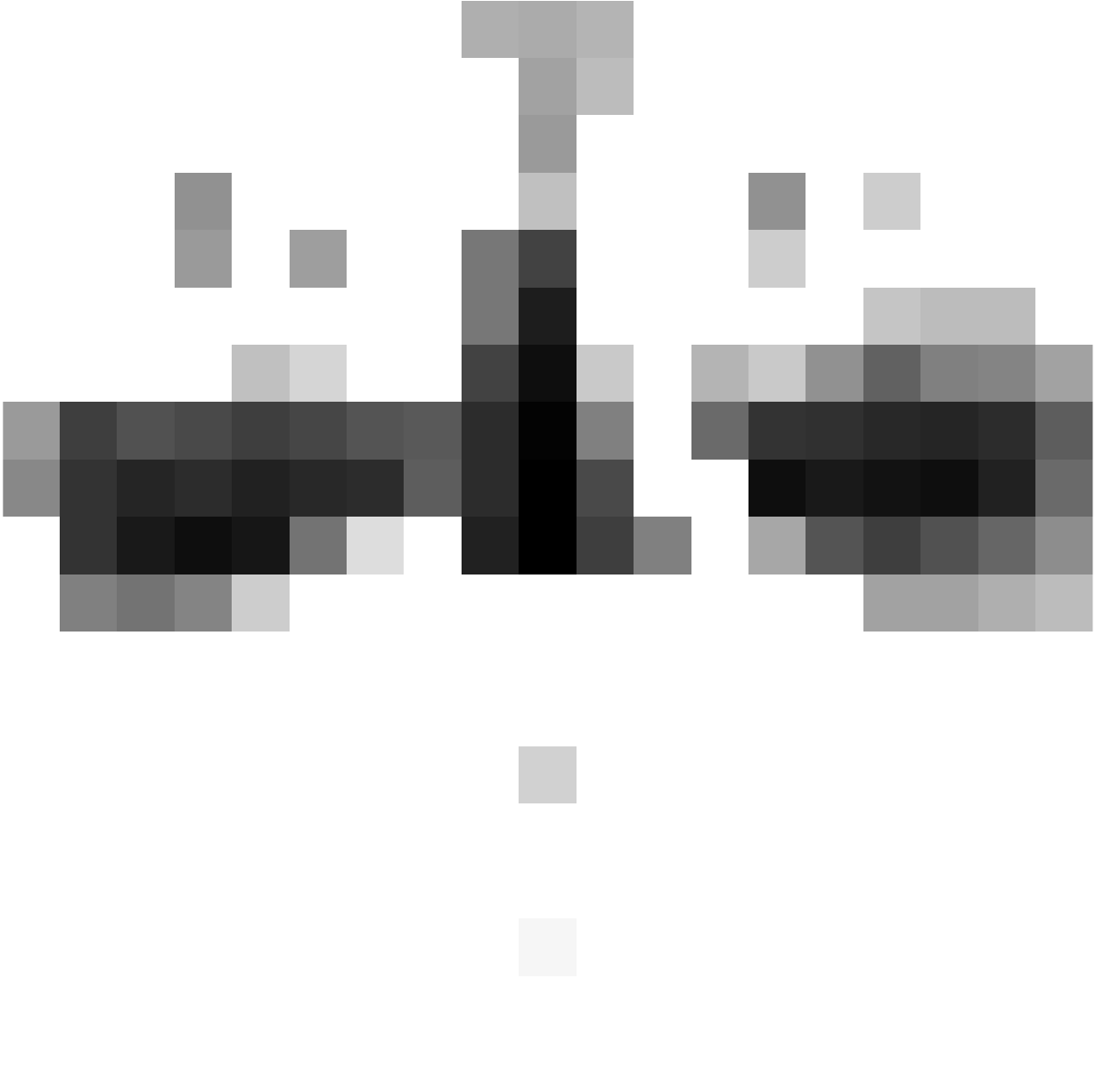}&\includegraphics[width=0.05\textwidth,height=0.05\textwidth]{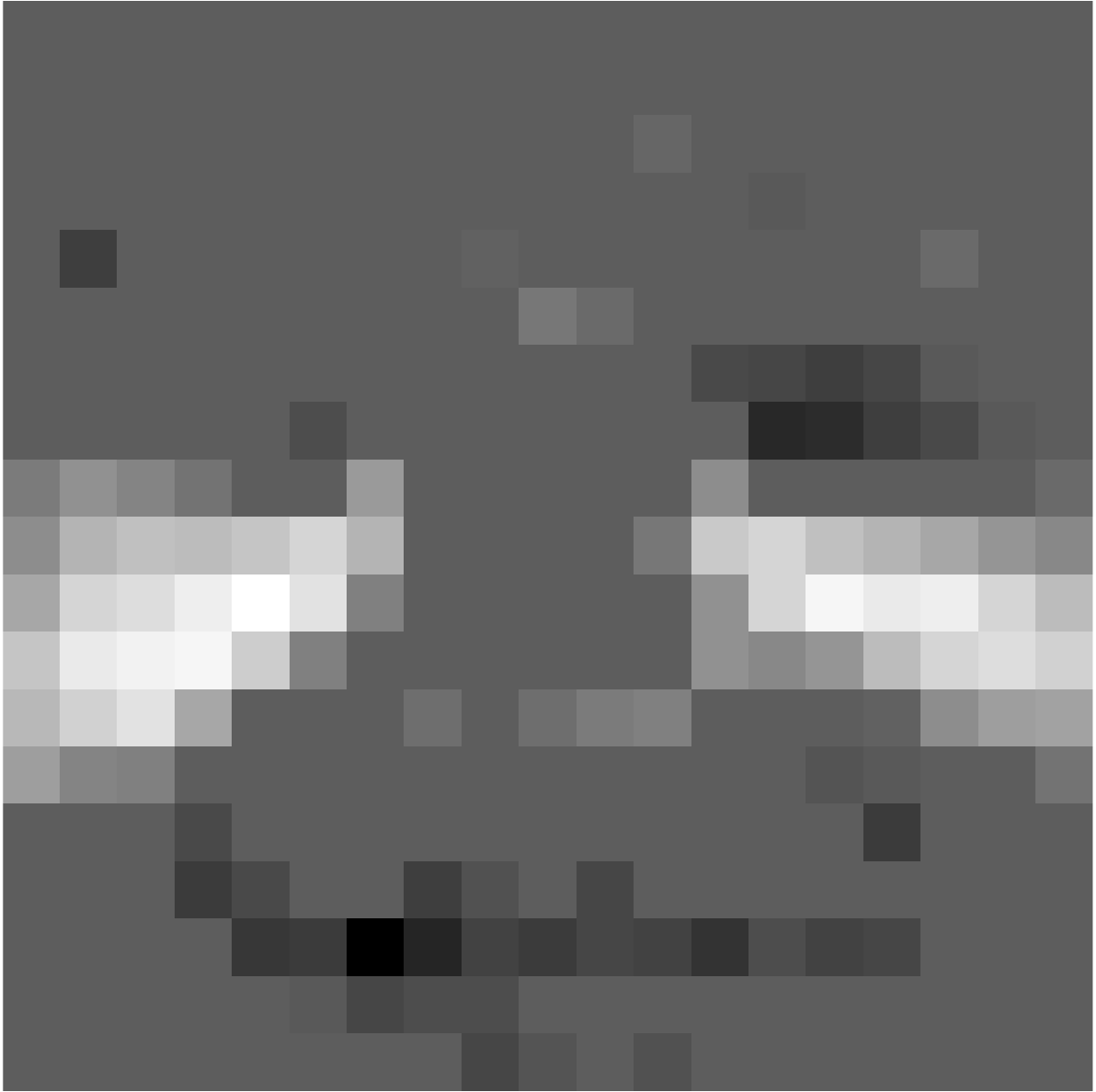}&\includegraphics[width=0.05\textwidth,height=0.05\textwidth]{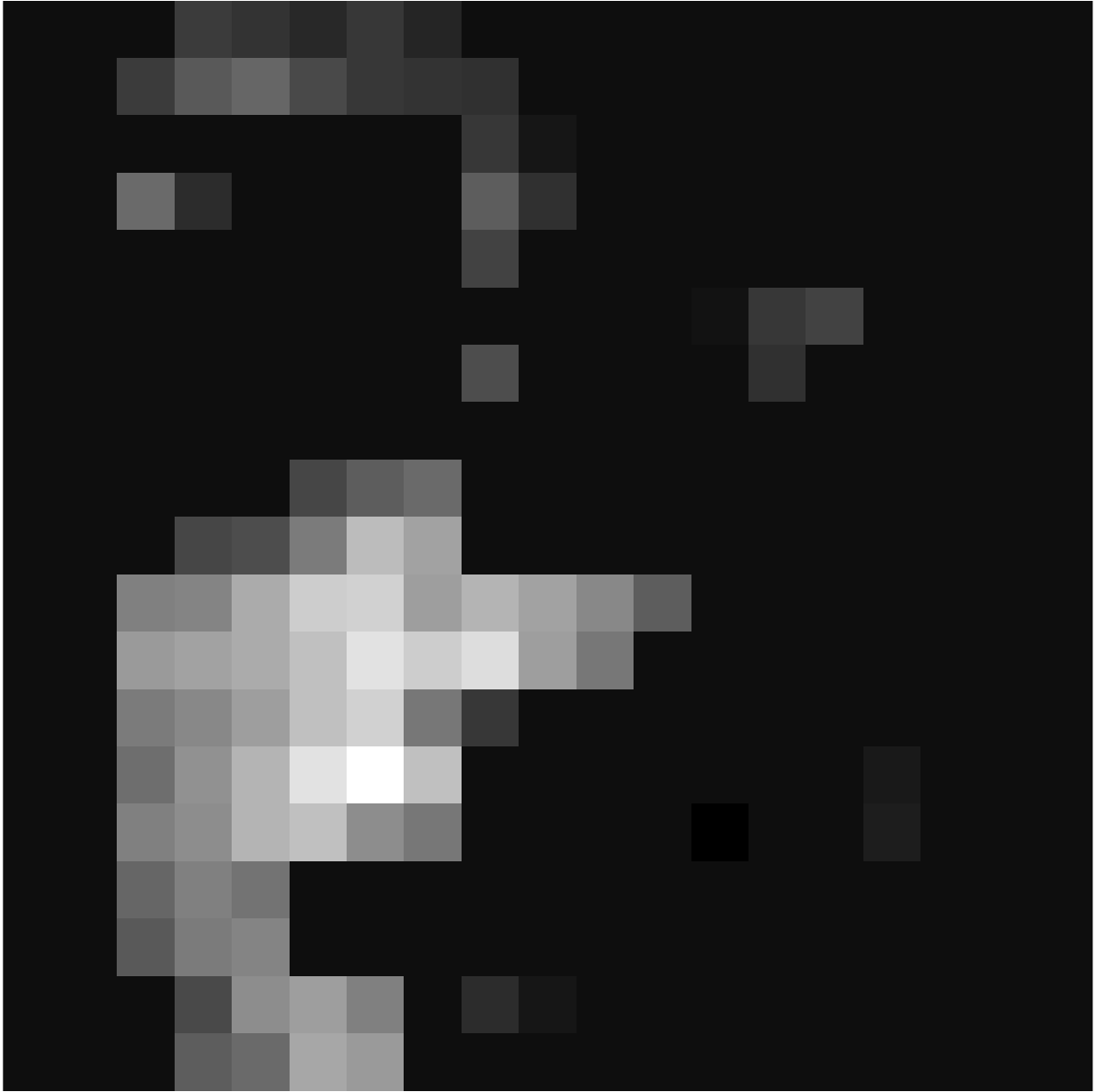}&\includegraphics[width=0.05\textwidth,height=0.05\textwidth]{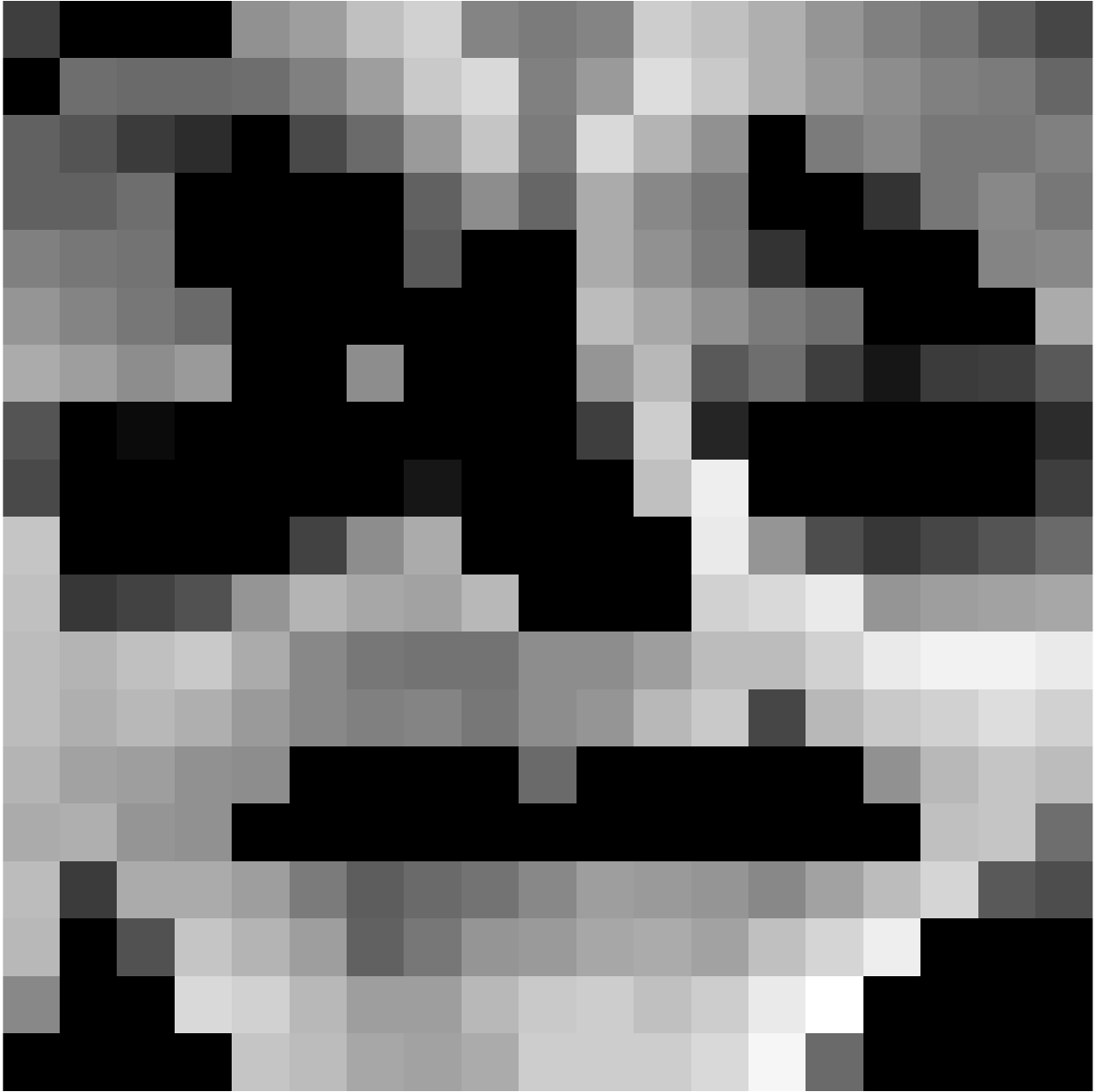}&\includegraphics[width=0.05\textwidth,height=0.05\textwidth]{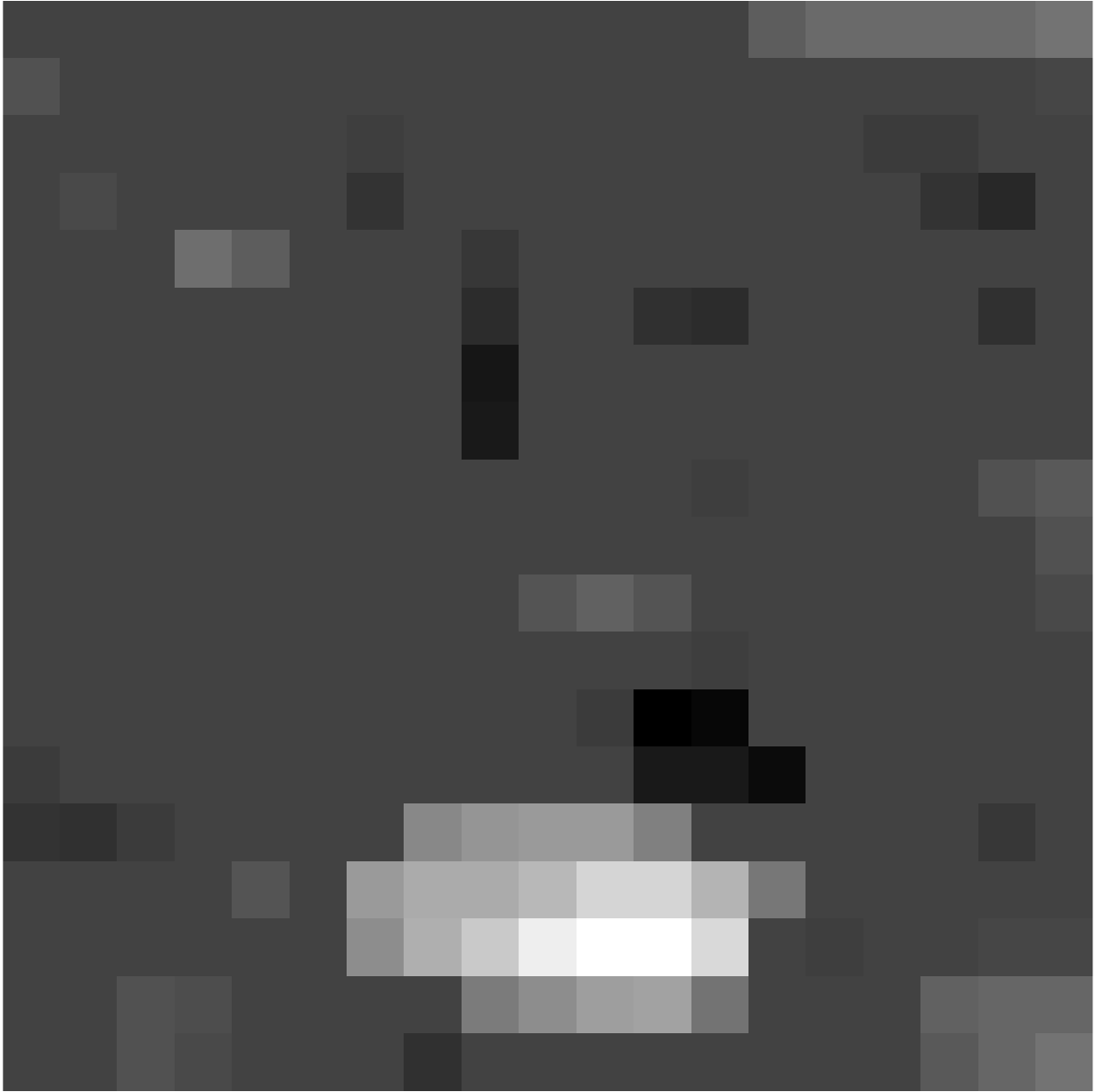}&\includegraphics[width=0.05\textwidth,height=0.05\textwidth]{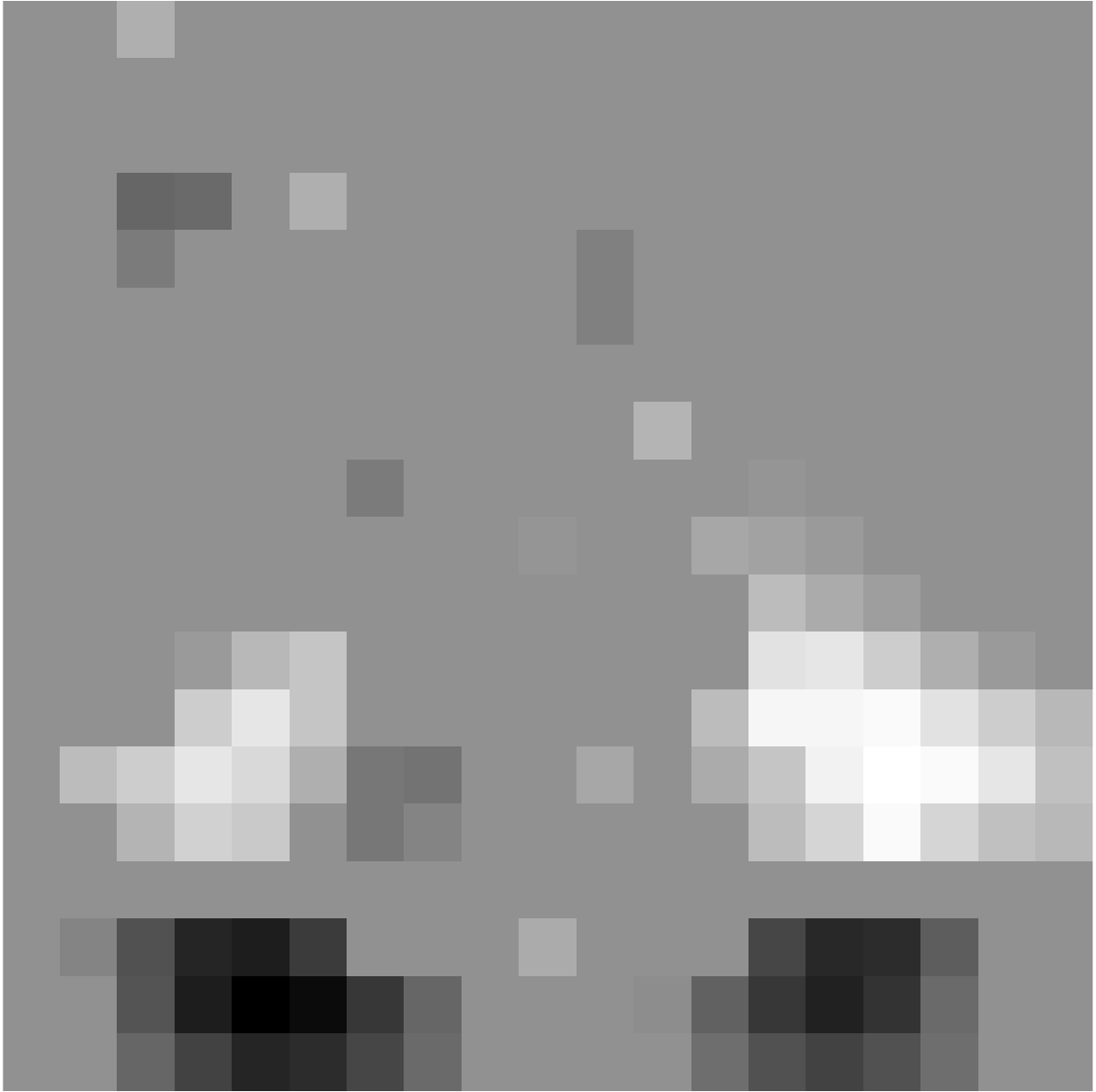}&\includegraphics[width=0.05\textwidth,height=0.05\textwidth]{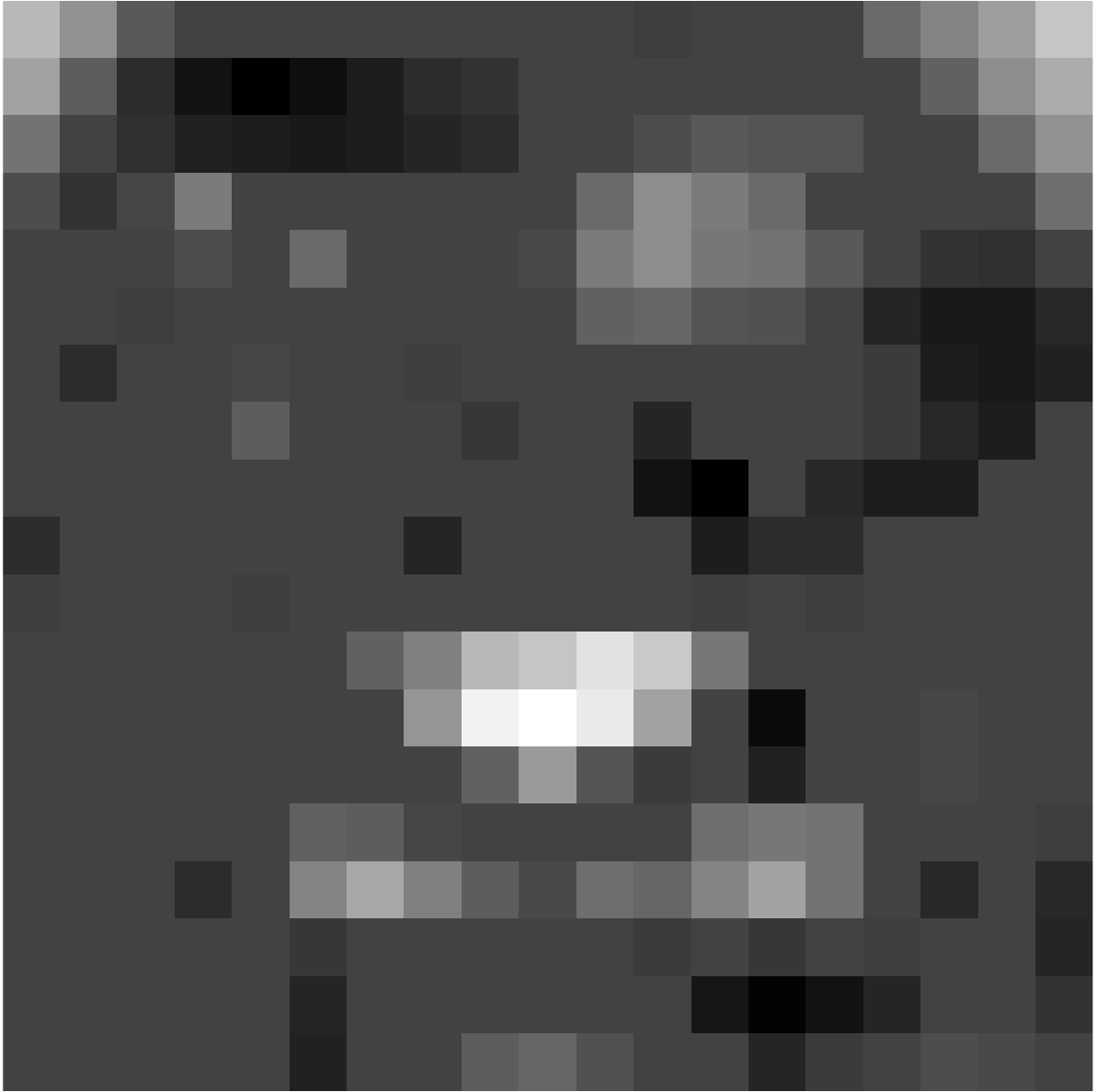} \\
  \includegraphics[width=0.05\textwidth,height=0.05\textwidth]{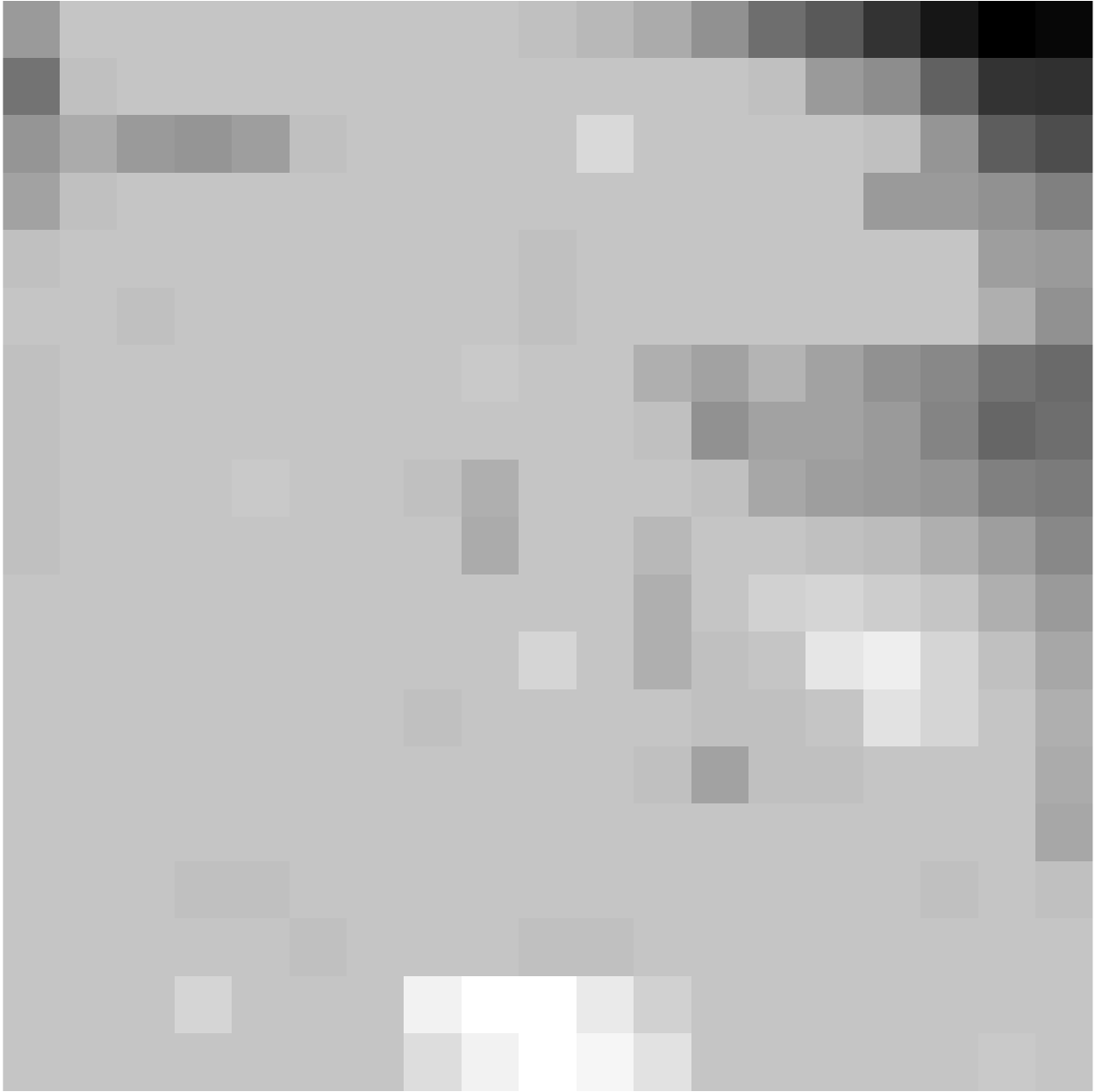}&\includegraphics[width=0.05\textwidth,height=0.05\textwidth]{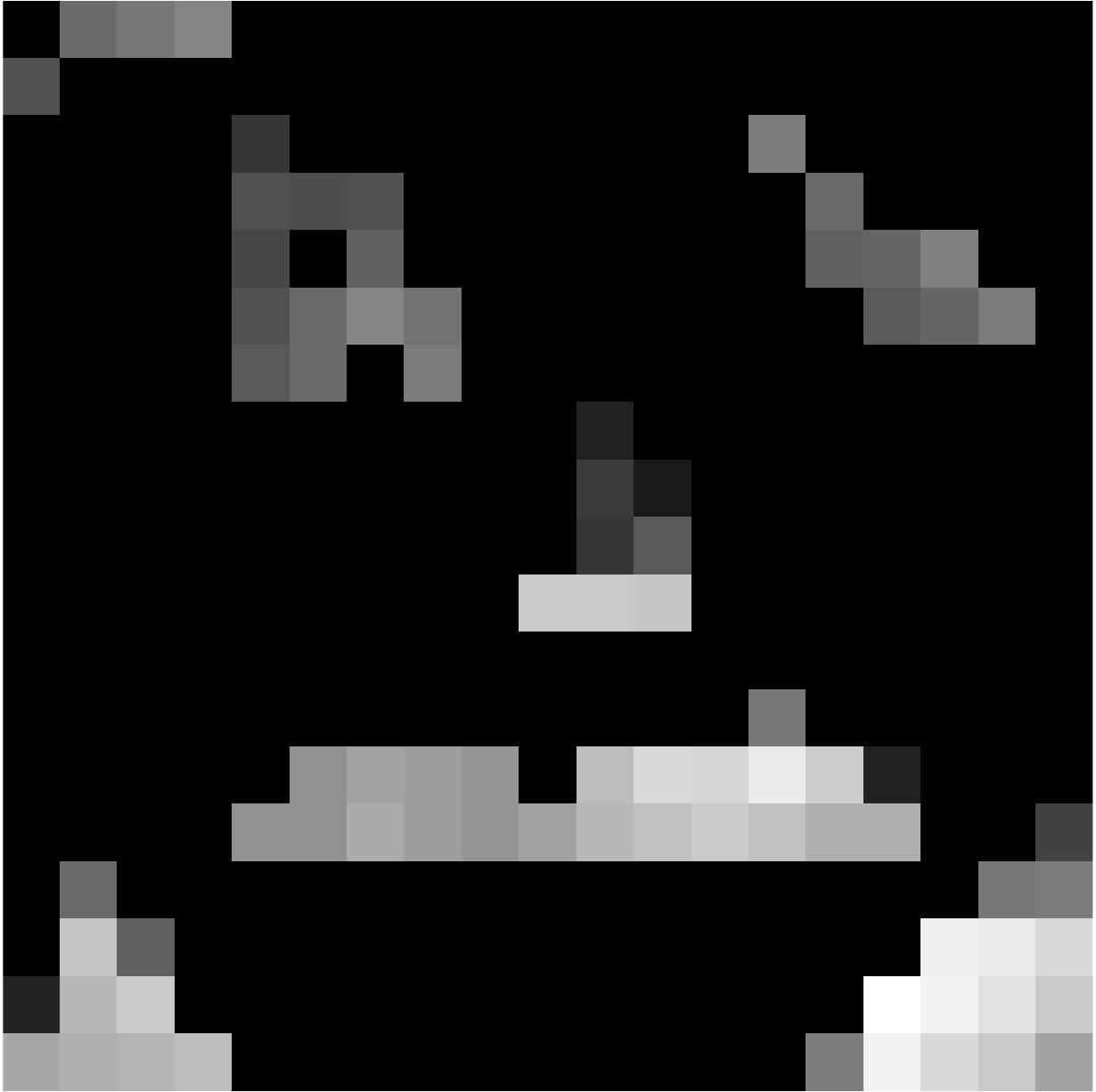}&\includegraphics[width=0.05\textwidth,height=0.05\textwidth]{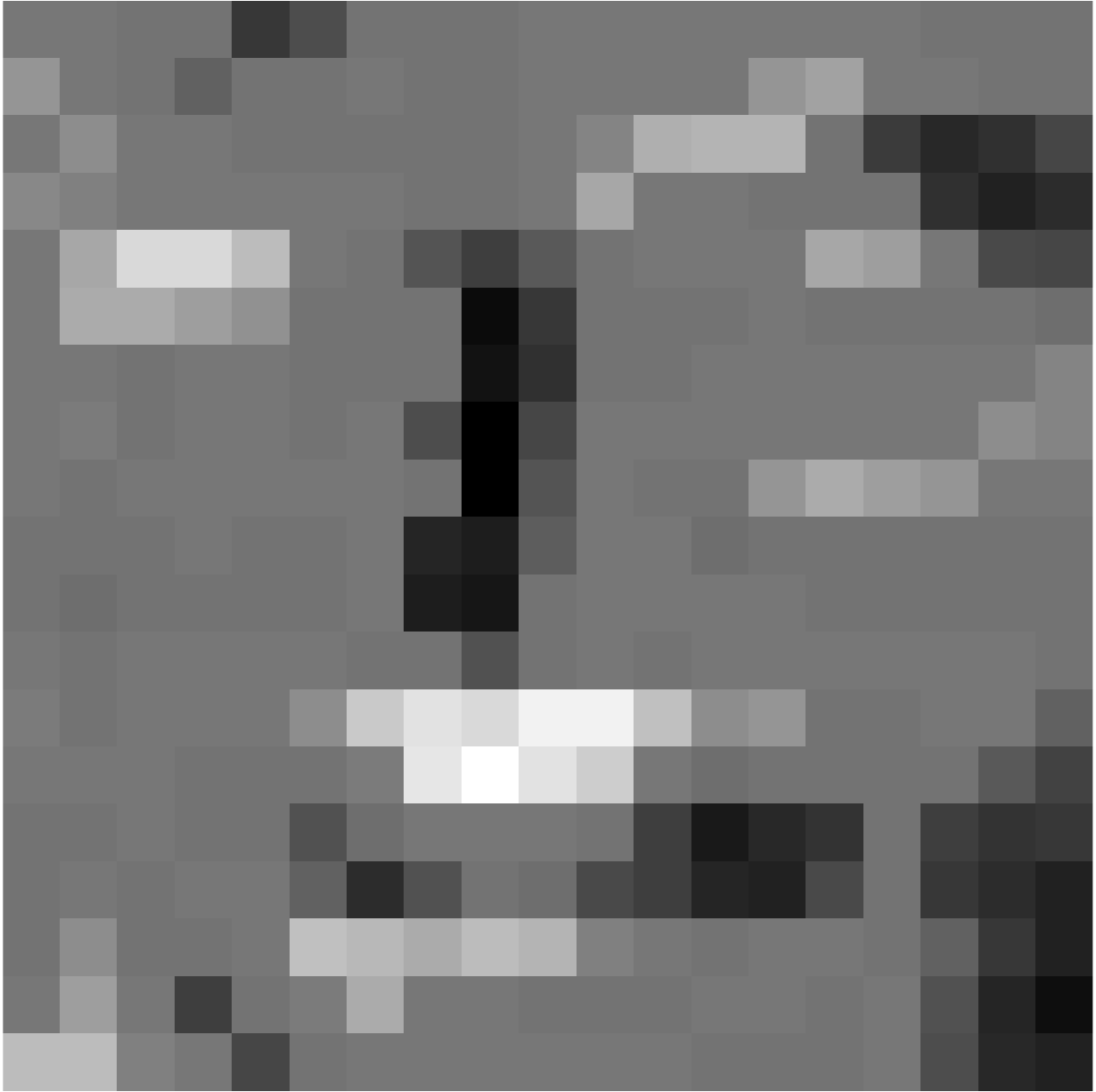}&\includegraphics[width=0.05\textwidth,height=0.05\textwidth]{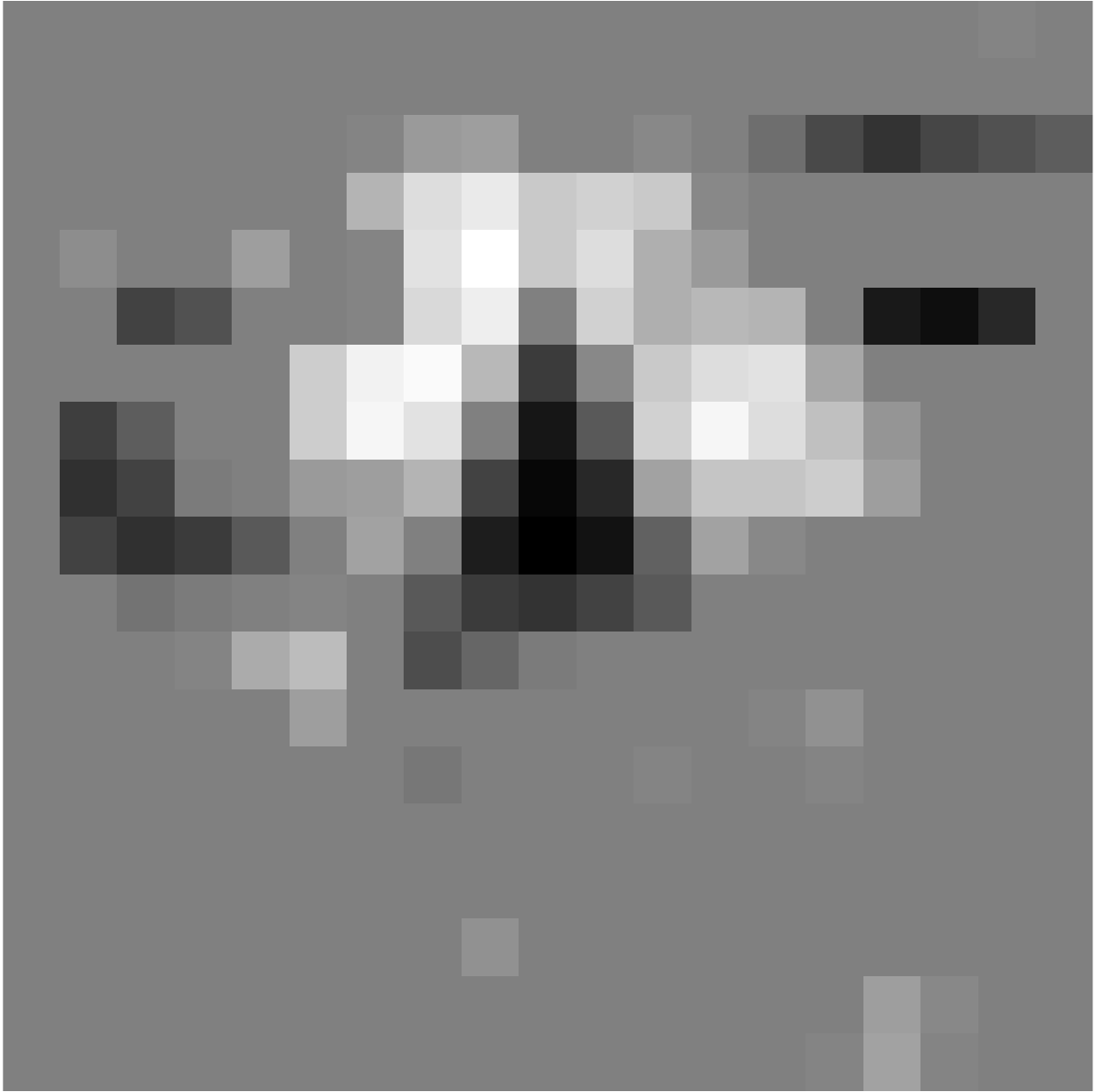}&\includegraphics[width=0.05\textwidth,height=0.05\textwidth]{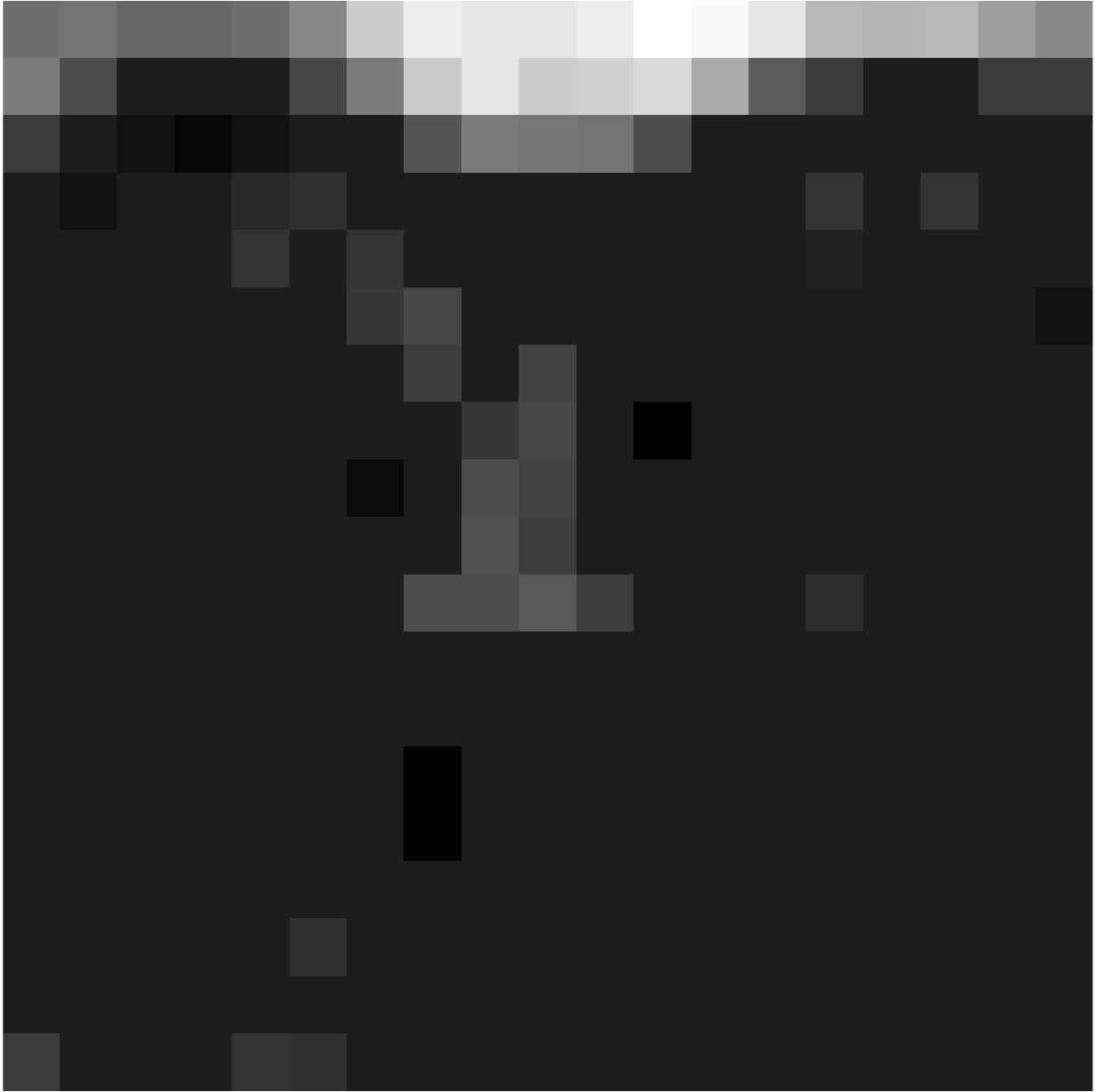}&\includegraphics[width=0.05\textwidth,height=0.05\textwidth]{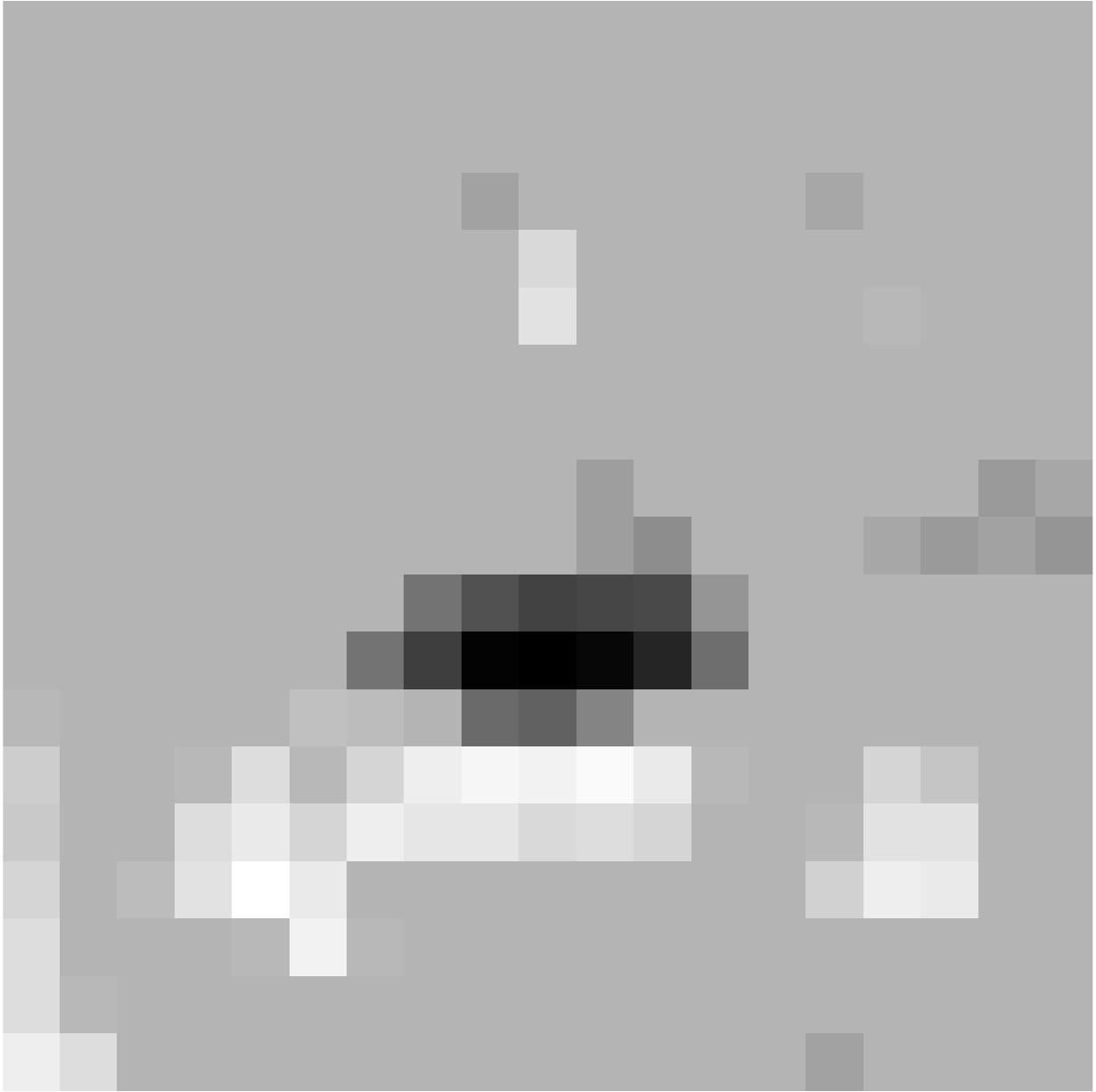}&\includegraphics[width=0.05\textwidth,height=0.05\textwidth]{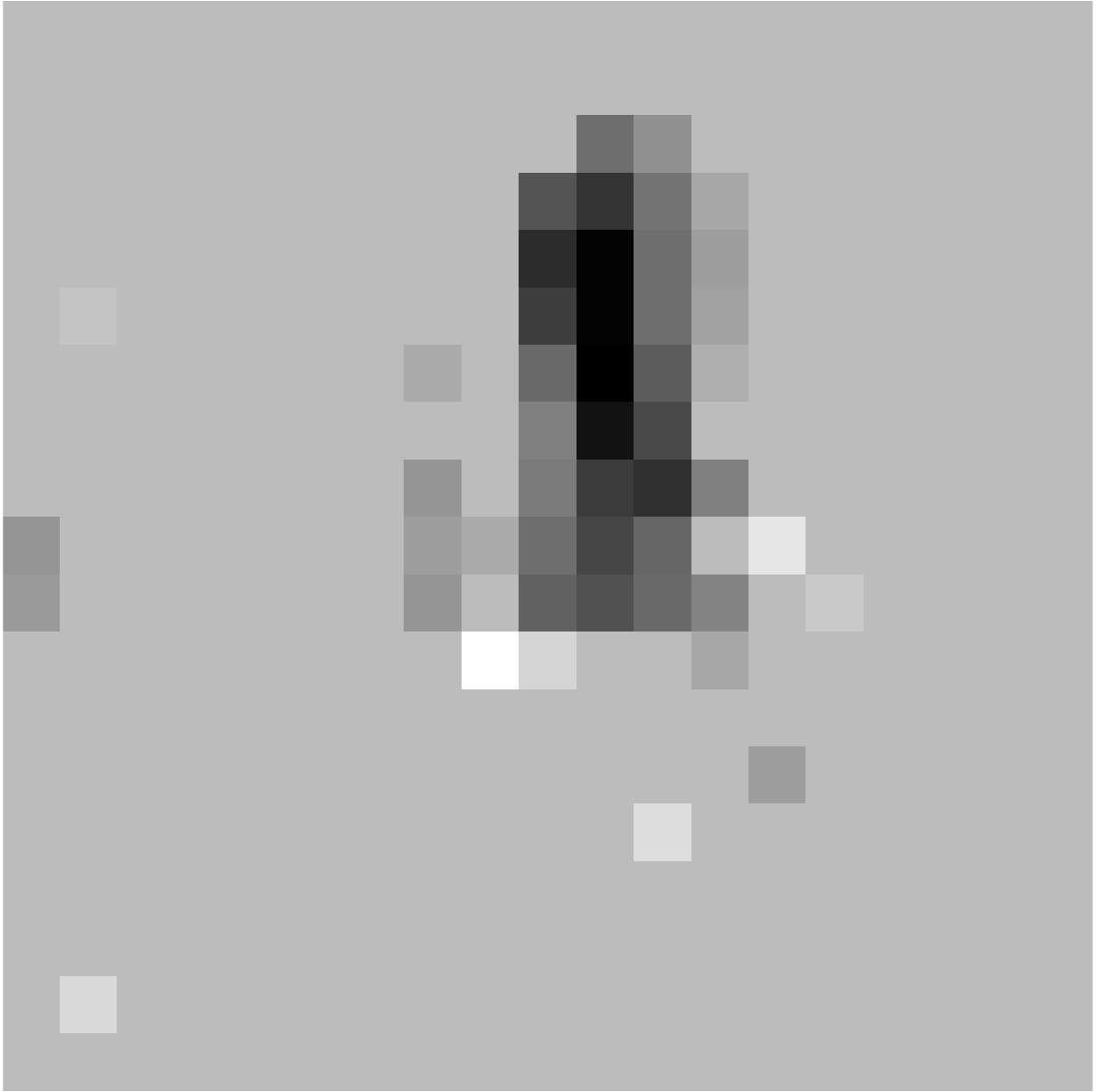} \\
  \includegraphics[width=0.05\textwidth,height=0.05\textwidth]{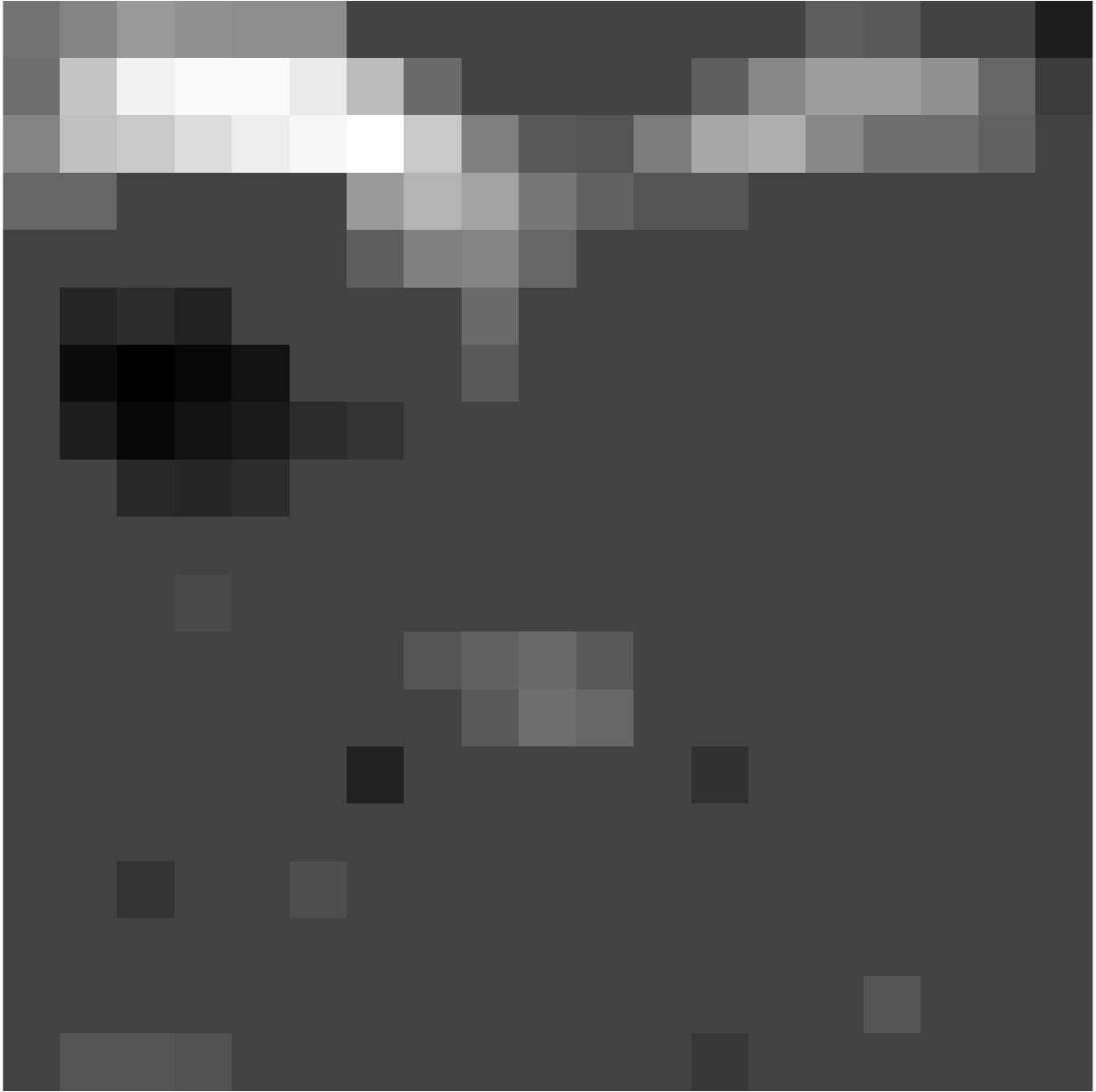}&\includegraphics[width=0.05\textwidth,height=0.05\textwidth]{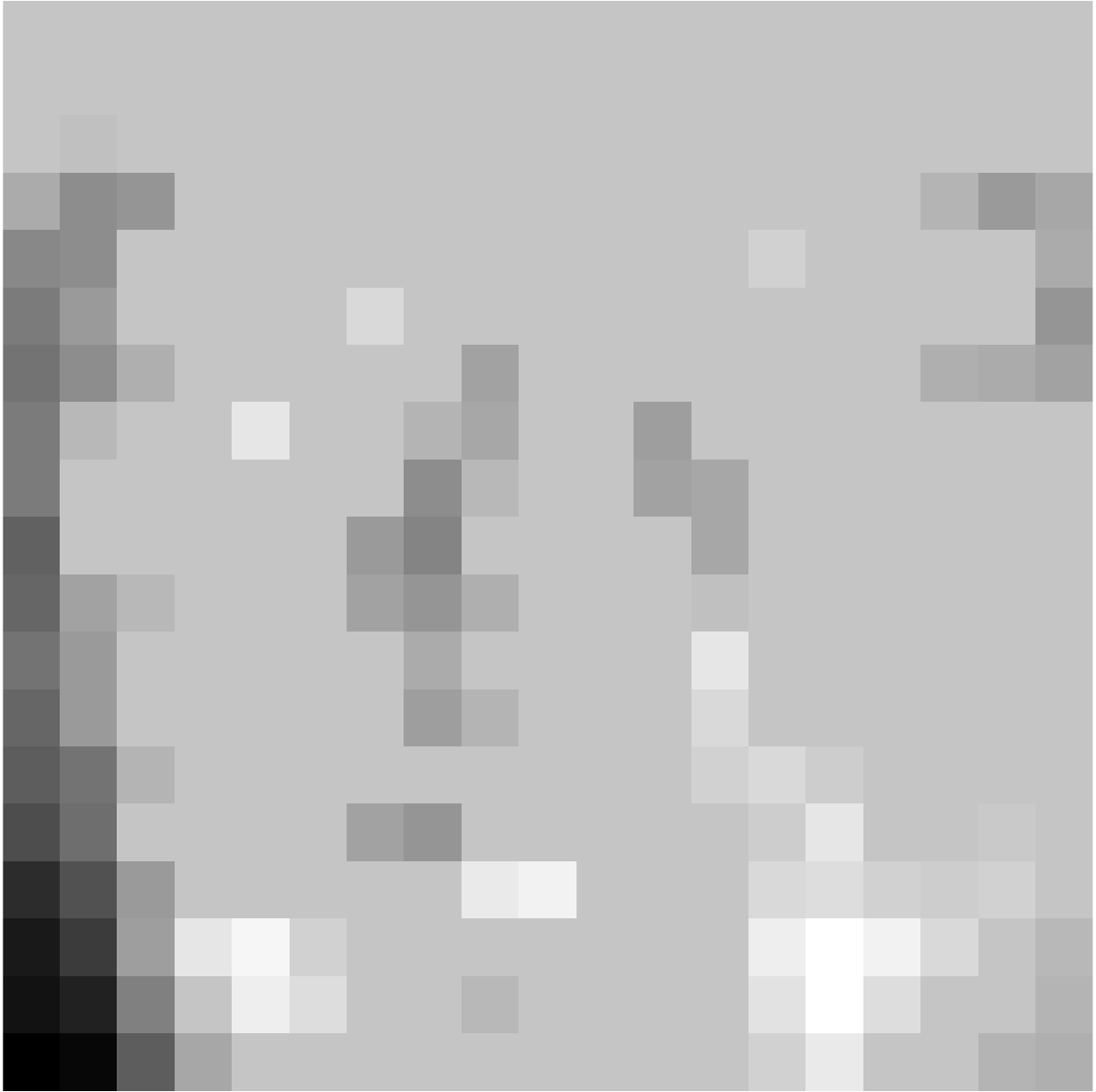}&\includegraphics[width=0.05\textwidth,height=0.05\textwidth]{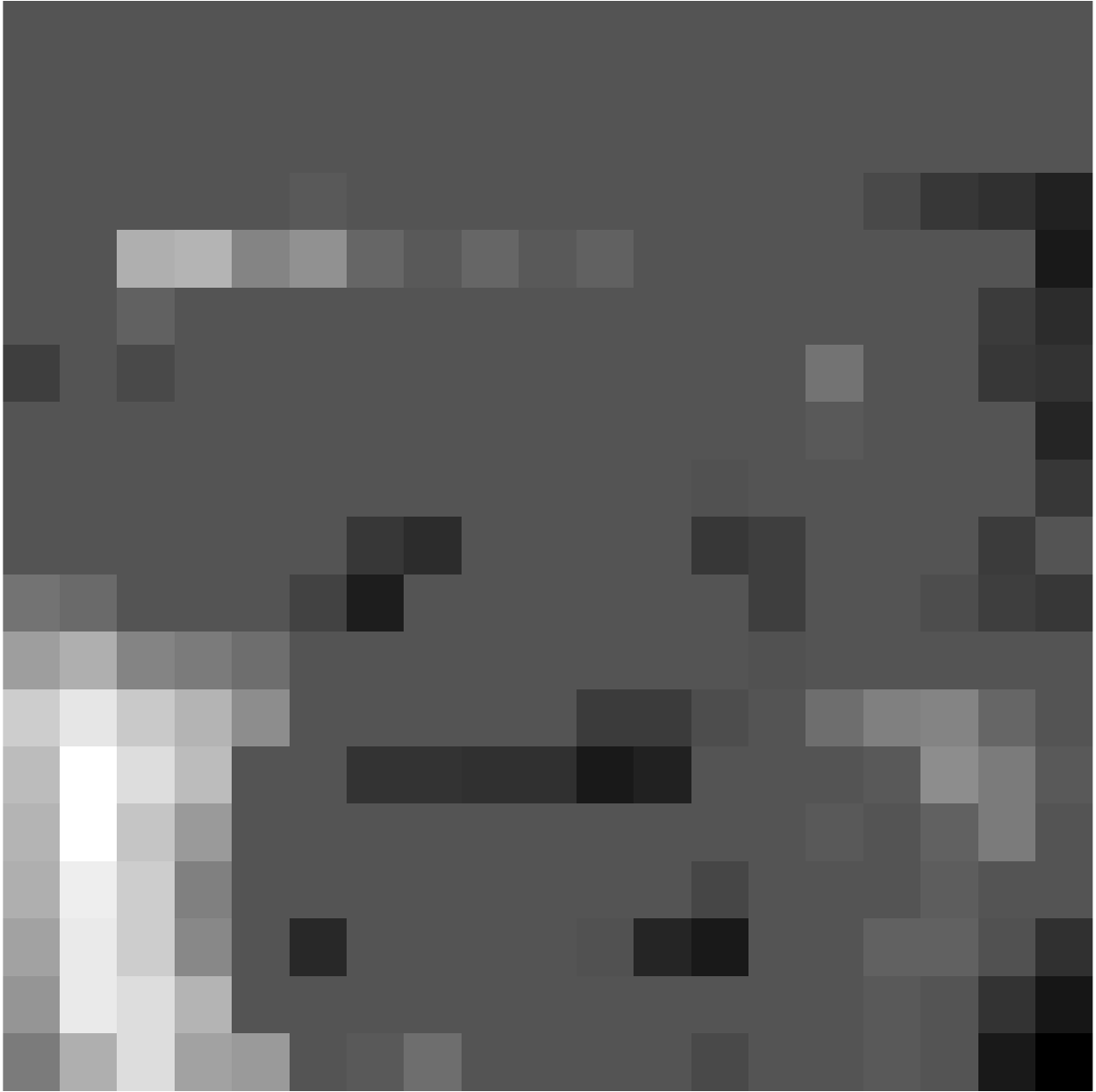}&\includegraphics[width=0.05\textwidth,height=0.05\textwidth]{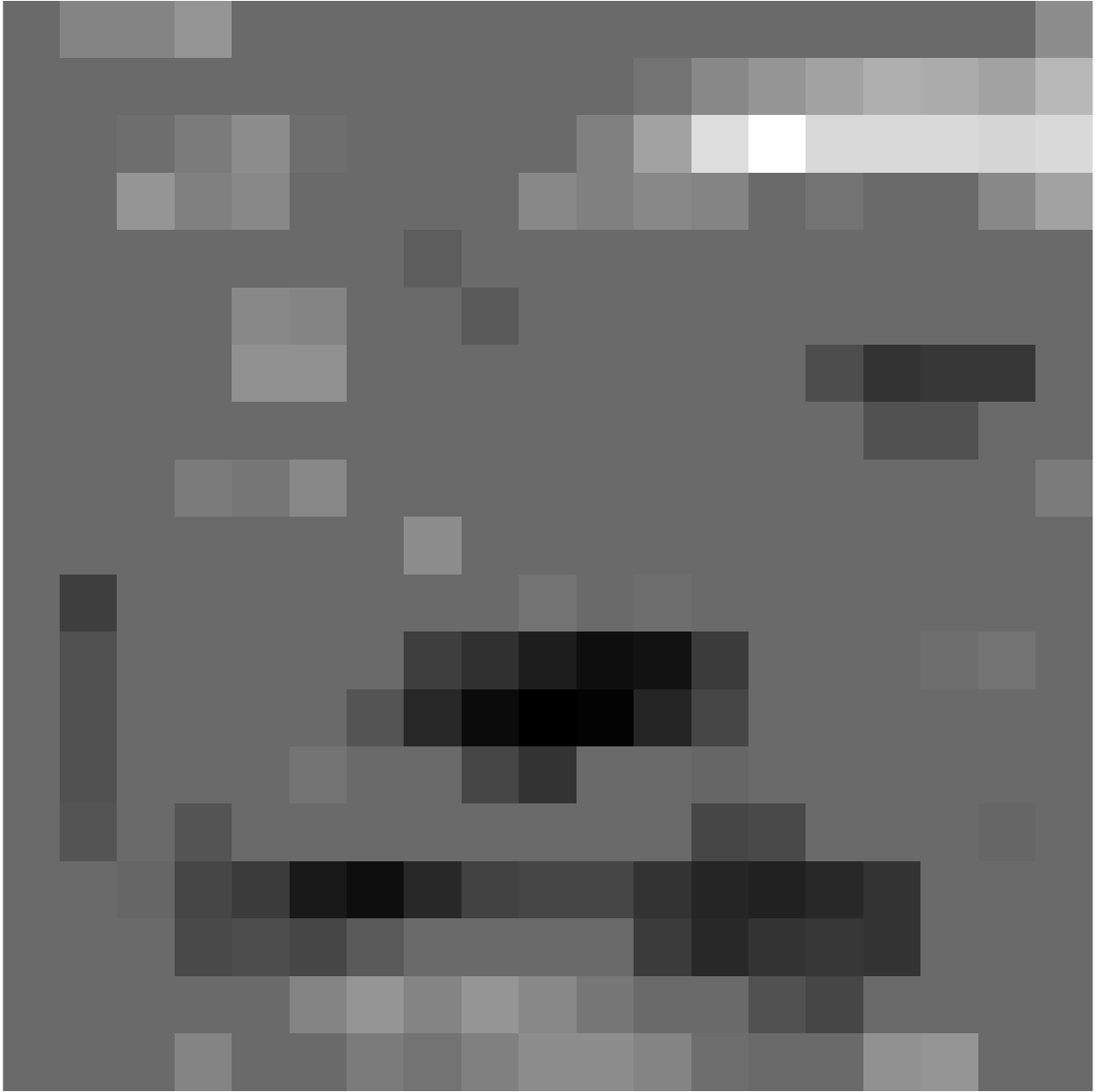}&\includegraphics[width=0.05\textwidth,height=0.05\textwidth]{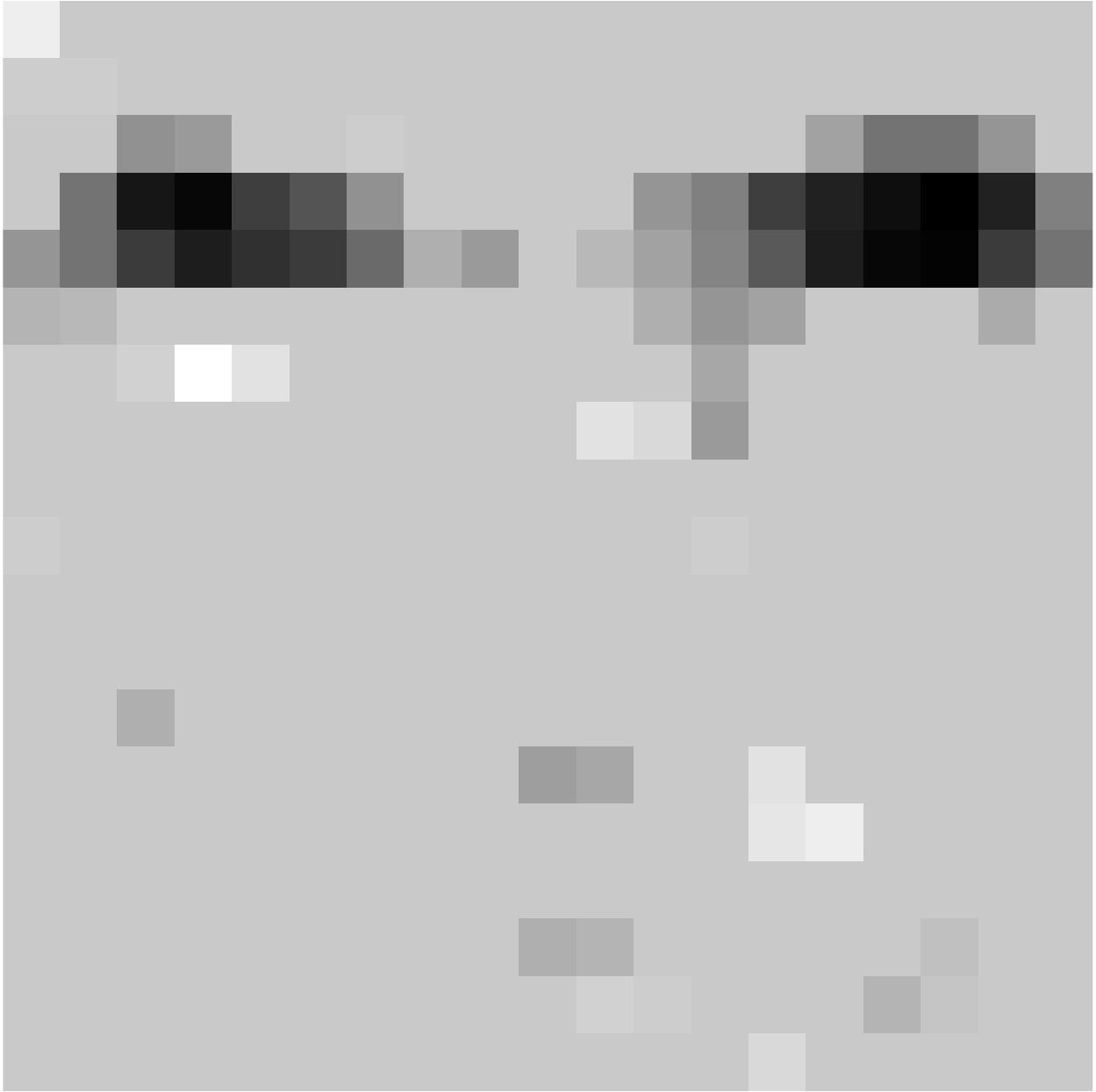}&\includegraphics[width=0.05\textwidth,height=0.05\textwidth]{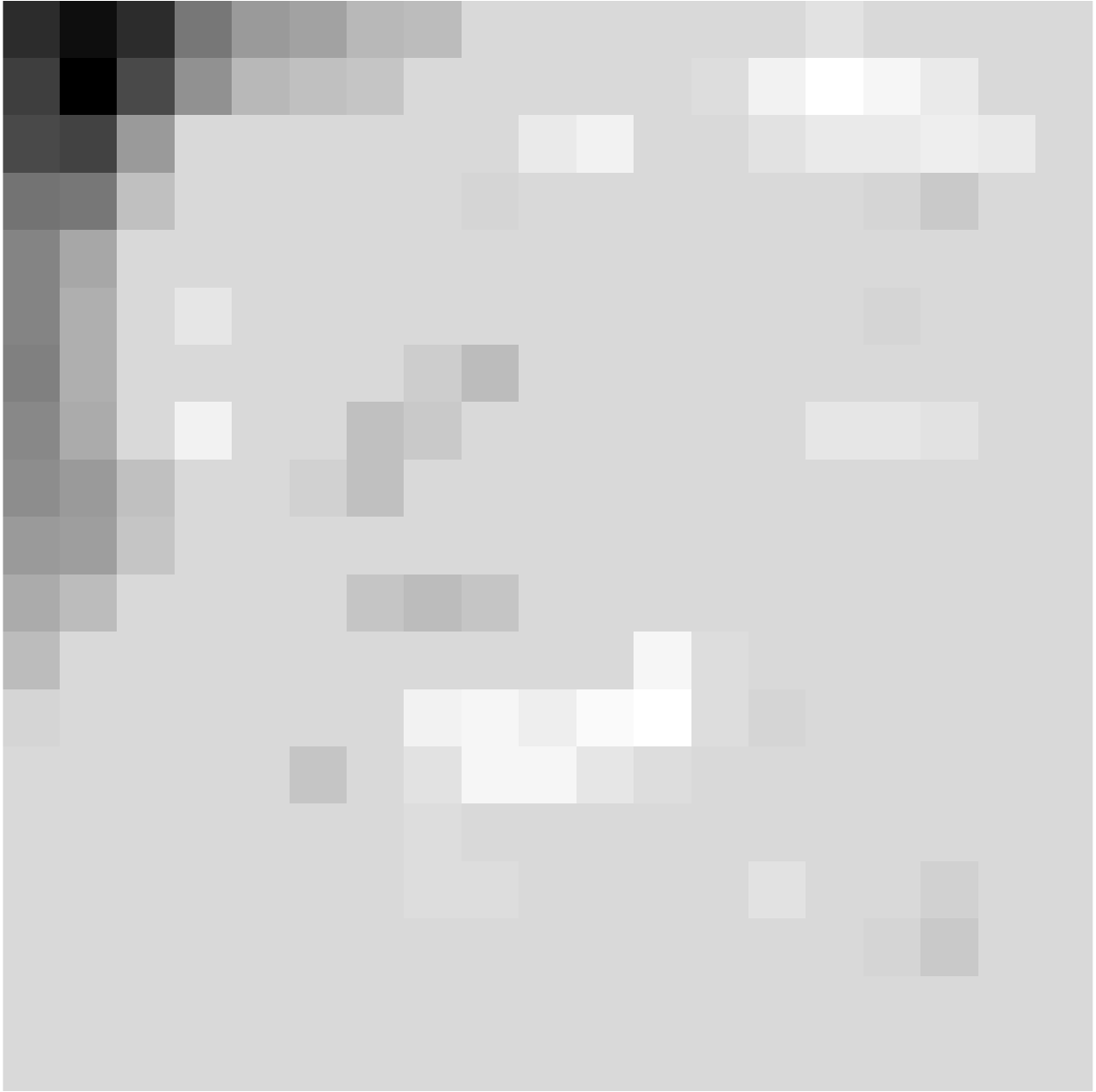}&\includegraphics[width=0.05\textwidth,height=0.05\textwidth]{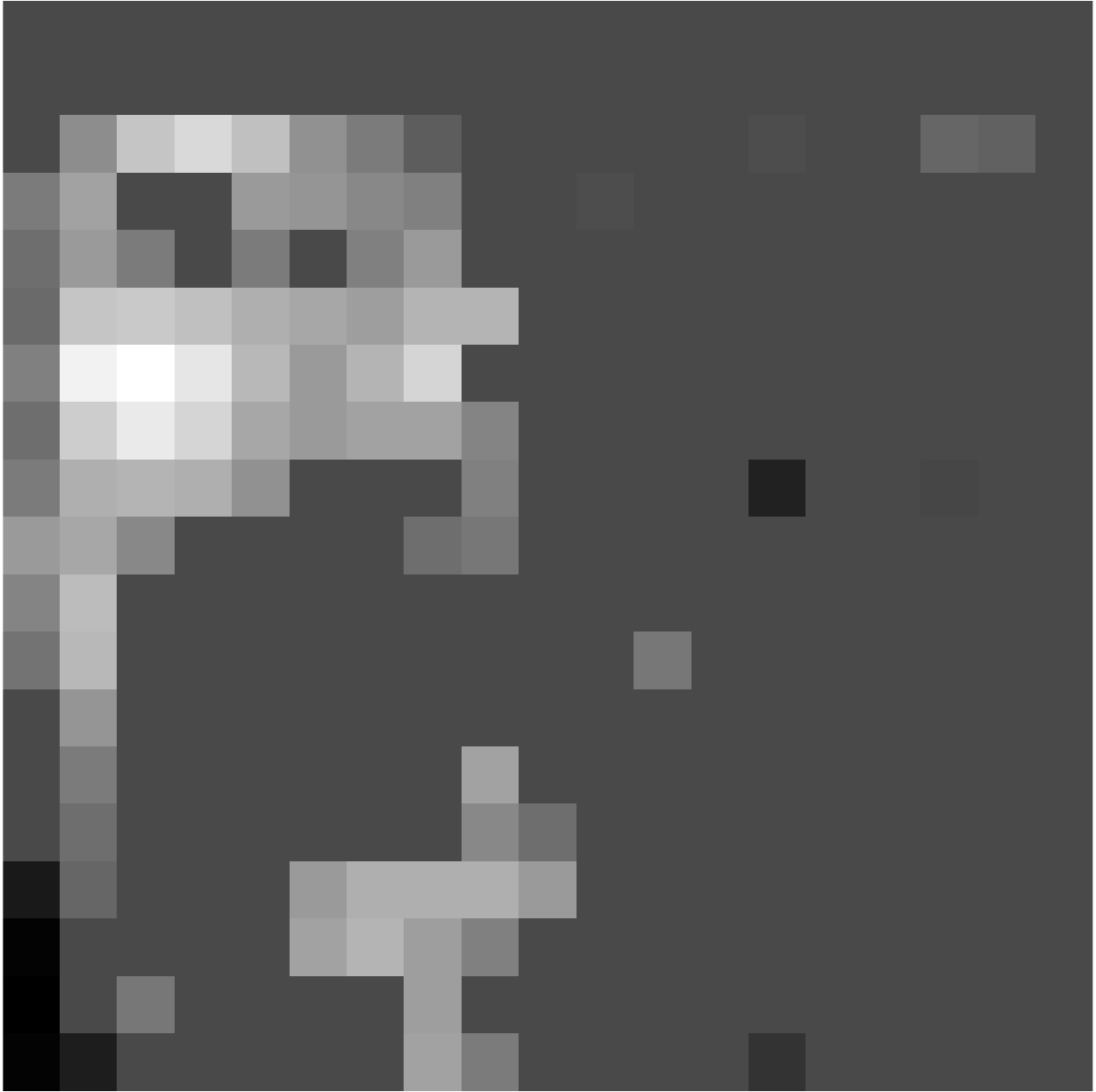}
 \end{tabular} \\
 \small b) sparse OVBSL
 \end{tabular}
 \caption{Eigenfaces obtained by sparse and non-sparse versions OVBSL on MIT-CBCL dataset.}
 \label{exp:fig_5}
\end{figure}
\subsubsection{Online eigenface learning}
In this section, we qualitatively evaluate the performance of sparse OVBSL as compared to its non-sparse version on a real dataset. Towards this, we use the MIT-CBCL face dataset \cite{sung1996learning}, which contains $n=2429$ face images  of size $19\times 19$ pixels. OVBSL processes the images as $K$-dimensional vectors with $K=361 (=19^2)$, in an online fashion. The subspace matrix estimated by both versions of OVBSL can be deemed as a {\it learned dictionary} of faces. In doing so, each image can be reconstructed by a linear combination of the atoms (eigenfaces) contained in the subspace matrix. The rank of the subspace is initialized for both algorithms to 50. Fig. \ref{exp:fig_5} shows the 20 leading eigenfaces in a descending order, according to their associated eigenvalues. Dark pixels correspond to negative values, while positive values are denoted with light colors. As it can be noticed, sparsity imposition from the sparse version of OVBSL leads to eigenfaces that present more localized features, contrary to its non-sparse counterpart, where features are spread out over the image. It should be also noted that both versions of OVBSL converged to a subspace matrix of low-rank. This fact resulted from the inherent advantageous characteristic of OVBSL to eliminate components presenting low variance, hence offering negligible information. 

\section{Conclusions}
In this paper, a novel online  variational Bayes subspace learning (OVBSL) algorithm from incomplete data was presented. Two basic merits of the proposed approach are: a) the imposition of low-rankness on the sought subspace by utilizing a novel group sparsity based heuristic, and b) the sparsity promotion on the subspace matrix. The former characteristic makes the algorithm robust in the absence of the knowledge of the true rank while the latter renders it amenable to sparse dictionary learning problems. OVBSL belongs to the family of Bayesian algorithms thus, contrary to its deterministic counterparts, no parameter fine-tuning is required. The effectiveness of the proposed algorithm is verified in a variety of experiments conducted on simulated and real data. Subspace tracking from partial observations, treated in this paper, can be also viewed as an online matrix completion task that has a major impact in numerous applications. By suitably extending and modifying the proposed Bayesian model and methodology, similar problems of high importance such as online nonnegative matrix factorization and online robust PCA can be potentially tackled. This extension as well as the unification of all the different schemes under a common umbrella is the subject of our current investigation. 
\appendix \label{sec:apendix}
The joint prior of $\mathbf{X}$ and $\mathbf{W}$ is expressed as 
\begin{align}
p(\mathbf{X},\mathbf{W} \mid \boldsymbol{\delta},\beta,\boldsymbol{\Gamma}) = \prod_{l=1}^{L}p(\boldsymbol{ \mathit{x}}_l, \boldsymbol{\mathit{w}}_l \mid \delta_l,\beta,\boldsymbol{\Gamma}_l) \label{eq:jointXW}
\end{align}
where
\begin{align}
p(\boldsymbol{\mathit{x}}_l, \boldsymbol{\mathit{w}}_l \mid \delta_l,\beta,\boldsymbol{\Gamma}_l) = \int_{0}^{\infty} p(\boldsymbol{\mathit{x}}_l \mid s_l,\beta) p(\boldsymbol{\mathit{w}}_l \mid s_l,\beta,\boldsymbol{\Gamma}_l)p(s_l \mid \delta_l)ds_l
\label{eq:jointxlwl}
\end{align}
Using (\ref{prior:X}), (\ref{prior:W}) and (\ref{prior:s}) in (\ref{eq:jointxlwl}) yields
\begin{align}
p(\boldsymbol{\mathit{x}}_l,\boldsymbol{\mathit{w}}_l \mid \delta_l,\beta, \boldsymbol{\Gamma}_l) 
= \int_{0}^{\infty} (2\pi)^{-\frac{n+K}{2}}\beta^{\frac{n+K}{2}}|\boldsymbol{\Lambda}\boldsymbol{\Gamma}_l|^{\frac{1}{2}}s_l^{-\frac{3}{2}}\exp \left(-\frac{\beta s_l}{2}(\left\|\boldsymbol{\mathit{x}}_l\right\|_{2,\boldsymbol{\Lambda}}^2 + \left\|\boldsymbol{\mathit{w}}_l\right\|_{2,\boldsymbol{\Gamma}_l}^2) - \frac{\delta_l}{2s_l}\right)ds_l, \label{eq:jxl1}
\end{align}
where $\left\|\boldsymbol{\mathit{x}}_l\right\|_{2,\boldsymbol{\Lambda}}^2 = \boldsymbol{\mathit{x}}_l^T\boldsymbol{\Lambda}\boldsymbol{\mathit{x}}_l$ and $\|\boldsymbol{\mathit{w}}_l\|^2_{2,\boldsymbol{\Gamma}_l} = \boldsymbol{\mathit{w}}^T_l\boldsymbol{\Gamma}_l\boldsymbol{\mathit{w}}_l$. Using in (\ref{eq:jxl1}) the expression of the GIG distribution for $s_l$ with parameters $a = \beta(\left\|\boldsymbol{\mathit{x}}_l\right\|_{2,\boldsymbol{\Lambda}}^2 + \left\|\boldsymbol{\mathit{w}}_l\right\|_{2,\boldsymbol{\Gamma}_l}^2), b = \delta_l$ and $p = -1/2$, we easily get 
\begin{align}
p(\boldsymbol{\mathit{x}}_l,\boldsymbol{\mathit{w}}_l \mid \delta_l,\beta,\boldsymbol{\Gamma}_l)  = (2\pi)^{-\frac{n+K}{2}}\beta^{\frac{n+K}{2}}|\boldsymbol{\Lambda}\boldsymbol{\Gamma}_l|^{\frac{1}{2}} 2 {\cal K}_{-\frac{1}{2}}\left(\beta^{\frac{1}{2}}\delta_l^{\frac{1}{2}}(\left \|\boldsymbol{\mathit{x}}_l\right\|_{2,\boldsymbol{\Lambda}}^2 + \left\|\boldsymbol{\mathit{w}}_l\right\|_{2,\boldsymbol{\Gamma_l}}^2)^{\frac{1}{2}}\right) \left( \frac{\beta (\left \|\boldsymbol{\mathit{x}}_l\right\|_{2,\boldsymbol{\Lambda}}^2 + \left\|\boldsymbol{\mathit{w}}_l\right\|_{2,\boldsymbol{\Gamma}_l}^2)}{\delta_l} \right)^{\frac{1}{4}} \label{eq:jxl2}
\end{align}
By employing the identity,
\begin{align}
{\cal K}_{-\frac{1}{2}}(x) = \left( \frac{\pi}{2x}\right)^{\frac{1}{2}}\exp (-x)  \nonumber
\end{align}
in (\ref{eq:jxl2}) and after some straightforward calculations, we end up with the following expression for the joint distribution of $\mathbf{x}_l$ and $\mathbf{w}_l$,
\begin{align}
p(\boldsymbol{\mathit{x}}_l, \boldsymbol{\mathit{w}}_l \mid \delta_l,\beta,\boldsymbol{\Gamma}_l) = (2\pi)^{-\frac{n+K-1}{2}}\beta^{\frac{n+K}{2}}|\boldsymbol{\Lambda}\boldsymbol{\Gamma}_l|^{\frac{1}{2}}\delta_l^{-\frac{1}{2}}\exp \left(-\beta^{\frac{1}{2}}\delta_l^{\frac{1}{2}}(\left \|\boldsymbol{\mathit{x}}_l\right\|_{2,\boldsymbol{\Lambda}}^2 + \left\|\boldsymbol{\mathit{w}}_l\right\|_{2,\boldsymbol{\Gamma}_l}^2)^{\frac{1}{2}} \right).
\end{align}
Then, from (\ref{eq:jointXW}) 
\begin{align}
p(\mathbf{X},\mathbf{W} \mid \boldsymbol{\delta},\beta,\boldsymbol{\Gamma}) = (2\pi)^{-\frac{(n+K-1)L}{2}}\beta^{\frac{(n+K)L}{2}}|\boldsymbol{\Lambda}|^{\frac{L}{2}}\left( \prod_{l=1}^{L}\delta_l^{\frac{1}{2}} |\boldsymbol{\Gamma}_l|^{\frac{1}{2}}\right)\exp \left(-\beta^{\frac{1}{2}}\sum_{l=1}^{L}\delta_l^{\frac{1}{2}}(\left \|\boldsymbol{\mathit{x}}_l\right\|_{2,\boldsymbol{\Lambda}}^2 + \left\|\boldsymbol{\mathit{w}}_l\right\|_{2,\boldsymbol{\Gamma}_l}^2)^{\frac{1}{2}} \right)
\end{align} 
which is a multi-parameter (with respect to the $\delta_l$'s) Lalpace-type distribution defined on the columns of the matrix $[\boldsymbol{(\Lambda}^{1/2}\mathbf{X})^T \;\; \left(\boldsymbol{\Gamma}\odot\mathbf{W}\right)^T]^T$. Such a distribution is known to impose column sparsity and thus, due to the form of the matrix, joint column sparsity on $\mathbf{X}$ and $\mathbf{W}$. 

\bibliographystyle{IEEEtran}
\bibliography{IEEEabrv,refs_report}
\end{document}